\documentclass[a4paper,fleqn]{cas-sc}

\usepackage[authoryear]{natbib}
\usepackage{graphicx} 
\usepackage{float}
\usepackage{algorithm}  
\usepackage{algpseudocode}
\usepackage{color}
\usepackage{setspace}
\usepackage{array}
\usepackage{multirow}
\usepackage{todonotes}
\usepackage{subcaption}
\usepackage{url}
\usepackage{threeparttable}
\usepackage{booktabs}
\usepackage{pifont}
\usepackage{longtable}
\usepackage{siunitx}
\usepackage{cleveref}

\newcommand{\xmark}{\ding{55}}
 
\newif\ifshowcover
\showcoverfalse
 
\def\tsc#1{\csdef{#1}{\textsc{\lowercase{#1}}\xspace}}
\tsc{WGM}
\tsc{QE}
\tsc{EP}
\tsc{PMS}
\tsc{BEC}
\tsc{DE}

\usepackage{lineno}

\begin{document}
\let\WriteBookmarks\relax
\def\floatpagepagefraction{1}
\def\textpagefraction{.001}
\shorttitle{CLOSP:  A Unified Semantic Space for SAR, MSI, and Text in Remote Sensing}
\shortauthors{D. Rege Cambrin et al.}

\title [mode = title]{CLOSP: A Unified Semantic Space for SAR, MSI, and Text in Remote Sensing}

\author[polito]{Daniele {Rege Cambrin}}[orcid=0000-0002-5067-2118]
\credit{Conceptualization, Methodology, Software, Investigation, Data Curation, Visualization, Writing - Original Draft}

\author[polito]{Lorenzo Vaiani}
\credit{Visualization, Software, Conceptualization, Writing - Original Draft}

\author[polito]{Giuseppe Gallipoli}
\credit{Validation, Writing - Original Draft, Data Curation}

\author[polito]{Luca Cagliero}
\credit{Supervision, Writing - Review \& Editing}

\author[polito]{Paolo Garza}
\credit{Supervision, Writing - Review \& Editing}

\address[polito]{Politecnico di Torino, Corso Duca Degli Abruzzi, Turin, Italy}

\begin{abstract}
Retrieving relevant imagery from vast satellite archives is crucial for applications like disaster response and long-term climate monitoring. However, most text-to-image retrieval systems are limited to RGB data, failing to exploit the unique physical information captured by other sensors, such as the all-weather structural sensitivity of Synthetic Aperture Radar (SAR) or the spectral signatures in optical multispectral data. To bridge this gap, we introduce CrisisLandMark, a new large-scale corpus of over 647,000 Sentinel-1 SAR and Sentinel-2 multispectral images paired with structured textual annotations for land cover, land use, and crisis events harmonized from authoritative land cover systems (CORINE and Dynamic World) and crisis-specific sources.
We then present CLOSP (\underline{C}ontrastive \underline{L}anguage \underline{O}ptical \underline{S}AR \underline{P}retraining), a novel framework that uses text as a bridge to align unpaired optical and SAR images into a unified embedding space enabling effective text-based retrieval from heterogeneous sources. Our experiments show that CLOSP achieves a new state-of-the-art, improving retrieval nDGC@1000 by 54\% over existing models. Additionally, we find that the unified training strategy overcomes the inherent difficulty of interpreting SAR imagery by transferring rich semantic knowledge from the optical domain with indirect interaction. Furthermore, GeoCLOSP, which integrates geographic coordinates into our framework, creates a powerful trade-off between generality and specificity: while the CLOSP excels at general semantic tasks, the GeoCLOSP becomes a specialized expert for retrieving location-dependent crisis events and rare geographic features. This work highlights that the integration of diverse sensor data and geographic context is essential for unlocking the full potential of remote sensing archives.
\end{abstract}


\maketitle 

\printcredits

\doublespacing

\section{Introduction}

\begin{figure}[pos=t]
  \centering
  \includegraphics[width=\textwidth]{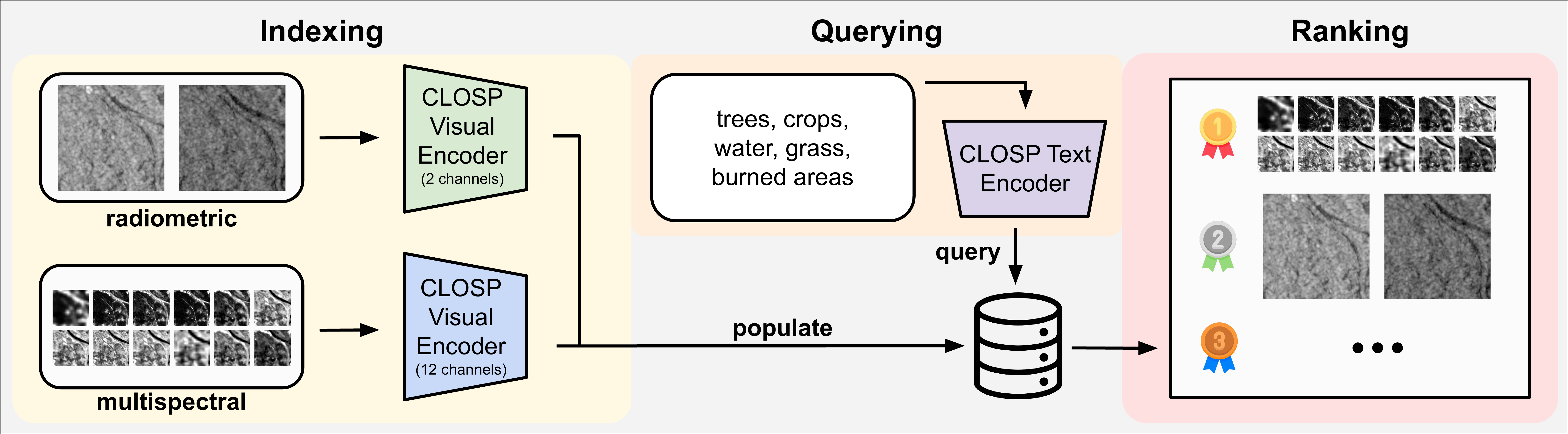}
  \caption{
  Retrieval process based on CLOSP, the multimodal text-image encoder proposed in the present study. 
  Optical 12-channel and SAR 2-channel images available in the newly released \textit{CrisisLandMark} corpus are encoded using the CLOSP visual encoder and indexed in the database. 
  A textual query related to land cover is embedded on the fly using the CLOSP text encoder. 
  The retriever returns a ranked list of the most pertinent images to the input query, which may contain both optical and SAR images.
  }
  \label{fig:teaser}
\end{figure}

Satellite remote sensing has become an indispensable tool for monitoring our planet, providing critical data for many applications, including climate science, environmental management, disaster response, urban planning, and precision agriculture \citep{wulder2022,torres2012,drusch2012,goward2001,roy2014}. The large number of satellite missions provides a great volume and diversity of Earth observation data. However, this availability presents a significant challenge: effectively searching and retrieving relevant imagery from massive, heterogeneous, and multimodal archives, particularly when the search criteria are expressed in natural language. To address this challenge, the field of Text-to-Remote-Sensing-Image Retrieval (T2RSIR) has emerged, aiming to bridge the semantic gap between textual descriptions and the rich visual content of satellite imagery. The task involves indexing a vast database of images and retrieving a ranked list of the most pertinent examples that match a user's natural language query.

The advances in multimodal learning, particularly the adaptation of contrastive learning frameworks like CLIP \citep{Radford2021}, have driven the recent progress in T2RSIR. Models such as SkyCLIP \citep{wang2024}, SenCLIP \citep{liu2024}, and RemoteCLIP \citep{jain2024} have demonstrated success in aligning text with remote sensing imagery. However, these efforts have focused on high-resolution aerial or RGB satellite images. This reliance on the visible spectrum overlooks the information contained in other modalities and limits their applicability. For instance, missions like Sentinel-1 provide Synthetic Aperture Radar (SAR) provides all-weather, day-and-night acquisitions, making it highly effective for land cover mapping, change detection, and infrastructure monitoring \citep{clasen2024, cambrin2024, li2020}. Concurrently, multispectral imagery (MSI) from sensors like Sentinel-2 offers crucial spectral bands beyond RGB that are essential for analyzing vegetation health, water bodies, and soil properties \citep{thenkabail2004, gao1996, huete1988soil}. While some models like SatCLIP \citep{klemmer2024} or Llama3-MS-CLIP \citep{marimo2025} have begun to leverage multispectral data, they both disregard the role of SAR data, and SatCLIP aligns it only with geographic coordinates. The full potential of combining these diverse sensor types into a unified retrieval framework remains largely underexplored.

This points to two gaps in the current state-of-the-art. First, existing T2RSIR corpora are inadequate for developing and evaluating rich and robust multimodal systems. Many are limited by small scale (e.g., RSICD, RSITMD \citep{lu2017, zhiqiang2022}), a dependency on RGB-only aerial imagery, and a lack of rich sensor data like MSI and SAR. Furthermore, their annotations often consist of unstructured, free-form text from non-experts or language models \citep{wang2024, liu2025}, which can introduce ambiguity and limit verifiable quantitative analysis. While datasets with structured Land Use/Land Cover (LULC) annotations exist (e.g., Sen2Lucas \citep{jain2024}), they lack crisis-related information and still rely on the optical domain. Second, a direct consequence of this data scarcity is the absence of retrieval models capable of jointly interpreting textual queries, multispectral optical data, and SAR imagery. No existing architecture creates a shared semantic space that unifies these three modalities, preventing the development of retrieval systems that can leverage the unique physical information captured by different sensors.

To bridge this gap, we present \textbf{CrisisLandMark}, a new, large-scale T2RSIR corpus of over 647,000 image-text pairs from the Sentinel-1 (SAR) and Sentinel-2 (optical) missions. Our corpus is enriched with structured, machine-readable annotations for LULC and crisis events, derived from authoritative sources like CORINE Land Cover \citep{buttner2017} and Dynamic World \citep{brown2022}. This provides an unambiguous foundation for training and evaluation. Its characteristics are summarized in \Cref{tab:corpora_comparison}. Additionally, to leverage this rich dataset, we propose \textbf{CLOSP} (\textbf{C}ontrastive \textbf{L}anguage \textbf{O}ptical \textbf{S}AR \textbf{P}retraining), a novel multimodal architecture designed to align textual descriptions, optical MSI, and SAR data into a shared latent space. By using the structured text annotations as a common anchor, CLOSP effectively learns to associate the different visual modalities without requiring temporally aligned image pairs. Furthermore, we introduce GeoCLOSP, which integrates geographic coordinates into CLOSP to explore the trade-off between semantic generality and geographic specificity.

Our experimental results on the new CrisisLandMark benchmark demonstrate that CLOSP substantially outperforms existing state-of-the-art T2RSIR models. We show that its unified training strategy enables powerful knowledge transfer, significantly improving the semantic retrieval of challenging SAR imagery. The findings underscore the value of integrating diverse sensor data to build more powerful, robust, and versatile retrieval systems for large-scale Earth observation.

In summary, the main contributions of this paper are:
\begin{itemize}
    \item A new, large-scale T2RSIR corpus, \textbf{CrisisLandMark}, featuring over 647,000 paired examples of multispectral optical (Sentinel-2) and SAR (Sentinel-1) imagery with structured annotations for land use, land cover, and crisis events.
    \item A novel multimodal learning architecture, \textbf{CLOSP}, that is the first to create a shared semantic space aligning textual descriptions, optical MSI, and SAR imagery, with an extension, GeoCLOSP, that incorporates geographic locations.
    \item A comprehensive empirical validation demonstrating CLOSP's superiority over state-of-the-art baselines and quantifying the trade-off between semantic retrieval and geographic specificity.
\end{itemize}

The code and corpus are publicly available at \url{https://github.com/DarthReca/closp}. The remainder of this paper is organized as follows: Section 2 details the construction of the CrisisLandMark corpus. Section 3 presents the CLOSP and GeoCLOSP architectures. Section 4 describes the experimental setup and results, Section 5 discusses the findings, and Section 6 concludes the paper, presenting future works.

\begin{table}[pos=t]
    \centering
    \caption{Comparison with the existing T2RSIR datasets. \textit{LC}, \textit{MSI/SAR}, and \textit{Crisis} respectively indicate if the dataset contains annotations about land cover, multispectral or radiometric data, and crisis-related information.}
    \label{tab:corpora_comparison}
        \begin{tabular}{c|cccc} 
        \toprule
        Corpus                        & LC & MSI/SAR & Crisis & Samples (\#)   \\ 
        \midrule
        RSICD   \citep{lu2017}                       & \xmark   & \xmark    & \xmark       & 11k            \\
        RSITMD \citep{zhiqiang2022}                        & \xmark   & \xmark    & \xmark       & 5k             \\
        UCM-Captions   \citep{qu2016}             & \xmark   & \xmark    & \xmark        & 2k             \\
        SkyScript-Retrieval \citep{wang2024}           & \xmark   & \xmark    & \xmark        & 30k            \\
        Sen2Lucas \citep{jain2024}                    & \checkmark   & \xmark    & \xmark       & 235k           \\
        \textbf{CrisisLandMark (Ours)} & \checkmark   & \checkmark    & \checkmark       & \textbf{647k}  \\
        \bottomrule
        \end{tabular}
\end{table}

\section{Corpus}
\label{sec:corpus}
We present CrisisLandMark, a new corpus composed of Sentinel-1 and  Sentinel-2 data enriched with textual and geospatial annotations. Hereafter, we will describe the dataset composition and multimodal sources. 

\subsection{Data sources}
We consider the following datasets as satellite image sources: CaBuAr \citep{cambrin2023}, MMFlood \citep{montello2022}, Sen12Flood \citep{rambour2020}, QuakeSet \citep{cambrin2024}, and re-BEN~\citep{clasen2024}.

We align the spatial resolution of these datasets to 10 meters using interpolation and split the images into $120\times120$ patches to be consistent with the re-BEN dataset. 

We enrich the source images with a comprehensive set of textual and geospatial annotations.
To support crisis management applications, we incorporated the four specialized datasets (CaBuAr, MMFlood, Sen12Flood, Quakeset), retaining their original event tags (wildfire, flood, earthquake). For these crisis images, we generated corresponding LULC annotations (trees, bare, built, crops, grass, water, snow and ice, shrub and scrub, flooded vegetation) by querying the Dynamic World collection. For general LULC coverage in re-BEN, we mapped the high-quality CORINE LC provided with the dataset as detailed in \Cref{sec:source_text}. Spatial coordinates also characterize all the image patches of every dataset. 

The resulting corpus is primarily composed of general land cover scenes from re-BEN, which constitute 88\% of the images, providing a robust foundation for learning diverse visual semantics. The remaining 12\% from the specialized datasets offer critical examples for crisis retrieval tasks. \Cref{tab:corpus} provides a detailed breakdown of the data sources and their respective contributions of Sentinel-1 and Sentinel-2 imagery.

\begin{table}[pos=htb]
    \centering
        \caption{Corpus composition. We report the data source, the crisis event data, and the number of samples from each dataset, separately for Sentinel-1 (S1) and Sentinel-2 (S2) data.}
    \label{tab:corpus}
    \begin{tabular}{@{}lcccc@{}}
\toprule
Dataset                       & Task           & S1 (\#)    & S2 (\#)     & Crisis event \\ \midrule
re-BEN      & Classification & 286159 & 286214 & \xmark       \\
CaBuAr     & Segmentation   & \xmark & 3272   & wildfire     \\
QuakeSet   & Classification & 21430  & \xmark & earthquake   \\
MMFlood   & Segmentation   & 27880  & \xmark & flooding     \\
Sen12Flood & Classification & 2873   & 18975  & flooding     \\ \midrule
                              &                & 338342 & 308461 &              \\ \bottomrule
\end{tabular}

\end{table}

\subsection{Source Images}
The corpus is composed of Sentinel-1 GRD and Sentinel-2 L2A products. Three source datasets contain Sentinel-2 data (re-BEN, CaBuAr, and Sen12Flood), whereas four of them include Sentinel-1 data (re-BEN, QuakeSet, MMFlood, Sen12Flood). Specifically, Sentinel-2 L2A (\underline{L}evel \underline{2A} Surface Reflectance) products cover 12 spectral bands in ultra-blue, visible, near-infrared, and short-wave infrared. In contrast, Sentinel-1 GRD (\underline{G}round \underline{R}ange \underline{D}etected) products have two different polarizations, namely \underline{V}ertical transmit-\underline{V}ertical receive (VV) and \underline{V}ertical transmit-\underline{H}orizontal receive (VH).

\begin{table}[pos=htb]
    \caption{Channels of Sentinel-2 L2A. \textit{UB} is Ultra-Blue, \textit{VNIR} is Visible and Near Infrared, \textit{SWIR} is ShortWave Infrared.}
    \label{tab:bands}
    \centering
    \begin{tabular}{l|c|ll}
    \toprule
    Band & Central wavelength & Description & Applications \\
    \midrule
    B1 &	443 nm	& UB   & Coastal, Aerosol \\
    B2 &	490 nm	& Blue & Vegetation, Water \\
    B3 &	560 nm	& Green& Plant health \\
    B4 &	665 nm	& Red  & Vegetation, Soil \\
    B5 &	705 nm	& VNIR & Vegetation, Biomass \\
    B6 &	740 nm	& VNIR & Vegetation Stress \\
    B7 &	783 nm	& VNIR & Chlorophyll, Canopy\\
    B8 &	842 nm	& VNIR & Vegetation Health \\
    B8a &	865 nm	& VNIR & Vegetation \\
    B9 &	940 nm	& SWIR & Water vapor \\
    B11 &	1610 nm	& SWIR & Soil, Snow, Clouds \\
    B12 &	2190 nm	& SWIR & Mineralogy \\
    \bottomrule
\end{tabular}

\end{table}

\begin{figure}[pos=htb]
    \centering
    \subfloat[Sentinel-1 VV\label{fig:example_general_S1_1}]{\includegraphics[width=0.3\linewidth]{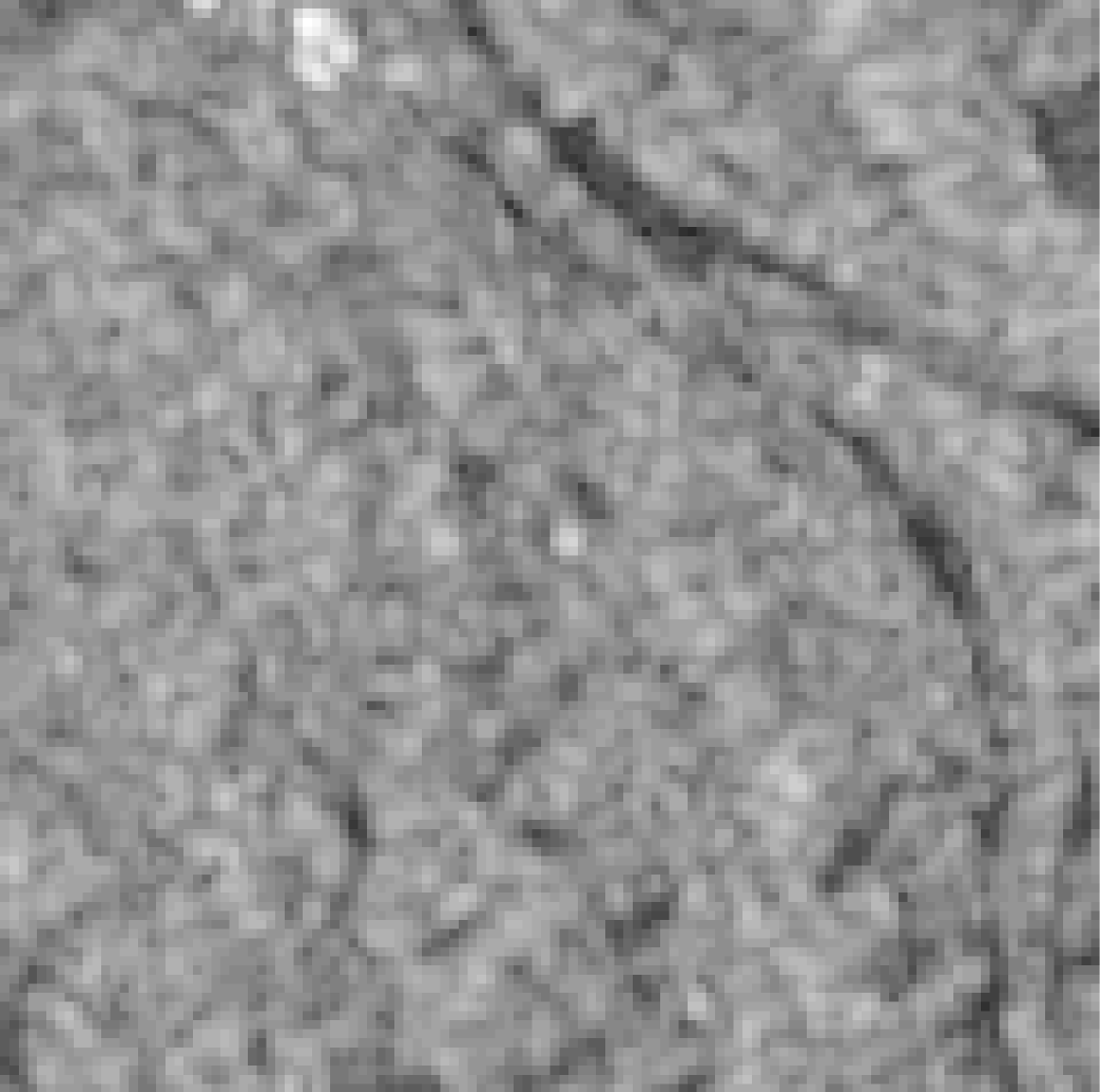}}
    \hfil 
    \subfloat[Sentinel-1 VH\label{fig:example_general_S1_2}]{\includegraphics[width=0.3\linewidth]{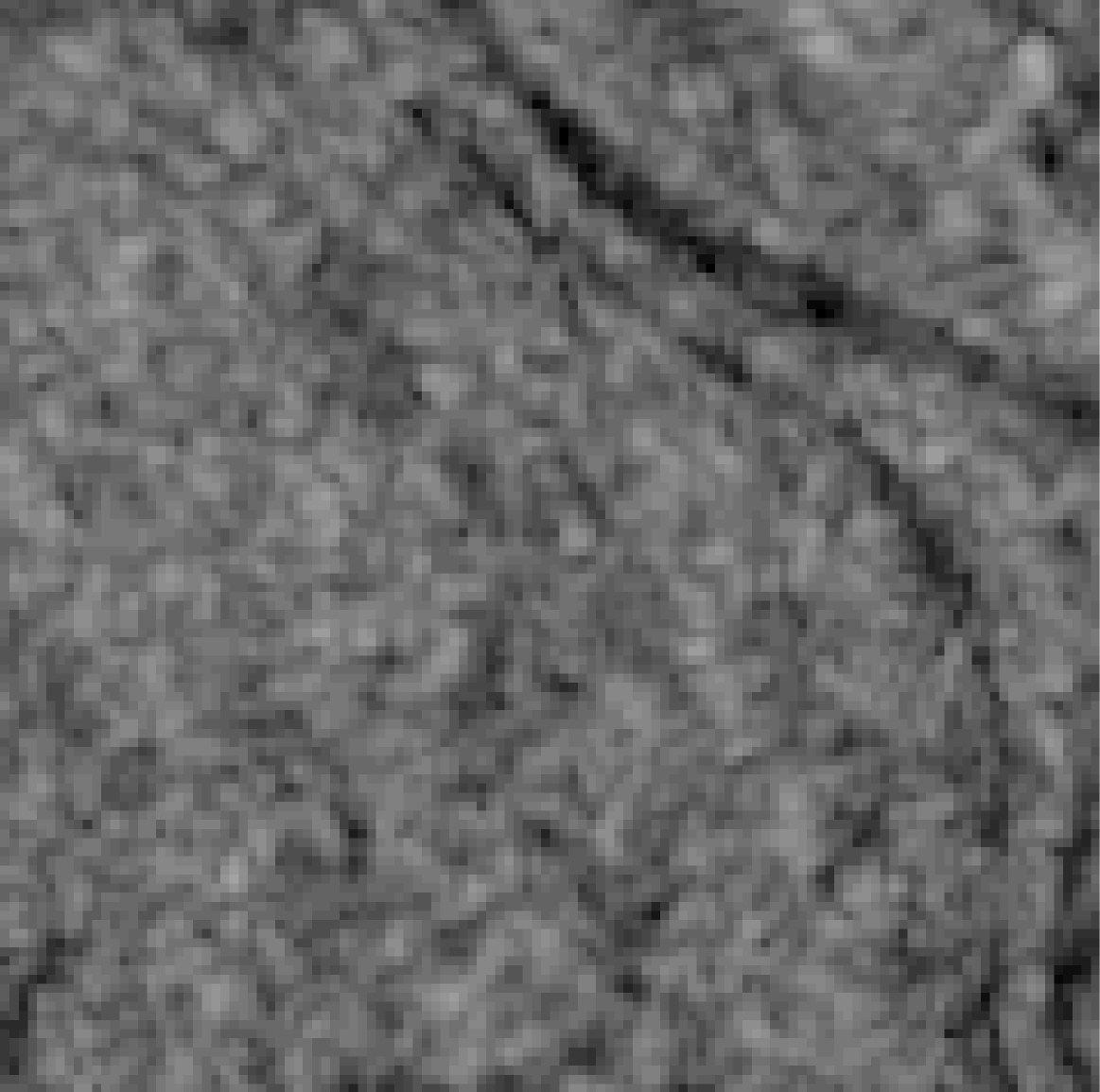}}
    \hfil
    \subfloat[Sentinel-2 RGB\label{fig:example_general_S2}]{\includegraphics[width=0.3\linewidth]{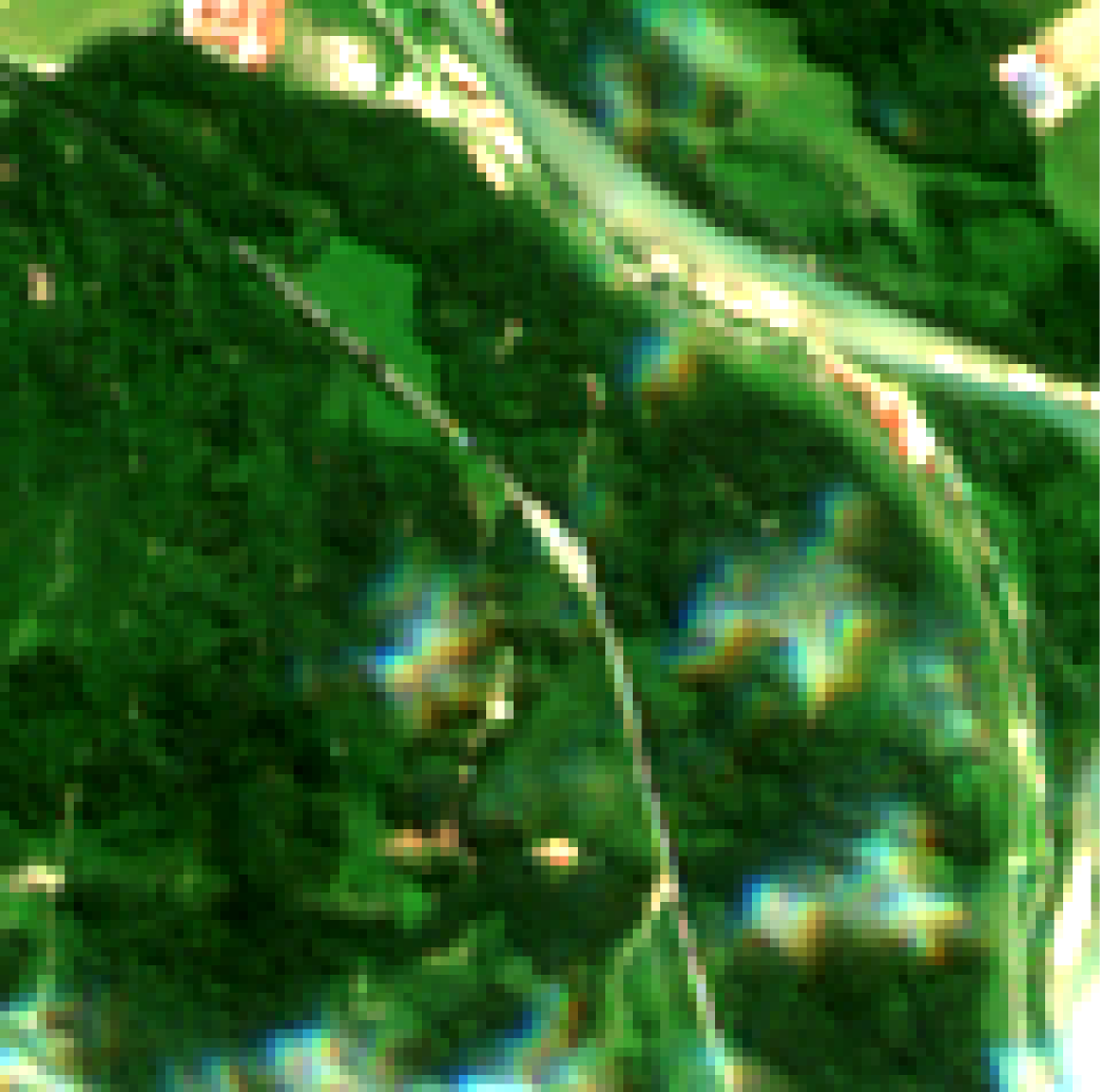}}
    \caption{Sample images from Sentinel-1 VV, and Sentinel-1 VH, and Sentinel-2 RGB of the same area. The scale of Sentinel-1 image values for each channel is the same.}
    \label{fig:example_general}
\end{figure}

\paragraph{Sentinel-1 GRD}
Sentinel-1 GRD products \citep{s1Processing,torres2012} are satellite images captured using radar technology that provides detailed, day-and-night observations of the Earth’s surface. Unlike regular cameras, radar can see through clouds and in darkness, making it ideal for consistent monitoring in all weather conditions. GRD products represent the Earth in two dimensions, with each pixel corresponding to a specific area on the ground. The dataset contains two channels: VV (Vertical transmit, Vertical receive) and VH (Vertical transmit, Horizontal receive), which refer to the orientation of the radar waves when they are sent out and received back. The VV polarization is more sensitive to smooth surfaces, while VH is more adequate for detecting rough ones. An example of a single channel of one of these products can be seen in \Cref{fig:example_general_S1_1,fig:example_general_S1_2}.

\begin{figure}[pos=!htb]
    \centering
    \subfloat{\includegraphics[width=0.3\linewidth]{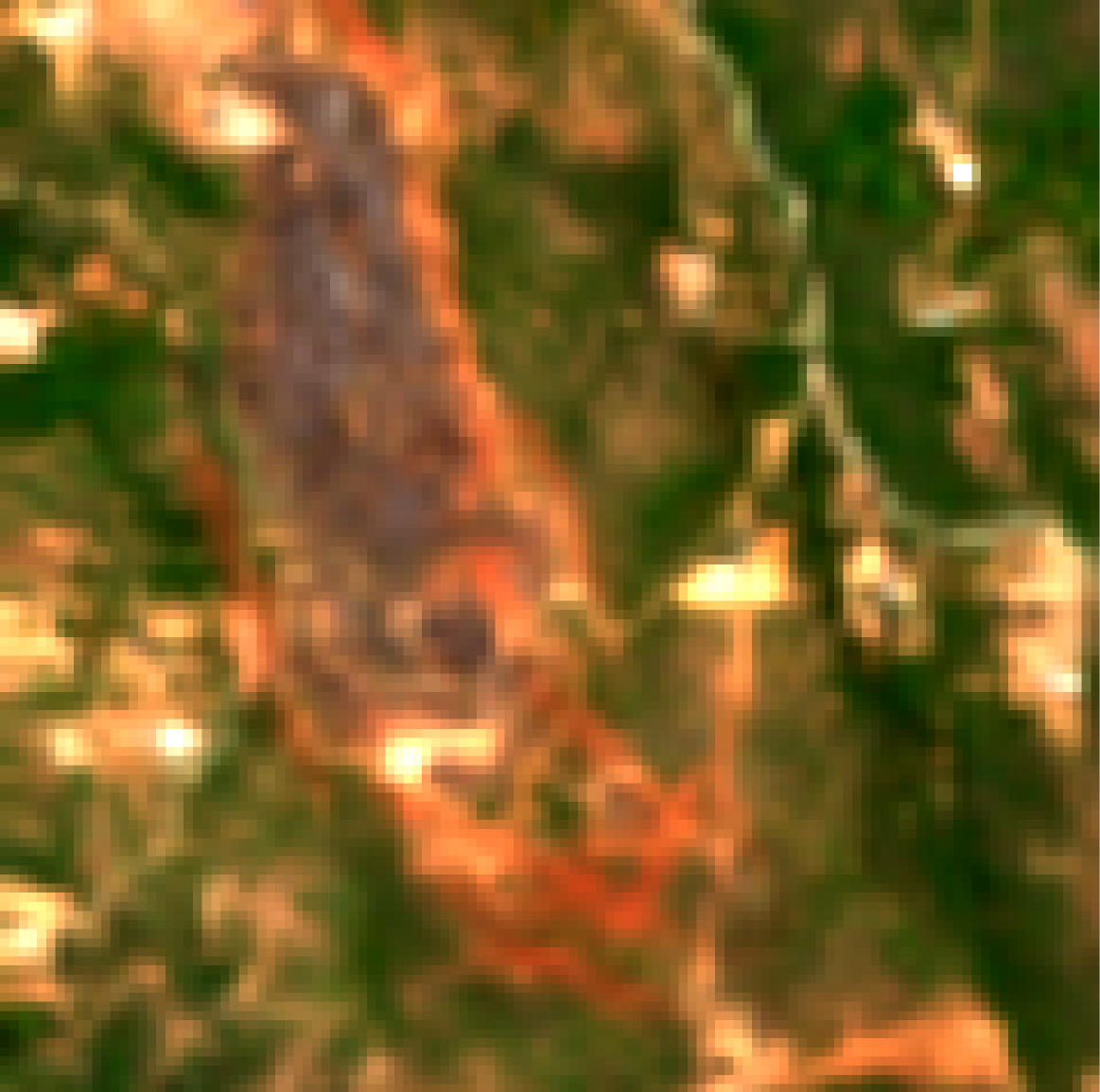}} \\\vspace{1cm}
    \subfloat{\includegraphics[width=0.2\linewidth]{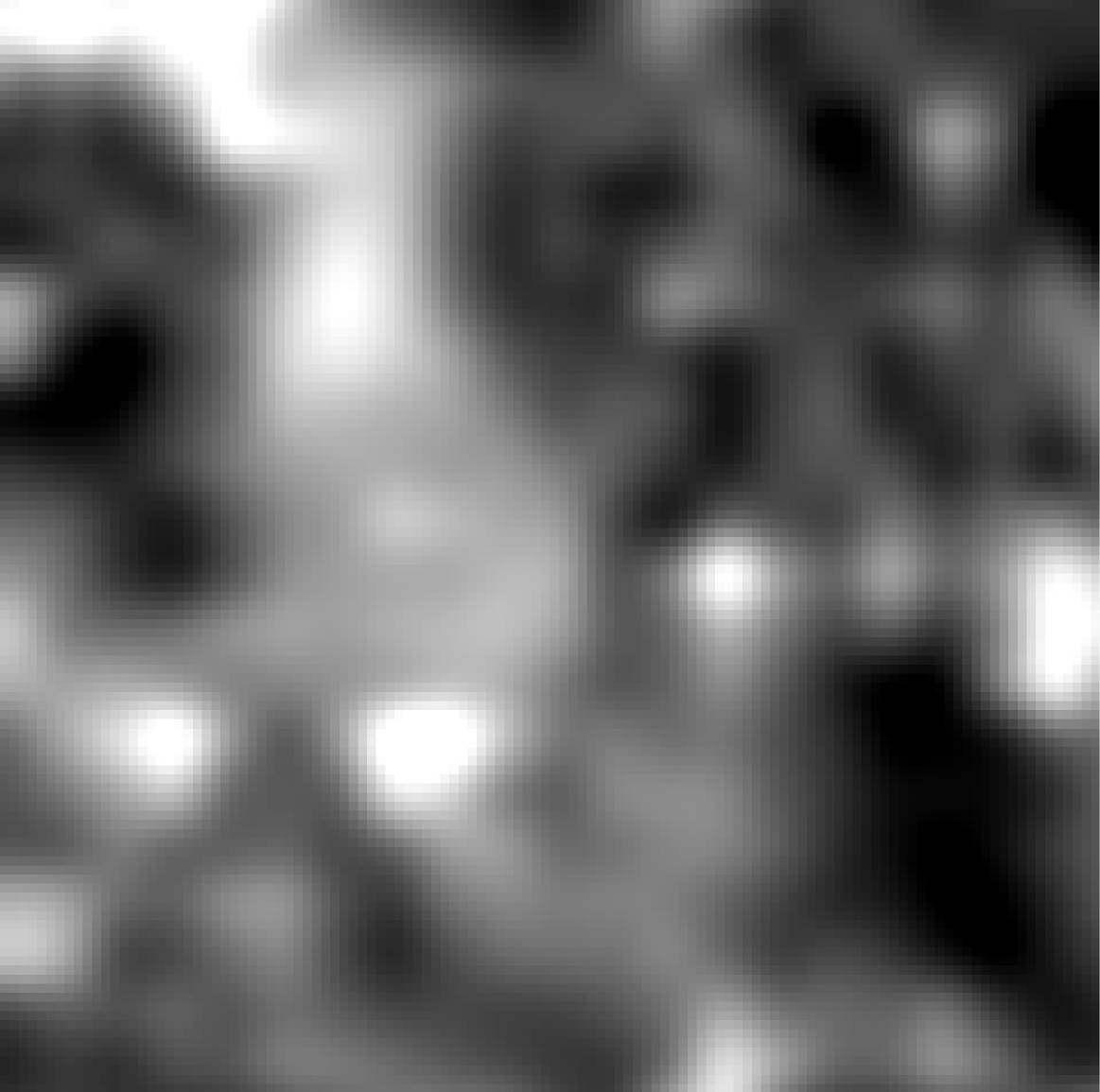}} \hfil
    \subfloat{\includegraphics[width=0.2\linewidth]{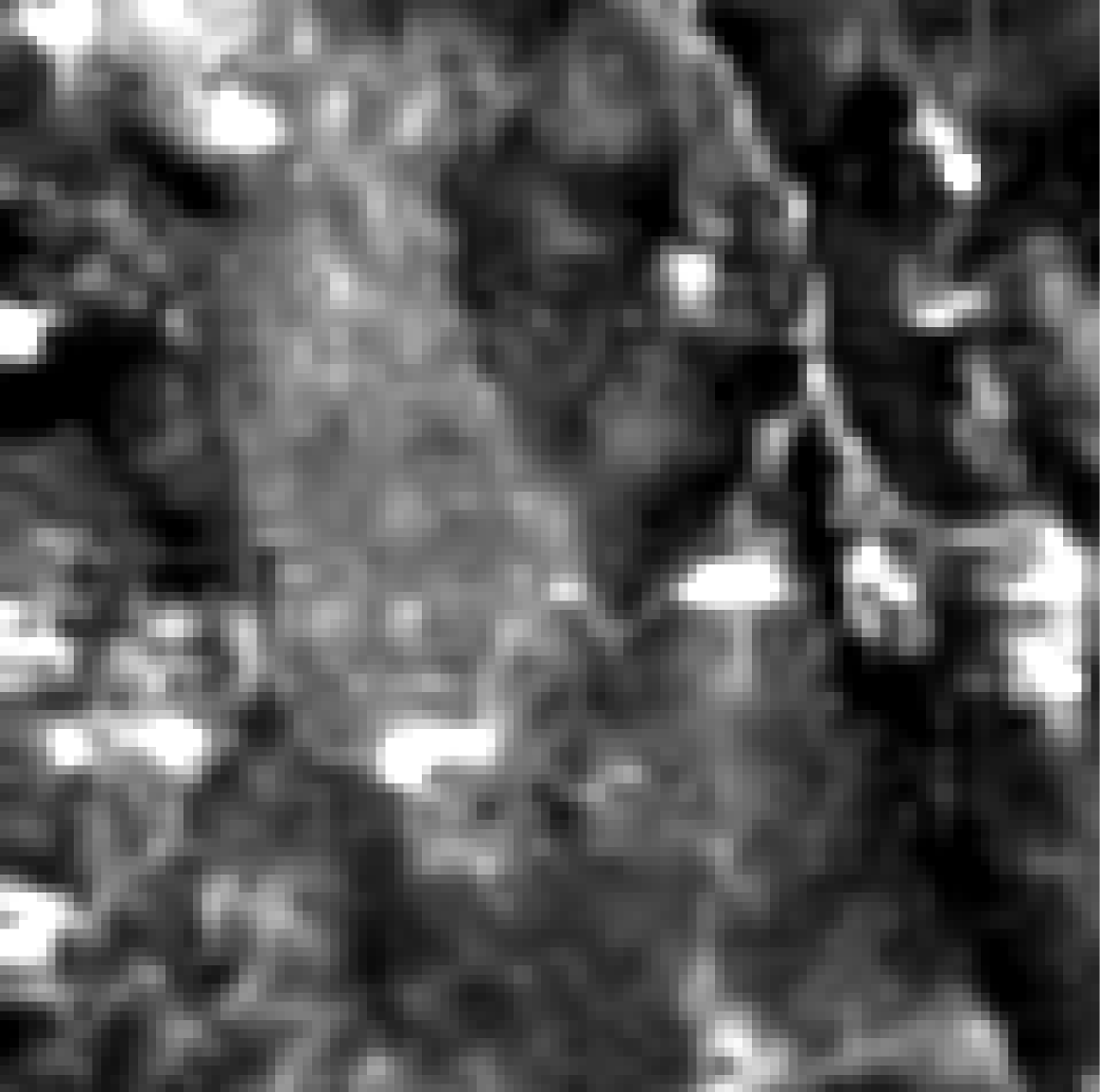}} \hfil
    \subfloat{\includegraphics[width=0.2\linewidth]{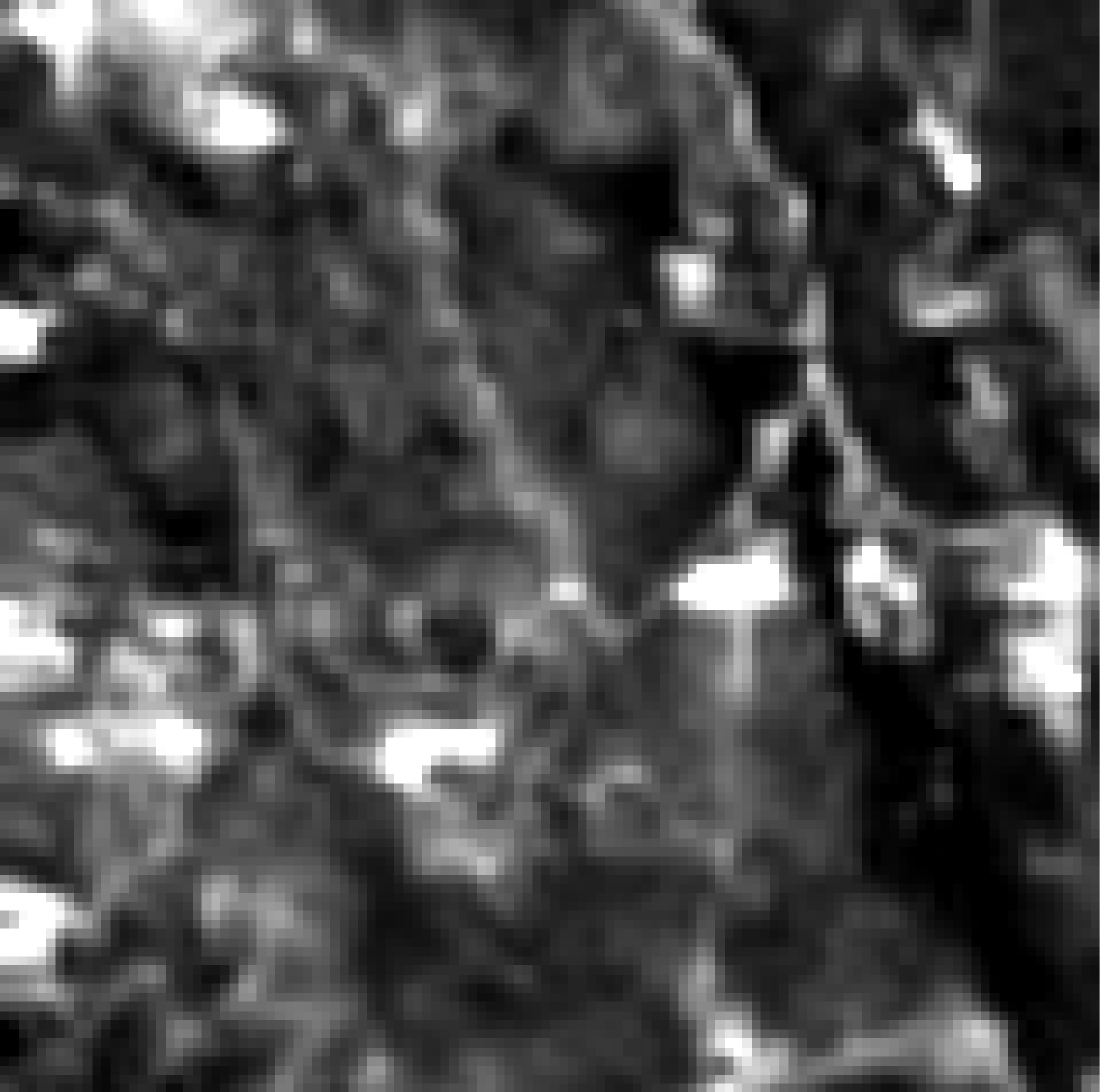}} \hfil
    \subfloat{\includegraphics[width=0.2\linewidth]{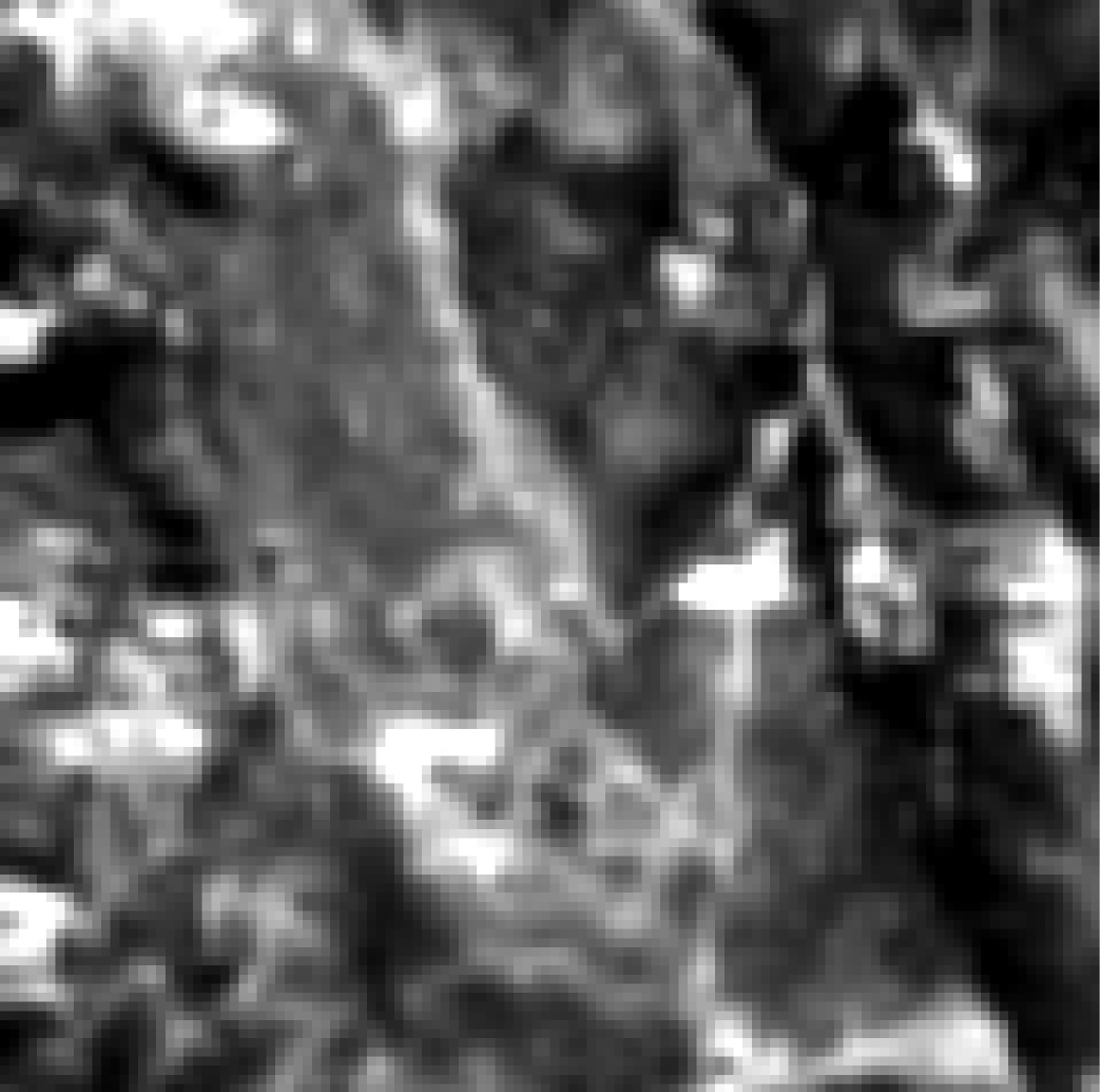}} \\
    \subfloat{\includegraphics[width=0.2\linewidth]{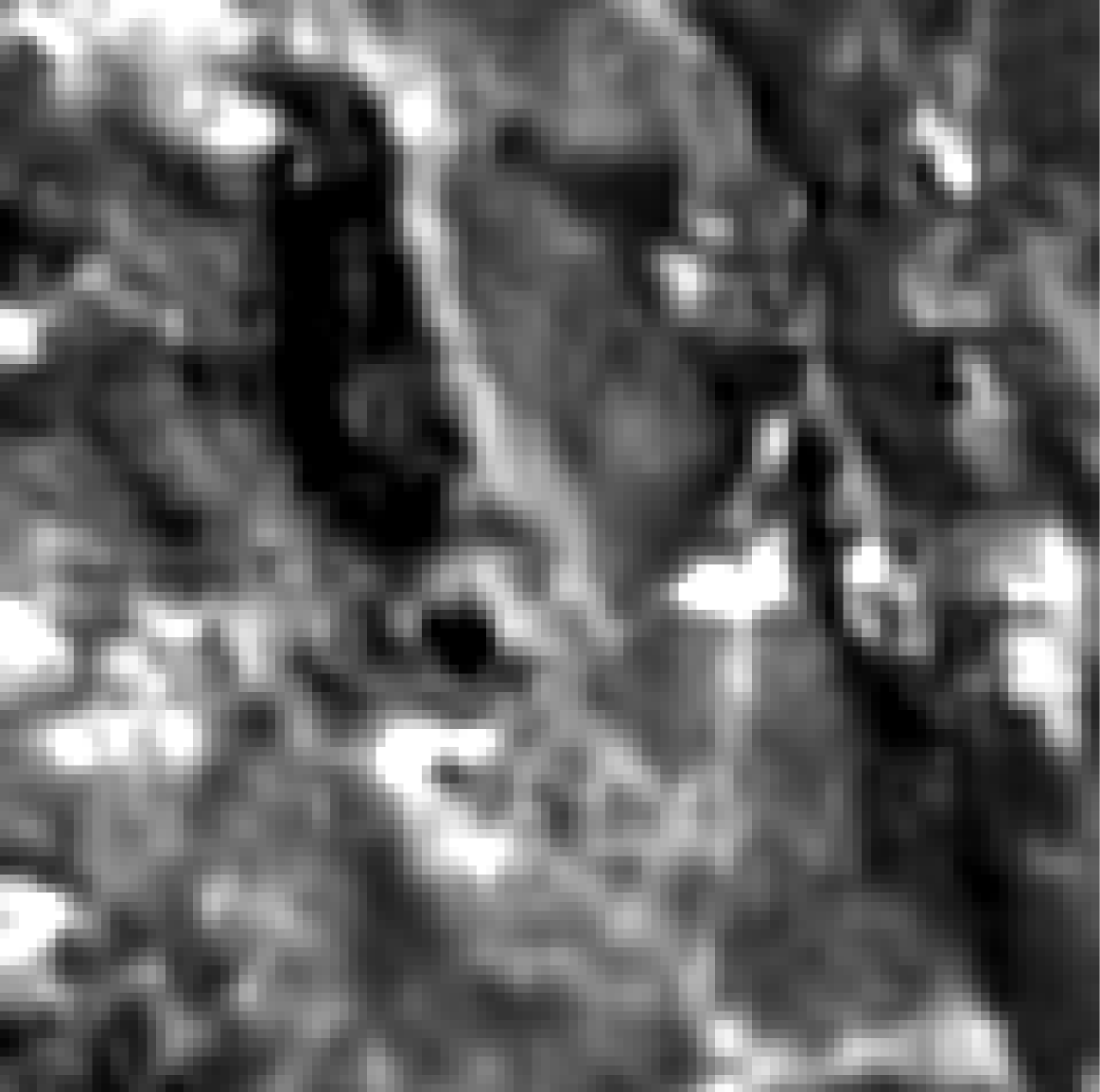}} \hfil
    \subfloat{\includegraphics[width=0.2\linewidth]{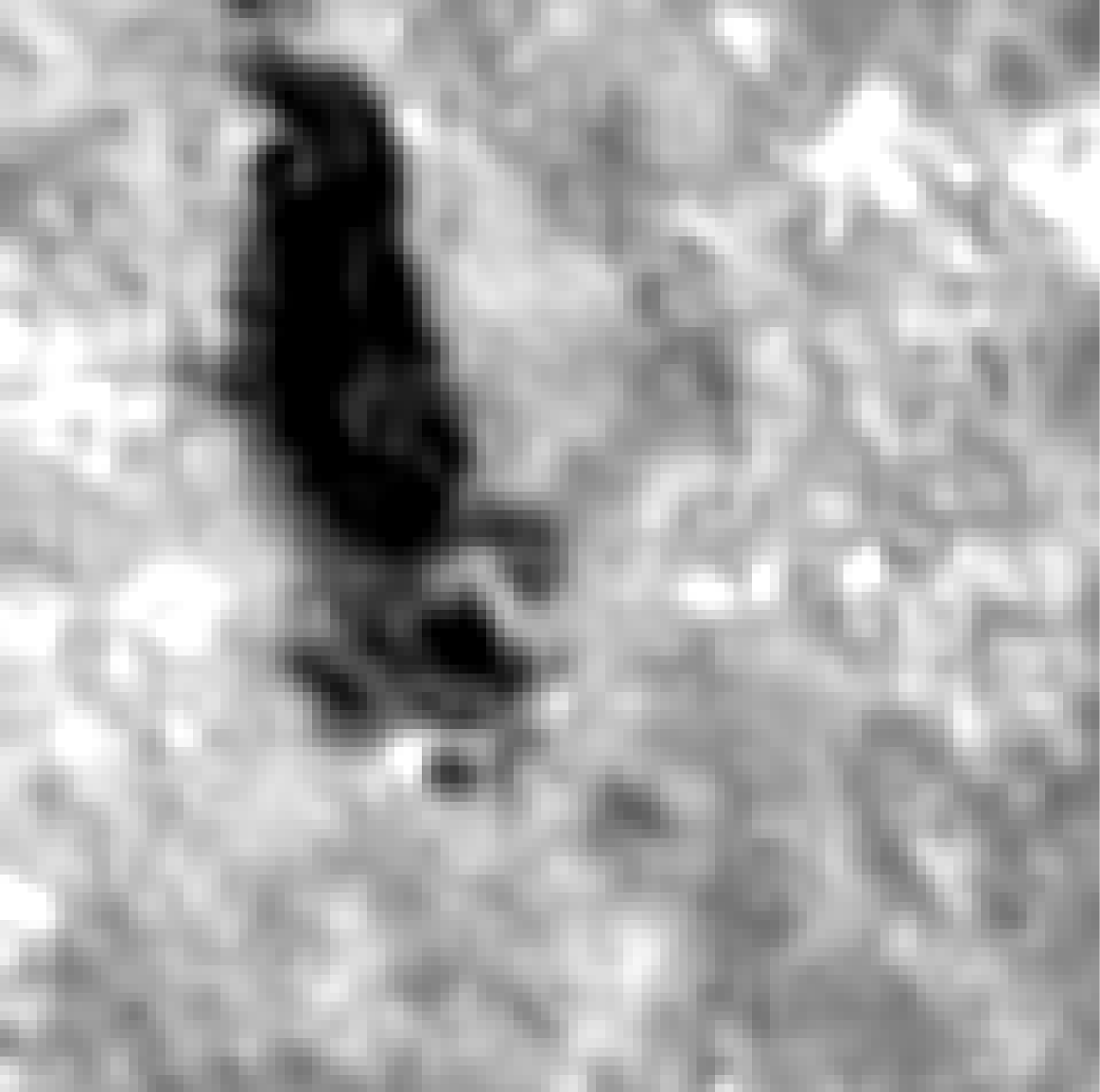}} \hfil
    \subfloat{\includegraphics[width=0.2\linewidth]{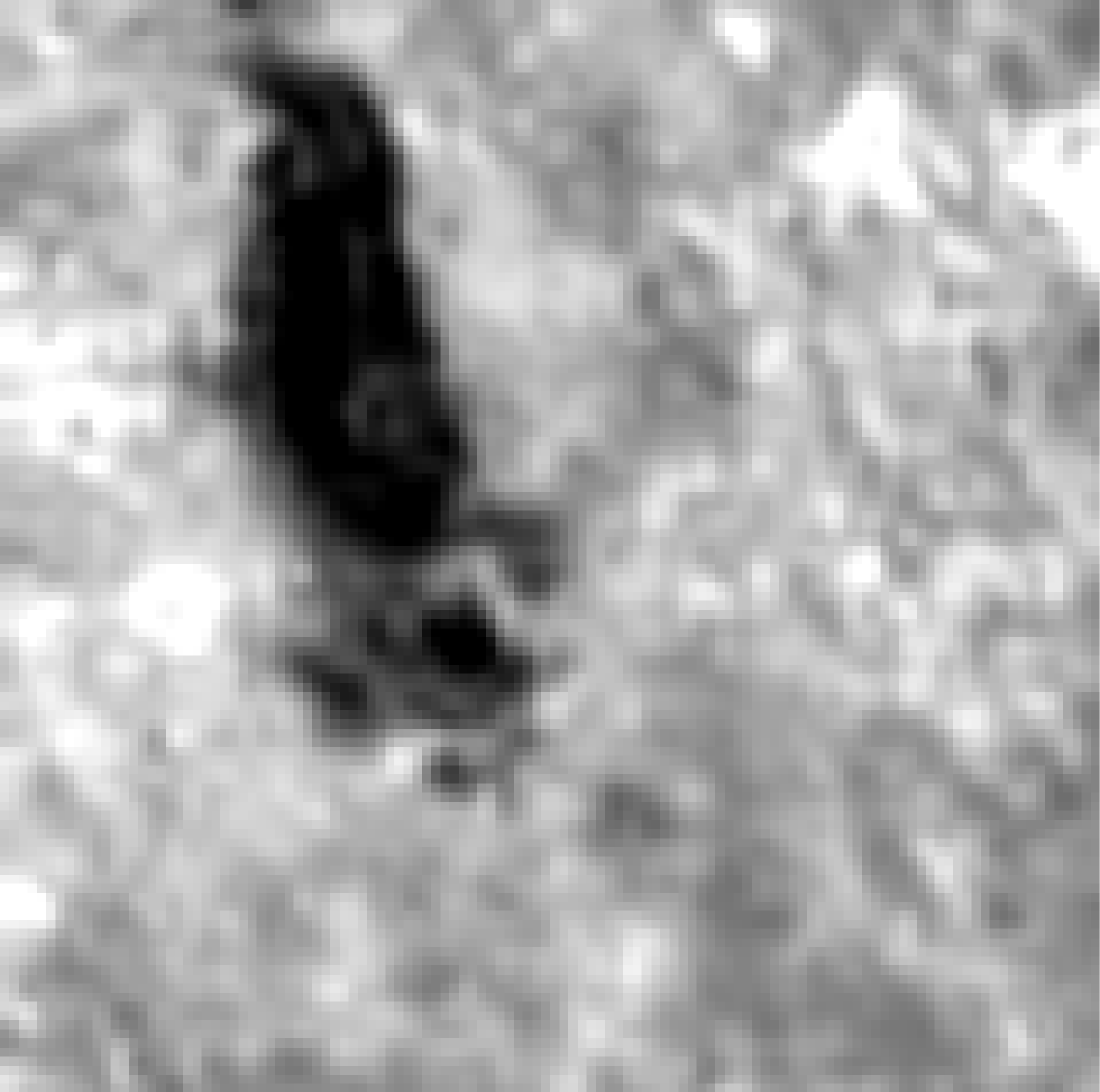}} \hfil
    \subfloat{\includegraphics[width=0.2\linewidth]{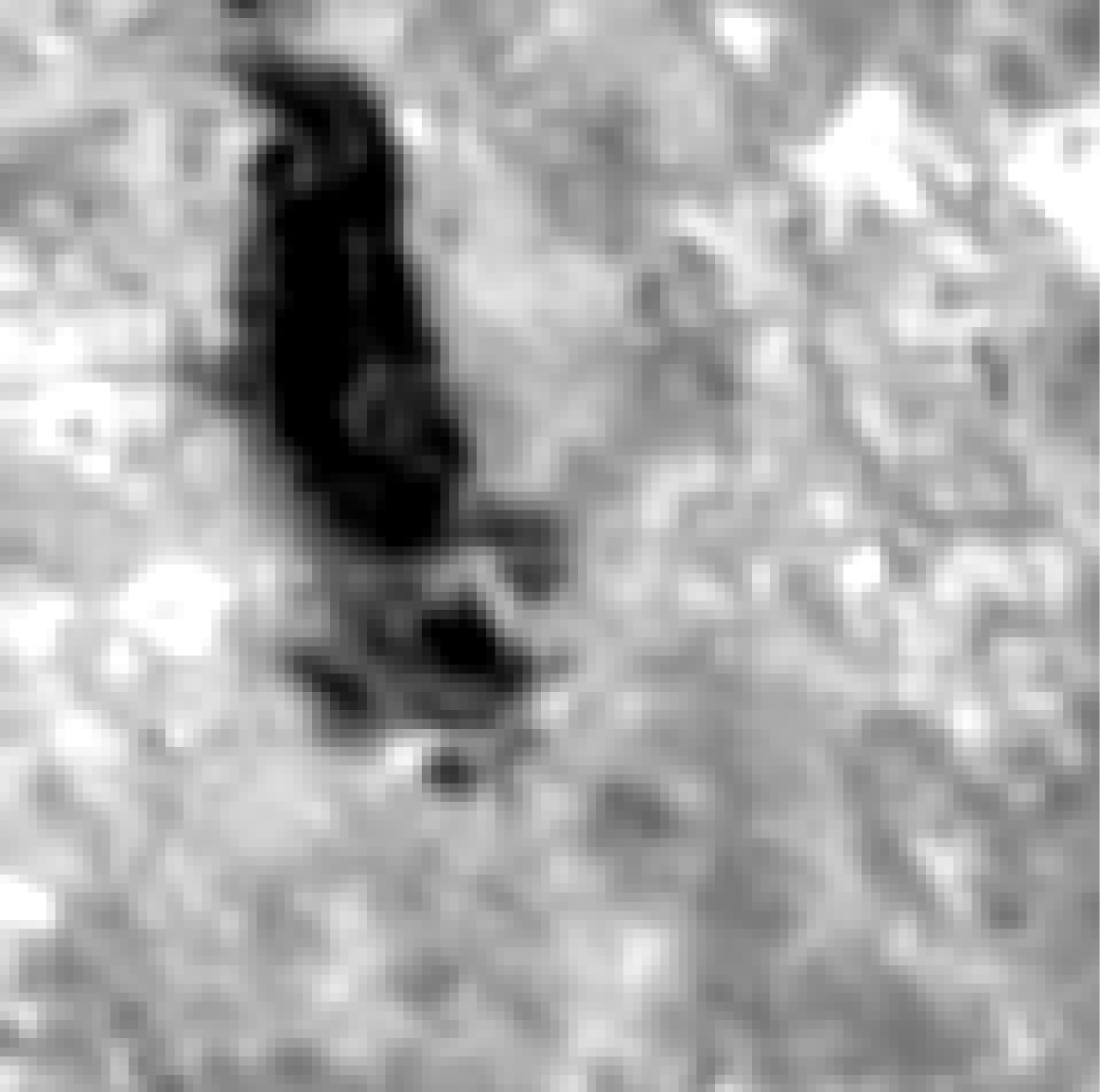}} \\
    \subfloat{\includegraphics[width=0.2\linewidth]{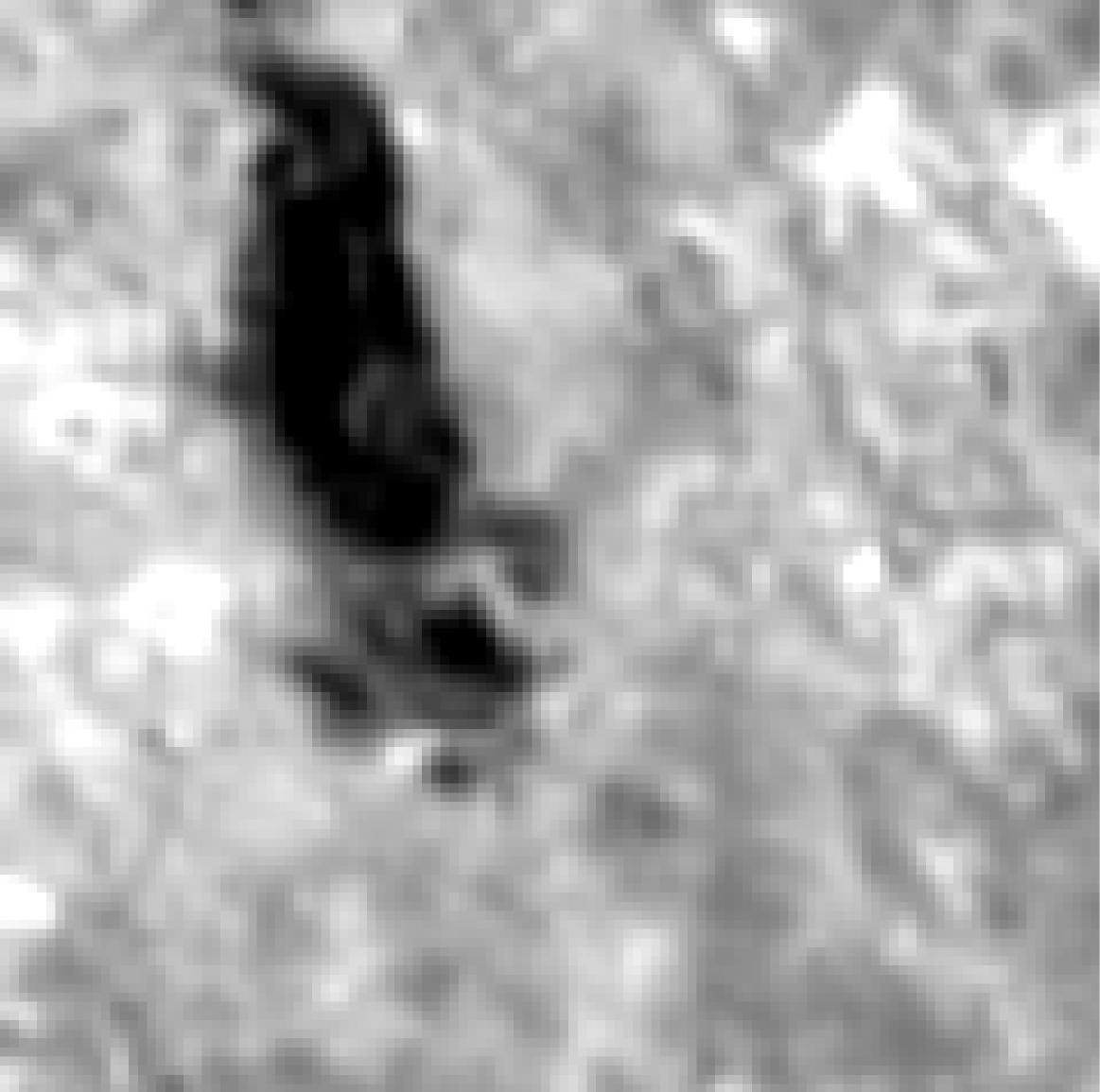}} \hfil
    \subfloat{\includegraphics[width=0.2\linewidth]{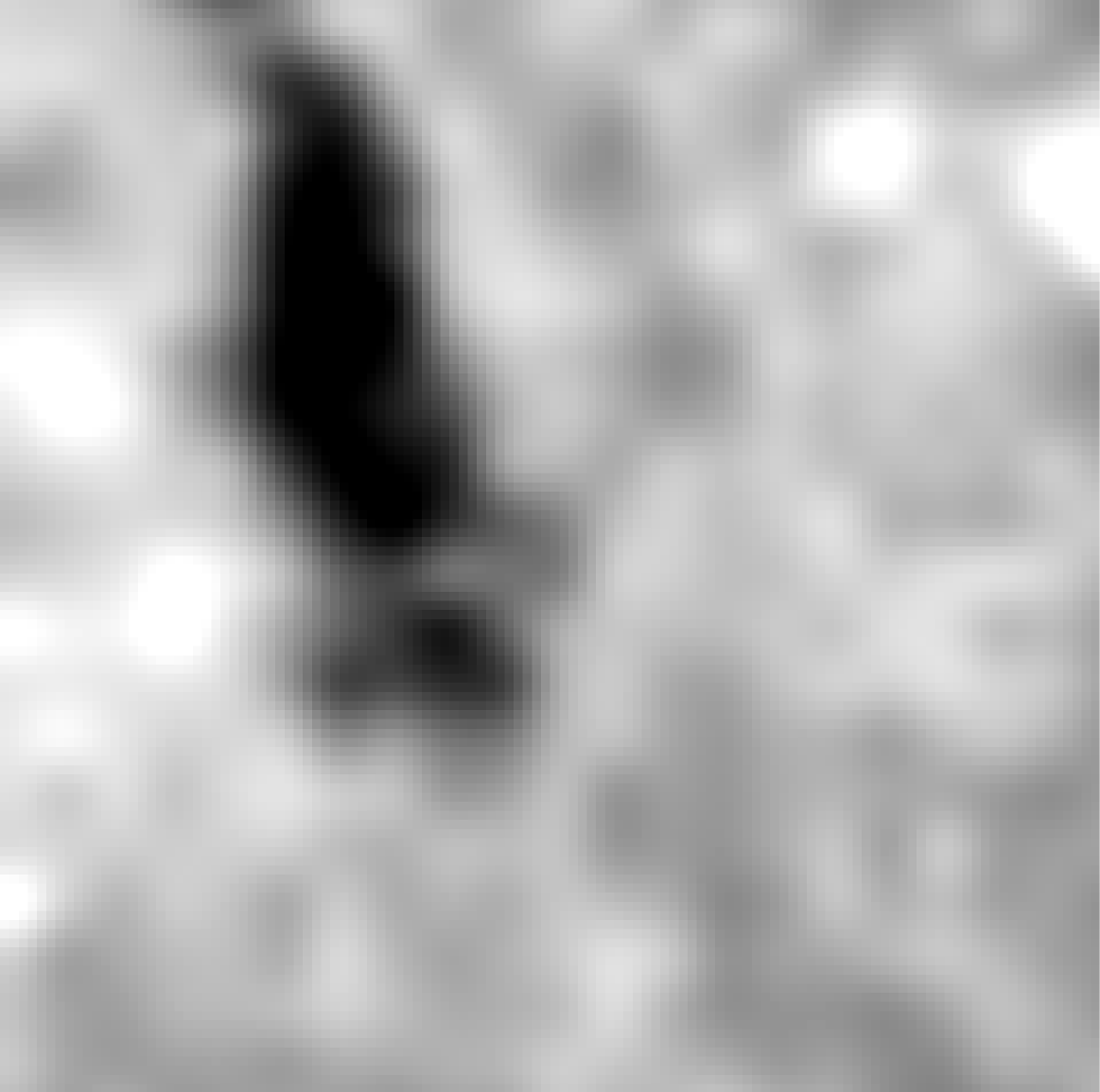}} \hfil
    \subfloat{\includegraphics[width=0.2\linewidth]{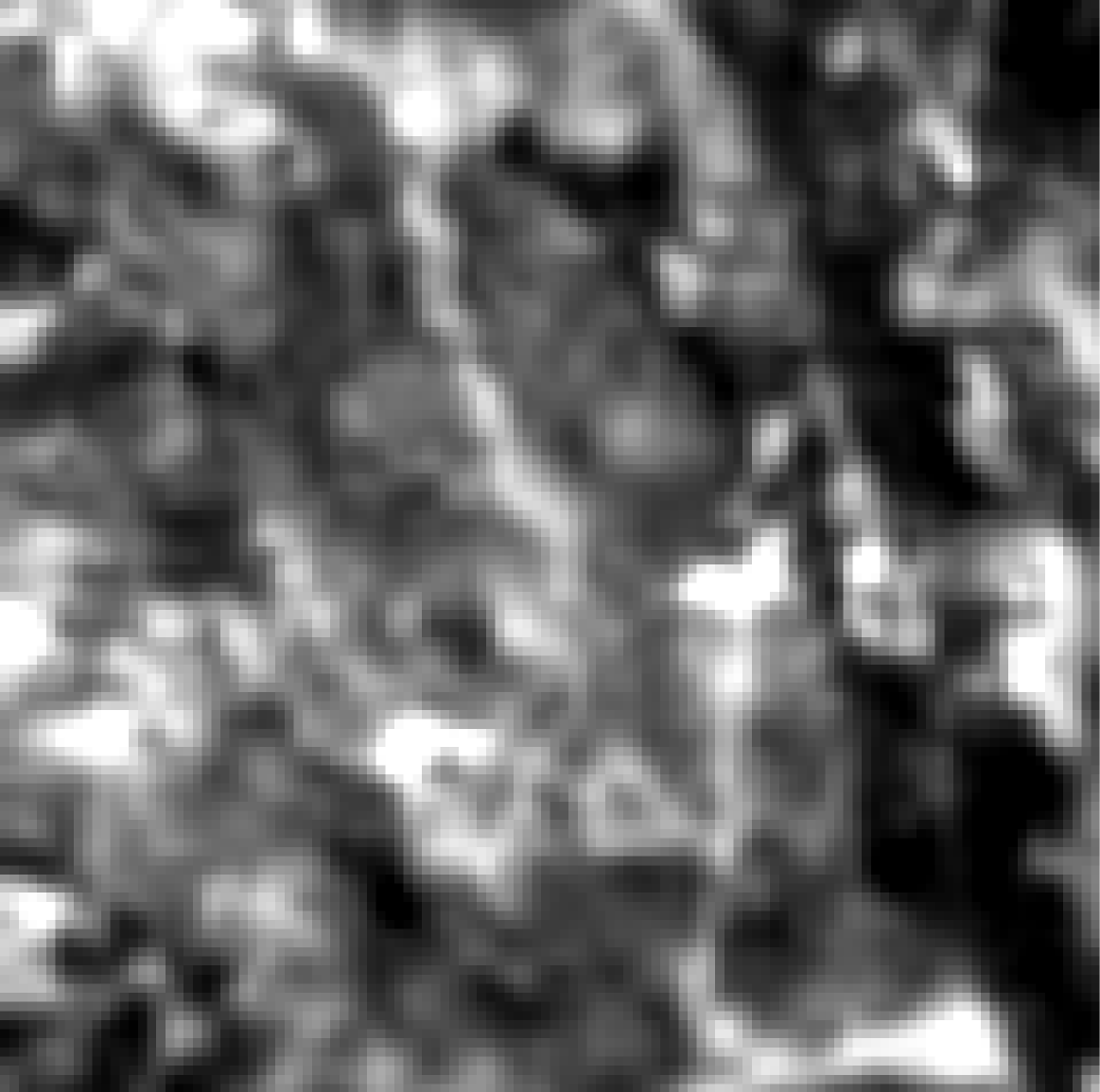}} \hfil
    \subfloat{\includegraphics[width=0.2\linewidth]{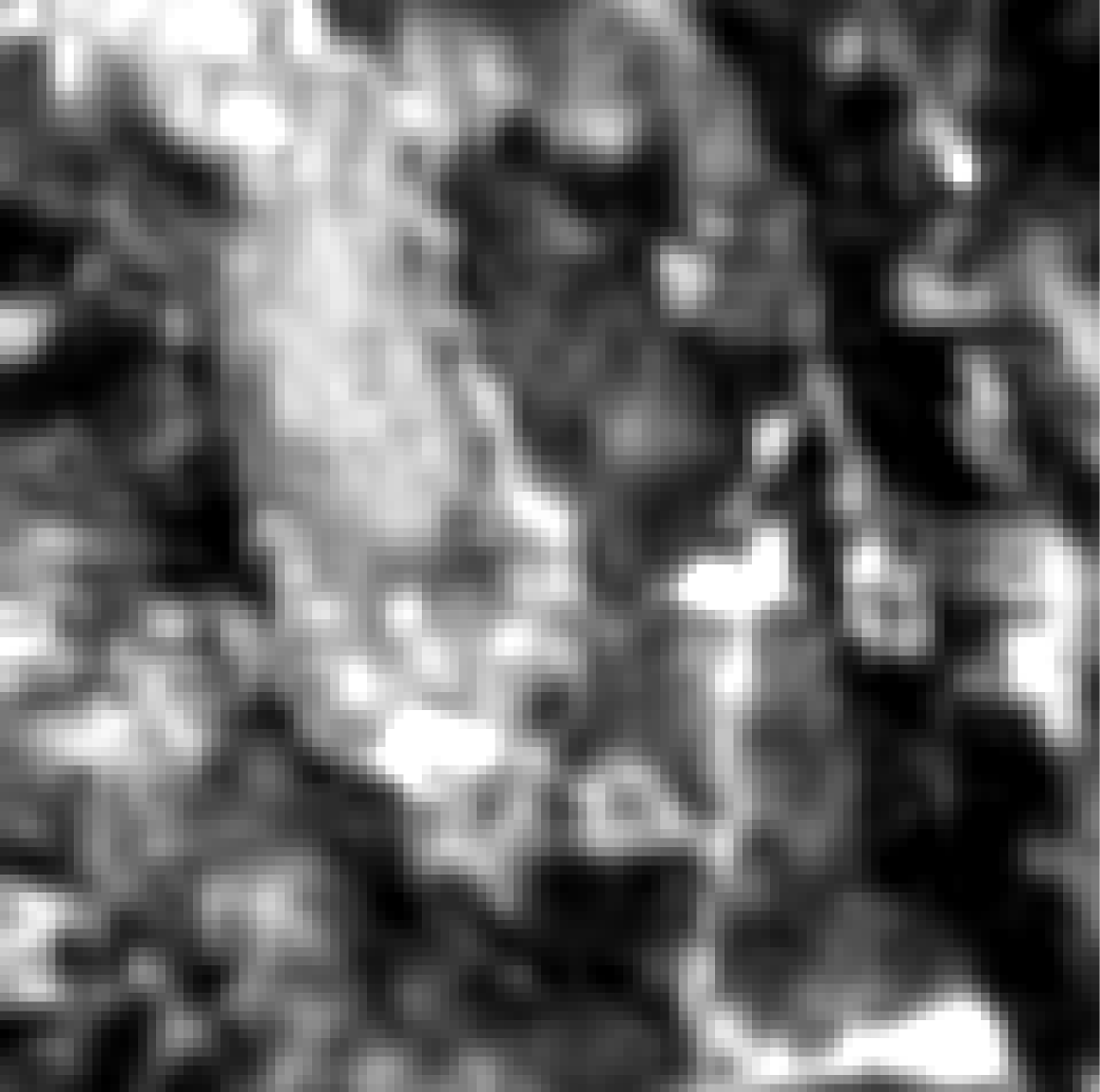}} \\
    \caption{Sample image of a wildfire seen through RGB and all the bands of Sentinel-2. Some bands are at a lower resolution.}
    \label{fig:example_wildfire}
\end{figure}

\paragraph{Sentinel-2 L2A}
Sentinel-2 Level-2A (L2A) products \citep{drusch2012,s2Processing} are high-resolution satellite images that provide detailed, true-color views of the Earth's surface, optimized for environmental and land monitoring. These images are processed to correct atmospheric effects like haze or cloud cover, making them more accurate and ready for analysis. They include 13 spectral bands, ranging from visible light (what we see) to infrared, enabling insights into vegetation health, water quality, soil conditions, and urban development. After the atmospheric correction, one band was removed, resulting in 12 bands, as seen in \Cref{tab:bands}. One example image of the RGB bands can be seen in \Cref{fig:example_general_S2} paired with VV and VH channels (\Cref{fig:example_general_S1_1,fig:example_general_S1_2}) of a Sentinel-1 GRD product of the same area. The different vision achieved with different spectral bands is shown in \Cref{fig:example_wildfire} in the case of a wildfire.

\subsection{Source text}
\label{sec:source_text}
We design the dataset to support
standard land use and land cover applications rather than relying on unstructured annotations. 
This is motivated by the fact that LULC applications are typically 
designed to address the need of institutions for planning regulations and
to assess the evolution of certain specific areas regarding ecological aspects and human activities.

The re-BEN dataset annotations are based on 43 classes, grouped into categories, provided by \underline{C}ORINE \underline{LC} (CLC)~\citep{buttner2017} at its finer level of detail.
CLC is a European Environment Agency (EEA) program that provides consistent, detailed information on land use and land cover across Europe. It is updated periodically (i.e., 1990, 2000, 2006, 2012, 2018) and helps track land use changes over time, such as urban expansion, deforestation, or changes in agricultural practices. CLC uses a consistent hierarchical classification system across Europe, ensuring data comparability between countries with a minimum mapping unit of 25 hectares. Data are primarily gathered using satellite imagery but are cross-checked with ground observations and other sources to ensure accuracy. However, CLC is geographically limited to Europe and updated too infrequently for global crisis analysis.

Since CORINE is limited to specific years and only available for Europe, we enrich the four crisis-event datasets (that cover different years and areas than CORINE) with the 9 LULC classes of \underline{D}ynamic \underline{W}orld (DW)~\citep{brown2022} (i.e., Water, Trees, Grass, Flooded Vegetation, Crops, Shrub and Scrub, Built area, Bare Ground, Snow), which are contained in CORINE categories. The crisis annotation is taken from the original dataset.
Dynamic World is a high-resolution, near-real-time global land cover system developed by Google, leveraging deep learning models and Sentinel-2 satellite imagery. It provides 10-meter spatial resolution land cover classifications of the entire globe, making it applicable to diverse ecosystems and regions. Unlike traditional datasets, it offers frequent updates, enabling the detection of rapid and fine-scale land cover changes. The crisis datasets also contain special keywords to denote \textit{burned area}, \textit{earthquake damaged}, and \textit{flooded area}.
To create a unified and scalable label space, we harmonized all annotations using the 9-class system from DW. Thanks to its near-real-time global coverage, it is ideal for analyzing diverse ecosystems and detecting rapid land cover changes. With this solution, we map the higher quality of CLC annotation to the same classes of DW. The description and distribution of the classes can be found in \Cref{tab:classes_description}. The mapping is detailed in \Cref{sec:clc_mapping}.

\begin{table}[pos=!htb]
    \centering
    \caption{The harmonized classes of the CrisisLandMark corpus, their descriptions, and the percentage of images in which each class appears.}
    \label{tab:classes_description}
        \begin{tabular}{p{3.5cm} p{7.5cm} r}
        \toprule
        \textbf{Label} & \textbf{Description} & \textbf{Images (\%)} \\
        \midrule
        \multicolumn{3}{l}{\textit{Dynamic World Land Cover Classes}} \\
        \midrule
        Trees & Any significant cover of trees. & 69.00 \\
        Crops & Land cultivated for agriculture. & 57.28 \\
        Shrub and Scrub & Areas dominated by shrubs or low, woody vegetation. & 35.80 \\
        Water & Open and permanent water bodies. & 28.92 \\
        Grass & Land covered predominantly by grasses and other non-woody vegetation. & 27.18 \\
        Built & Artificial, man-made surfaces and structures. & 18.65 \\
        Flooded Vegetation & Areas where vegetation (e.g., forests, crops) is temporarily inundated with water. & 7.62 \\
        Bare & Land with little to no vegetation cover. & 6.95 \\
        Snow and Ice & Surfaces permanently or seasonally covered by snow or ice. & 2.38 \\
        \midrule
        \multicolumn{3}{l}{\textit{Crisis Event Classes}} \\
        \midrule
        Flooded Area & Land temporarily submerged by water due to a flood event. & 7.69 \\
        Earthquake Damage & Visible structural damage to built areas or significant land deformation caused by seismic activity. & 3.31 \\
        Burned Area & Land showing evidence of recent fire, characterized by burn scars and the destruction of vegetation. & 0.54 \\
        \bottomrule
        \end{tabular}
\end{table}

\paragraph{Data splits}
We divide the resulting dataset into a training set (20\% of the total) for training our model and a corpus for retrieval (80\% of the total). To ensure the same label distribution, we employ a stratified multi-label sampling~\citep{sechidis2011}. According to the $\chi^2$ test, the two distributions of labels are similar with $p \approx 1.0$. Therefore, the training split can effectively align models to the task of interest. 

\subsection{Queries}
\begin{figure}[pos=htb]
    \centering
    \subfloat{\includegraphics[width=0.45\linewidth]{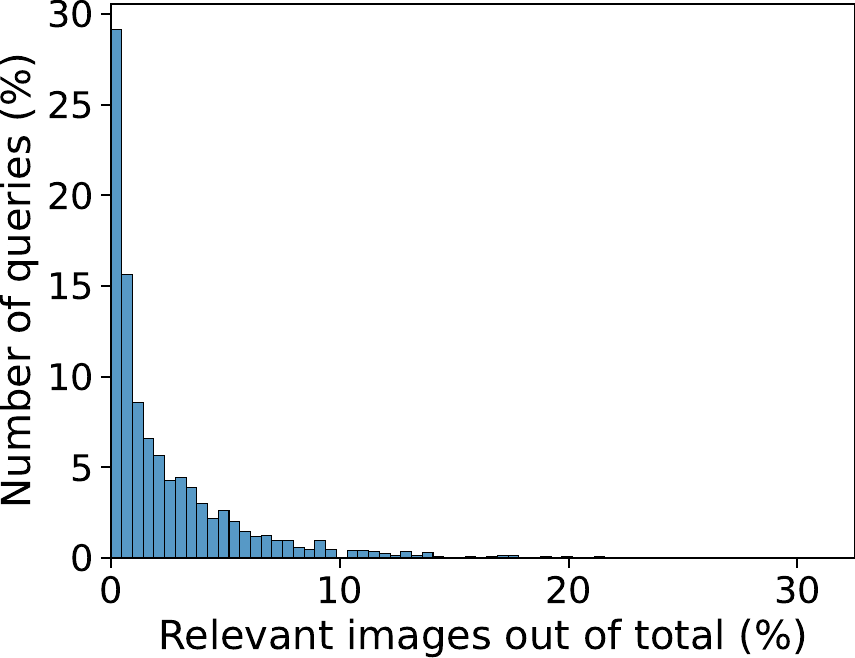}}
    \hfil
    \subfloat{\includegraphics[width=0.45\linewidth]{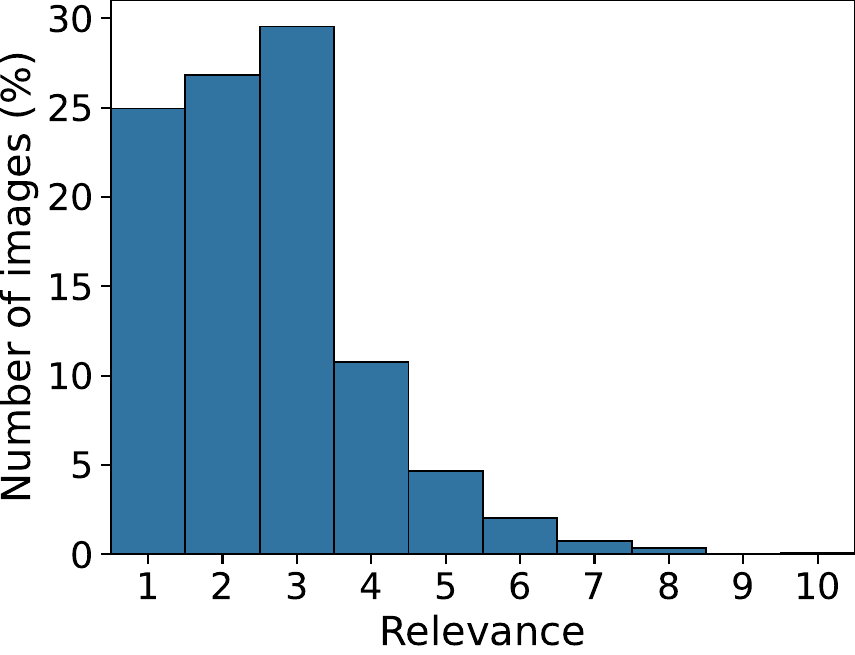}}
    \caption{Distribution of relevant images for each query and distribution of relevance scores.}
    \label{fig:relevant_distribution}
\end{figure}

We define the query set by taking every unique combination (with length from 1 to 12) of one or more labels that co-occur in at least one image within our corpus. This process resulted in a total of 2,047 distinct multi-label queries. We define a graded relevance $rel$ based on the IoU between the query labels $L_q$ and the image labels $L_i$ to provide a more fine-grained evaluation:
\begin{equation}
    rel = round\left(10\cdot\frac{|L_q \cap L_i|}{|L_q \cup L_i|}\right)
\end{equation}
For binary relevance operators (e.g., precision and recall), we consider a threshold of 5 for defining a relevant item. The distribution of relevance scores and relevant images per query is shown in \Cref{fig:relevant_distribution}.

\section{Methodology}

We present a new contrastive learning architecture, \underline{C}ontrastive \underline{L}anguage  \underline{O}ptical \underline{S}AR \underline{P}retraining (CLOSP), to address the T2RSIR task on the newly proposed corpus. 

\begin{figure}[pos=!htb]
    \centering
    \includegraphics[width=\linewidth]{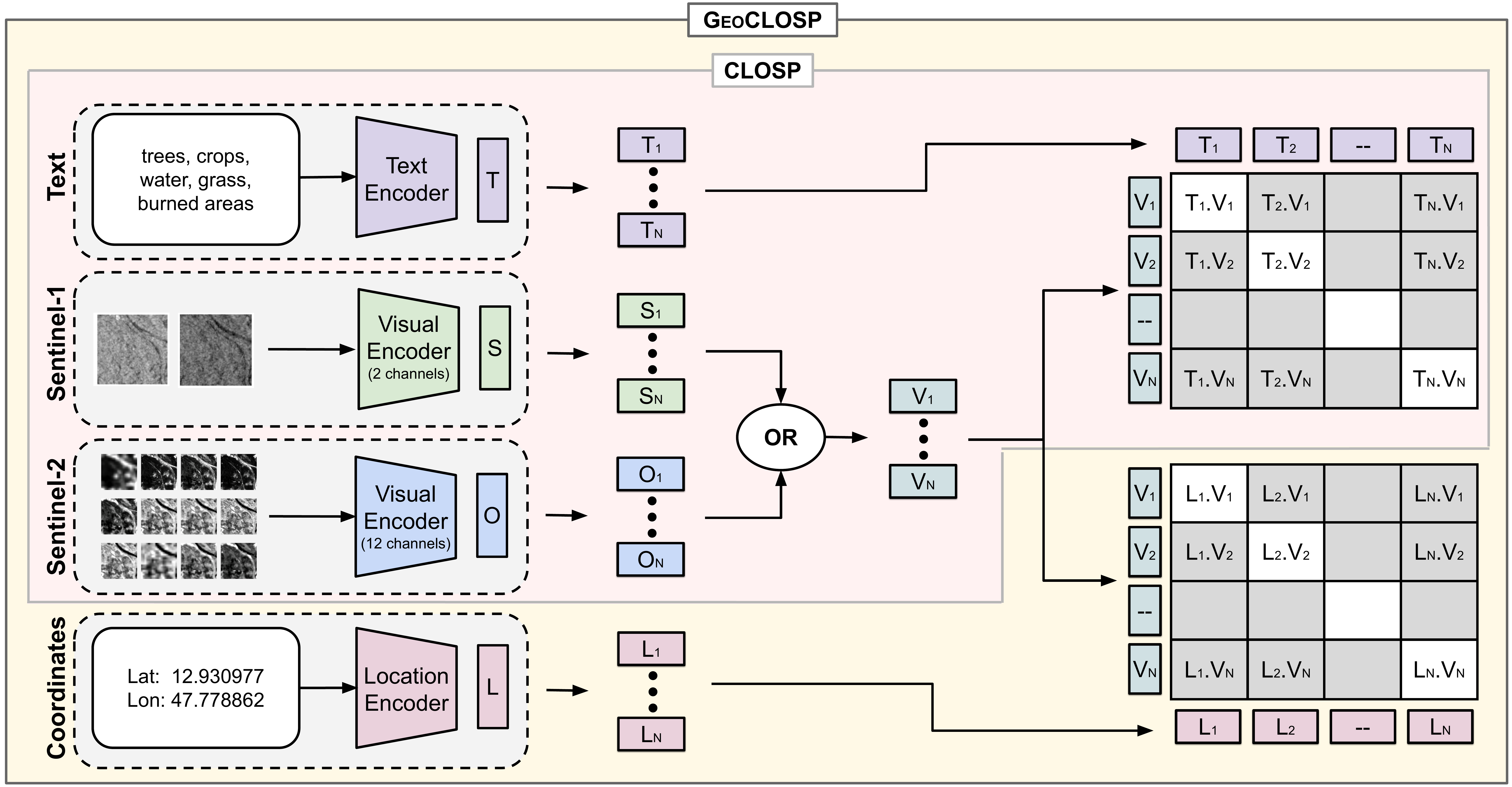}
    \caption{The CLOSP model aligns textual descriptions with SAR and optical satellite images. For a batch of $N$ elements ($M$ SAR and $M$ MSI), one modality—either SAR or optical—is selected for each element of the batch, and the corresponding image embeddings are paired with their associated textual embeddings. The model is trained to maximize alignment for these positive pairs (represented by white cells in the matrix) while ensuring that negative pairs, formed by combining textual and image embeddings from different items within the batch (represented by grey cells in the matrix), are effectively separated. GeoCLOSP extends the CLOSP architecture by incorporating a location encoder, which aligns the geographical coordinates of an item with the corresponding satellite image, in parallel with the image-text alignment.}
    \label{fig:osclip}
\end{figure}

The remainder of this section is organized as follows.
Section~\ref{sec:task} formulates the task, whereas Sections~\ref{sec:arch} and~\ref{sec:archgeo} respectively describe CLOSP and its variant that also leverages geospatial data (GeoCLOSP). 

\subsection{Task definition}
\label{sec:task}
Let $D$ be a large-scale remote sensing corpus where each item $i \in D$ consists of a satellite image and its associated metadata. A key characteristic of this task is the focus on imagery beyond the visible spectrum, where an image $i$ can be a multispectral optical product (e.g., from Sentinel-2) or a Synthetic Aperture Radar (SAR) product (e.g., from Sentinel-1). Each image $i$ is described by a set of structured textual annotations, $L_i$, representing its land cover, land use, and any relevant crisis events. Let $Q$ be the query space, where a query $q \in Q$ is composed of a set of one or more labels, $L_q$, from a predefined vocabulary (e.g., \{``water'', ``trees'', ``burned area''\}).

The Text-to-Remote-Sensing-Image Retrieval (T2RSIR) task aims to learn a relevance scoring function, $f: Q \times D \to \mathbb{R}$, that quantifies the semantic relevance of a candidate image $i$ to a given query $q$. The relevance is determined by the thematic alignment between the query labels $L_q$ and the image's labels $L_i$. The ultimate objective is to leverage this function, for any given query $q$, to retrieve and return a ranked list of the top-$k$ images from the corpus $D$ that maximize the relevance score $f(q, i)$.

For instance, given a query constructed from the labels (e.g., ``forest, burned area''), the retriever should return a ranked list of satellite images of Sentinel-1 and Sentinel-2 data that align with the textual description based on their relevance to the input query (see Figure~\ref{fig:teaser}).

\subsection{The  CLOSP architecture}
\label{sec:arch}

The CLOSP architecture is a contrastive learning network whose training schema is depicted in \Cref{fig:osclip}.
It employs three different encoders, 
one for text, one for Sentinel-1 (SAR) data, and one for Sentinel-2 (optical) data.
The encoders are trained to project their output representations into a shared latent space, aligning text with satellite imagery associated with the same item. Due to differences in satellite configurations, aligned images for the two visual modalities (i.e., SAR and optical) are rarely available. For example, the mean revisit time for a given area is 12 days for Sentinel-1 and 10 days for Sentinel-2. Thus, finding paired images for both modalities of the same area with a close timestamp, e.g., within the same day, is often not possible. 

To overcome this limitation, we align textual information simultaneously with both visual modalities during training, but without enforcing any direct alignment between the Sentinel-1 and Sentinel-2 images themselves. In this way, the text also acts as an anchor, linking each image modality independently within a shared latent space.

During a training iteration, the text encoder generates a textual embedding for each of the $N$ elements ($M$ SAR and $M$ MSI) in the current batch. Then, each image element (either SAR or MSI) is processed by the corresponding vision encoder to produce visual embeddings. Finally, following the CLIP pretraining strategy and employing symmetric cross-entropy loss (see \Cref{eq:loss_img,eq:loss_txt}), the textual and visual embeddings corresponding to the same item are simultaneously aligned (represented as white cells in the matrix in \Cref{fig:osclip}) while being separated from all other elements (depicted as grey cells in \Cref{fig:osclip}). The total loss $L$ is presented in \Cref{eq:total_loss}.

\begin{align}
    L_{\text{img}} = -\frac{1}{N} \sum_{i=1}^{N} \log \frac{\exp(\mathbf{i}_i \cdot \mathbf{t}_i / \tau)}{\sum_{j=1}^{N} \exp(\mathbf{i}_i \cdot \mathbf{t}_j / \tau)}\label{eq:loss_img} \\
    L_{\text{txt}} = -\frac{1}{N} \sum_{i=1}^{N} \log \frac{\exp(\mathbf{t}_i \cdot \mathbf{i}_i / \tau)}{\sum_{j=1}^{N} \exp(\mathbf{t}_i \cdot \mathbf{i}_j / \tau)}\label{eq:loss_txt} \\
    L = \frac{L_{\text{img}} + L_{\text{txt}}}{2}\label{eq:total_loss}
\end{align}

Where $N$ is the number of image-text pairs in a given batch. The vectors $\mathbf{i}_k$ and $\mathbf{t}_k$ represent the embeddings for the $k$-th image and its corresponding text caption, respectively. The parameter $\tau$ is a learnable temperature hyperparameter that scales the logits before the softmax operation.

\subsection{The GeoCLOSP architecture}
\label{sec:archgeo}

Unlike natural images, geospatial imagery is inherently associated with geographic coordinates. Ideally, image embeddings within the same region should exhibit similarities. To also consider the geographical position of the satellite images, we propose GeoCLOSP, an extension of the CLOSP architecture tailored to align geographical coordinates with the visual modality. 
Inspired by SatCLIP~\citep{klemmer2024}, we utilize 
a Sinusoidal Representation Network (SIREN)~\citep{sitzmann2020} and spherical harmonic (SH) positional encoding~\citep{russwurm2024} as a location encoder to embed the geographic coordinates associated with each image-text pair. 
The training schema, shown in \Cref{fig:osclip}, builds upon the standard CLOSP alignment of text and visual modalities while introducing the alignment of image and location embeddings (see \Cref{eq:loss_loc,eq:loss_iloc}). This dual alignment ensures that images with similar textual content share similar representations, and that images geographically close to each other also exhibit similarity. The loss $L_g$ follows the previous definition with two additional components as shown in \Cref{eq:total_geo_loss}.

\begin{align}
    L_{\text{loc}} = -\frac{1}{N} \sum_{i=1}^{N} \log \frac{\exp(\mathbf{l}_i \cdot \mathbf{i}_i / \tau)}{\sum_{j=1}^{N} \exp(\mathbf{l}_i \cdot \mathbf{i}_j / \tau)}\label{eq:loss_loc} \\
    L_{\text{iloc}} = -\frac{1}{N} \sum_{i=1}^{N} \log \frac{\exp(\mathbf{i}_i \cdot \mathbf{l}_i  / \tau)}{\sum_{j=1}^{N} \exp(\mathbf{i}_j \cdot  \mathbf{l}_i  / \tau)\label{eq:loss_iloc}} \\
    L_g = 0.5 \frac{L_{\text{img}} + L_{\text{txt}}}{2} + 0.5\frac{L_{\text{loc}} + L_{\text{iloc}}}{2}\label{eq:total_geo_loss}
\end{align}

Where $\mathbf{l}_k$ represents the embeddings for the $k$-th location associated with the corresponding image. An analysis on the impact of the weighting between the two components is reported in \Cref{sec:loss_weighting}.

\section{Experiments}

In this section, after introducing the baseline methods, empirical settings, and performance metrics, we discuss the main experimental results achieved by CLOSP, GeoCLOSP, and the state-of-the-art models on the newly proposed corpus.

\subsection{Baselines}
\label{sec:baselines}

We compare the performance of CLOSP with that of CLIP \citep{ilharco2021,Radford2021}, SkyCLIP \citep{wang2024}, RemoteCLIP \citep{liu2024}, SenCLIP \citep{jain2024}, and Llama3-MS-CLIP \citep{marimo2025} (named Llama3-CLIP for simplicity) on the CrisisLandMark corpus. To the best of our knowledge, these models are the latest and best-performing T2RSIR models. We also fine-tune them on the training set of our dataset (hereafter we denote them as <ModelName>-T). This provides an overview of the contribution of other spectral and radiometric information compared to RGB only. We also trained two specialized models (Text-SAR and Text-MS) to highlight the benefits of the unified space: we refer to the evaluation of these models together as BiCLIP. We also trained a CLOPS-RGB model using an RGB encoder for Sentinel-2 and a False Color encoder for Sentinel-1 to additionally highlight the limitations of the interactions between RGB and SAR. Additionally, we provide a dummy baseline for each set of experiments. We provide CLOSP with three different backbones: ResNet-50 (CLOSP-RN), ViT-Small (CLOSP-VS), and ViT-Large (CLOSP-VL). All the specific settings regarding the analyzed models' composition and efficiency are reported in \Cref{tab:models}.

\begin{table}[pos=htb]
    \centering
    \caption{Models and respective textual and vision backbones. Each CLOSP suffix indicates the employed vision backbone. The GFLOPs are calculated for the worst case at indexing time (i.e., with 12-channel optical images).}
    \label{tab:models}
    \begin{tabular}{l|ccc}
    \toprule
                & Vision Backbone & Textual Backbone & GFLOPs   \\\midrule
    CLIP        & ResNet50        & CLIP-Transformer  & 3      \\
    SkyCLIP     & ViT-L           & CLIP-Transformer  & 103    \\
    RemoteCLIP  & ResNet-50       & CLIP-Transformer  & 3     \\
    SenCLIP     & ResNet-50       & CLIP-Transformer  & 3    \\
    Llama3-CLIP & ViT-B        & CLIP-Transformer     & 23    \\
    CLOSP-RN    & ResNet-50       & MiniLM            & 3     \\
    CLOSP-VS    & ViT-S           & MiniLM            & 9     \\
    CLOSP-VL    & ViT-L           & MiniLM            & 120   \\
    GeoCLOSP    & ResNet-50       & MiniLM            & 3     \\\bottomrule
\end{tabular}

\end{table}

\subsection{Training Settings}
\label{sec:setting}

When testing existing approaches, since they can deal with RGB only (i.e., three channels), we extracted the RGB bands from Sentinel-2 (originally 12 channels), and we created a false-color composite \citep{s1Processing} from Sentinel-1 channels (originally two channels). 
We fine-tune all models for 30 epochs, with batch size 64, Adam optimizer, cosine annealing learning rate scheduler with warmup, and maximum learning rate 1e-4. We leverage pre-trained encoders: a vision encoder from SSL4EO \citep{yi2023}, a text encoder from SentenceTransformers \citep{reimers2019}, and a location encoder from SatCLIP \citep{klemmer2024}. GeoCLOSP and specialized models (Text-SAR and Text-MS) are based on ResNet-50 for simplicity, faster training, and reduced risk of underfitting compared to ViT models.

\subsection{Text-to-Image Retrieval}
\label{sec:results}
In this section, we present the metrics used for retrieval evaluation, the retrieval settings, and the empirical results. 

\subsubsection{Evaluation Metrics}
\label{sec:metrics}

We evaluate the retrieval performance in terms of Recall (R), Precision (P), and normalized Discounted Cumulative Gain (nDCG) \cite{Manning2008}. We experiment with common cutoff levels $K \in \{10, 50, 100, 1000\}$ to balance retrieval efficiency and common user needs. nDGC provides the most comprehensive evaluation for graded relevance. For the evaluation of precision and recall, we use a standard relevance threshold of 5 (i.e., IoU of 0.5). The random baseline performance is based on analytical computation: for each query, $R@K = K/|D|$, $P@K = R_q/|D|$ and $nDCG@K = R_m \frac{\sum_{i=1}^{K} 1/log_2(i + 1)}{IDCG@K}$, where $R_m$ is the average relevance score and $R_q$ is the number of relevant items.

\subsubsection{Retrieval Settings}
We employ ChromaDB as a vector database to store the normalized embeddings. We use the inner product as the distance function, which is computationally faster for normalized vectors and is equivalent to cosine similarity. The stored embeddings have the following sizes: 384 for our CLOSP model, 768 for SkyCLIP, and 1024 for SenCLIP and RemoteCLIP, as in their original configurations. We retrieve 1000 images for each query to balance efficiency and completeness.
In this case, the dummy retriever is the random uniform sampling. To obtain a unified ranking for the two specialized models, we independently retrieve the images, apply min-max normalization to the scores from each model, and merge them, keeping only the top 1000 for a fair comparison. 

\subsubsection{Results}
The text-to-image retrieval performance of our proposed models against several baselines is detailed in \Cref{tab:results}. Note that due to the large number of relevant items for some queries, a perfect Recall@1000 score is not always reachable. However, the relative differences demonstrate, in any case, the model's capabilities. To confirm the observations, we performed a Friedman test followed by a Conover post-hoc test for multiple tests for statistical significance (p < 0.01) \citep{janex2006,Conover1998}.

Among the original, non-finetuned models, only SenCLIP provides a competitive solution, outperforming the dummy baseline. However, after being fine-tuned on the CrisisLandMark training set, the baselines show divergent behavior. While SkyCLIP-T and RemoteCLIP-T improve significantly (according to the statistical test) upon their original counterparts, SenCLIP-T's performance degrades significantly. This result may be due to overfitting. Consequently, SkyCLIP-T emerges as the strongest baseline competitor.

Our CLOSP family of models significantly (according to the statistical test) outperforms all baselines (including BiCLIP and CLOSP-RGB) across every evaluated metric. The top-performing model, GeoCLOSP, achieves an nDCG@1000 of 57.76\%, representing a nearly 20-point absolute improvement over the best baseline, SkyCLIP-T (37.88\%).

This demonstrates the significant advantage of our unified architecture. It is also important to note that while CLOSP-VL employs a much larger backbone than the other variants, it provides only marginal gains. CLOSP-RN and CLOSP-VS do not exhibit a statistically significant difference from each other in terms of recall and precision. Some visual examples are shown in \Cref{sec:visual_examples}.

\begin{table}[pos=htb]
    \centering
    \caption{Mean performance (\%) for each model in terms of nDCG, Precision (P), and Recall (R) at given cutoffs. \textbf{Bold} values indicate the best result for each metric.}
    \label{tab:results}
    \sisetup{detect-weight=true, detect-mode=true} %
\begin{tabular}{
  l
  S[table-format=2.2]
  S[table-format=2.2]
  S[table-format=2.2]
  S[table-format=1.2]
}
\toprule
Model & {nDCG@10} & {nDCG@1000} & {P@1000} & {R@1000} \\
\midrule
Dummy        & 28.03 & 33.64 &  2.50 & 0.15 \\ \midrule
CLIP         & 18.64 & 20.70 &  4.57 & 0.13 \\
RemoteCLIP   & 19.27 & 21.68 &  4.08 & 0.09 \\
SkyCLIP      & 24.88 & 28.21 &  7.10 & 0.21 \\
SenCLIP      & 34.01 & 37.60 & 18.15 & 0.52 \\
Llama3-CLIP  & 29.12 & 31.32 & 10.18 & 0.32 \\ \midrule
RemoteCLIP-T & 29.00 & 26.57 &  8.70 & 0.27 \\
SkyCLIP-T    & 33.46 & 37.88 & 17.57 & 0.58 \\
SenCLIP-T    & 20.59 & 23.76 &  5.41 & 0.13 \\
Llama3-CLIP-T& 33.40 & 37.75 & 16.87 & 0.50 \\ \midrule
BiCLIP       & 38.33 & 44.90 & 27.20 & 1.31 \\
CLOSP-RGB    & 37.39 & 44.56 & 25.47 & 0.81 \\ \midrule
CLOSP-RN     & 50.50 & 56.23 & 40.66 & 2.05 \\
CLOSP-VS     & 49.18 & 54.51 & 40.22 & 2.14 \\
CLOSP-VL     & 47.82 & 55.91 & 42.14 & \bfseries 2.32 \\
GeoCLOSP     & \bfseries 51.14 & \bfseries 57.76 & \bfseries 42.98 & 2.10 \\
\bottomrule
\end{tabular}

\end{table}

\begin{figure}[pos=htb]
    \centering
    \includegraphics[width=\linewidth]{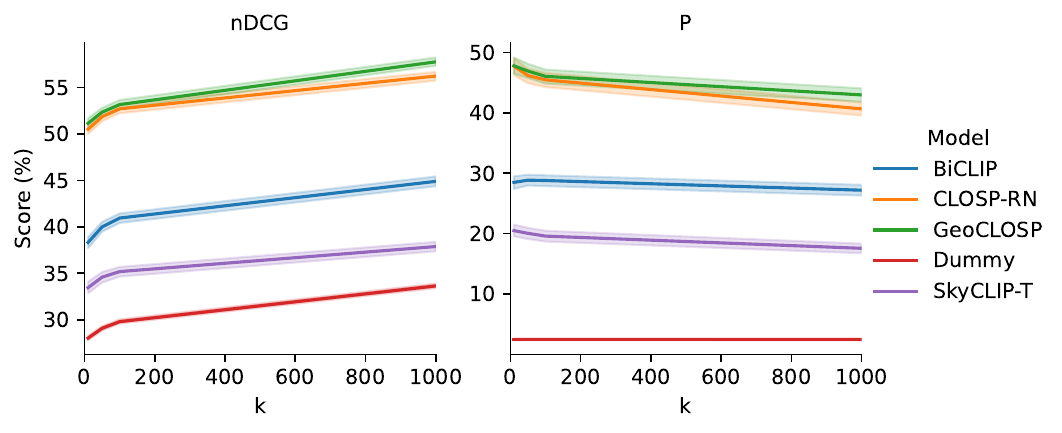}
    \caption{Mean nDCG (left) and Precision (right) performance at different cutoff levels. The bands represent the 95\% confidence intervals.}
    \label{fig:metrics_trend}
\end{figure}

To visualize the ranking quality at various cutoffs, \Cref{fig:metrics_trend} shows the nDCG and Precision curves for our CLOSP-RN and GeoCLOSP (which shares the same backbone) against the strongest baseline and the dummy predictor. The plots clearly illustrate that our models maintain a consistently higher performance across all values of k.

\subsection{Text-to-Image Retrieval by Modality}
To isolate and understand the contribution of each visual modality, we evaluated our models on the Sentinel-1 (SAR) and Sentinel-2 (optical) portions of the corpus separately. For this analysis, each specialized model of BiCLIP was evaluated only on its corresponding data subset (e.g., Text-SAR on Sentinel-1 images). Our unified CLOSP models were also evaluated on each subset individually to enable a direct, per-modality comparison.

\subsubsection{Results}
The retrieval performance of our unified models compared to specialized baselines for each modality is presented in \Cref{tab:modality_results}. The findings highlight the significant advantages of our joint training approach, particularly for the more challenging Sentinel-1 data. CLOSP-RGB does not provide comparable results for either Sentinel-2 or Sentinel-1, so we focus our comparison on the best baseline, i.e., BiCLIP.

For Sentinel-1 retrieval, all CLOSP variants dramatically outperform the specialized BiCLIP baseline. Our best variant, CLOSP-VL, achieves an nDCG@1000 of 55.65\%, which is +18 absolute improvement over BiCLIP's 37.35\%. This result provides strong evidence that the unified architecture facilitates a cross-modal knowledge transfer, where rich semantic concepts learned from the easier-to-interpret optical data are leveraged to disambiguate the complex backscatter signals in the more challenging SAR imagery, even without any direct interactions between the two visual encoders.

Conversely, for Sentinel-2 retrieval, our models remain highly competitive with the specialized baseline. CLOSP-RN, for instance, shows only a marginal performance decrease in nDCG@1000 compared to BiCLIP (54.80\% vs 55.72\%). A paired t-test confirms that while the gains on Sentinel-1 are statistically significant (p < 0.01), the small performance decrease on Sentinel-2 is also significant, though minor in magnitude. This result demonstrates that our unified training strategy provides great gains for the SAR modality at a negligible cost of a minor drop in optical performance, confirming the value of the joint training paradigm.

\begin{table}[pos=htb]
    \centering
    \caption{Mean performance (\%) for each model by modality in terms of nDCG, Precision (P), and Recall (R) at given cutoffs. \textbf{Bold} values indicate the best result for each metric. \textit{S1} is Sentinel-1 SAR data and \textit{S2} is Sentinel-2 multispectral data.}
    \label{tab:modality_results}
    
    \sisetup{detect-weight=true, detect-mode=true} %
        \begin{tabular}{
            @{}
            l
            S[table-format=2.2]
            S[table-format=2.2]
            S[table-format=2.2]
            S[table-format=2.2]
            S[table-format=2.2]
            S[table-format=2.2]
            S[table-format=1.2]
            S[table-format=1.2]
            @{}
        }
        \toprule
        Model & \multicolumn{2}{c}{nDCG@10} & \multicolumn{2}{c}{nDCG@1000} & \multicolumn{2}{c}{P@1000} & \multicolumn{2}{c}{R@1000} \\
        \cmidrule(lr){2-3} \cmidrule(lr){4-5} \cmidrule(lr){6-7} \cmidrule(lr){8-9}
              & {S1}    & {S2}    & {S1}    & {S2}    & {S1}    & {S2}    & {S1}    & {S2}    \\ \midrule
        BiCLIP    & 31.99 & \bfseries 49.56 & 37.35 & \bfseries 55.72 & 18.51 & \bfseries 39.65 & 0.78 & 2.24 \\
        CLOSP-RGB & 20.55 & 38.71 & 20.16 & 46.85 & 7.18 & 27.55 & 0.18 & 0.90 \\\midrule
        CLOSP-RN  & 45.93 & 49.33 & 53.60 & 54.80 & 37.51 & 38.46 & 1.76 & 1.92 \\
        CLOSP-VS  & 45.96 & 42.93 & 51.79 & 49.96 & 37.43 & 34.86 & 1.75 & 2.12 \\
        CLOSP-VL  & \bfseries 46.59 & 45.97 & \bfseries 55.65 & 53.85 & \bfseries 41.83 & 38.04 & \bfseries 2.12 & \bfseries 2.26 \\ \bottomrule
        \end{tabular}

\end{table}

\subsection{Zero-shot classification}
In this section, we analyze the performance of the CLOSP models in zero-shot multilabel classification over the whole corpus. Similarly to \cite{Radford2021} and \cite{wang2024}, we evaluate the “zero-shot” transfer capability of the model on an unseen dataset (corpus without the training set), instead of testing the model's generalizability on unseen object categories. This tests the model's generalization to new locations and scenes without any fine-tuning to assess how well the learned semantic representations can be used for direct classification. While some zero-shot evaluations focus on holding out entire classes, this approach is less suitable for foundational LULC classification. The selected classes form a comprehensive and interdependent set describing the terrestrial surface; training a model without knowledge of a fundamental class like ``water” would inhibit its ability to learn related concepts such as ``flooded vegetation”. Therefore, evaluating the transfer capability to unseen imagery is a more practical and meaningful test in this domain.

\subsubsection{Settings}
The multilabel classification is performed in a zero-shot fashion, leveraging the embeddings of the model. Given the text encoder $E_t$, the visual encoder $E_v$, and the set of 12 class (e.g; water, trees, burned area) keywords $C = \{c_1, c_2, ..., c_{12}\}$, the task can be formalized in the following way.
For each class keyword $c_j \in C$, we generate the class vector $v_j = E_t(c_j)$. For each image $i$ in the corpus $D$, we generate the visual embeddings $u_i = E_v(i)$. We create a similarity matrix $S$ of shape $|D| \times 12$ computing the cosine similarity between each $u_i$ and $v_j$. A class $c_j$ is predicted as present in an image $i$ if its similarity score exceeds a threshold $t$. 

To select a decision threshold $t$ for each model in a fair, zero-shot manner that accounts for their different score distributions, we adopted a global thresholding strategy. After the creation of $S$ for a specific model, the mean of this entire distribution of scores was then used as the single, model-specific classification threshold. While more complex methods exist, this simple, dynamic threshold was chosen to maintain a strict zero-shot setting without requiring a validation set.

We compare baselines and all CLOSP proposed models with a dummy classifier that always predicts the two most frequent classes, i.e., ``crops” and ``trees”.
Additionally, when evaluating the BiCLIP solution, since each specialized model makes predictions for its image type without item overlapping, we merge the predictions of all dataset elements before computing the final performances.
Finally, given the strong class imbalance, we evaluate the performance using macro-averaged precision (P), recall (R), and F1-score (F) over the classes. 

\subsubsection{Results}
The results of the zero-shot classification are presented in \Cref{tab:classification_results}. Our proposed CLOSP models demonstrate a clear advantage over existing baselines. Specifically, CLOSP-VS and CLOSP-RN achieve the highest performance with macro F1-scores of 41.82\% and 41.56\%, respectively, outperforming the best baseline model, i.e., SkyCLIP-T with an F1-score of 32.81\%.

An important finding is the effect of the location encoder. The GeoCLOSP model, which incorporates geographical data, shows no improvement over its location-unaware counterpart CLOSP-RN (41.24\% vs 41.56\% F1-score). This suggests that for this semantic classification task, the additional geographic information does not provide a benefit and may indicate a trade-off between learning semantic and spatial features.

Furthermore, we observe different model behaviors within our proposed family. For instance, CLOSP-VL exhibits very high precision (62.48\%) at the cost of lower recall (37.40\%), making it a more conservative classifier, while CLOSP-RN achieves a much higher recall (82.25\%) with more modest precision (35.59\%). All models substantially outperform the dummy classifier baseline, confirming the general effectiveness of the zero-shot approach.

To determine if differences are significant, we performed a Bayesian signed-rank test \citep{benavoli2014}, which is more suitable than frequentist tests for comparisons with a small number of classes (12 labels) \citep{benavoli2017}. 
CLOSP-VS, the best-performing model in terms of F1-score, is better than all baselines with high probability (> 0.9). CLOSP-RN achieves the highest recall, surpassing all models (probability larger than 0.9, except for SkyCLIP 0.88), proving its top-tier performance. CLOSP-VL, which achieves the best precision, also outperforms all baselines with high probability (> 0.9). Comparisons among the CLOSP variants were less conclusive, underscoring their similarities.

\begin{table}[pos=htb]
    \centering
        \caption{Zero-shot classification performance (\%) for each model in terms of F1-Score (F), Precision (P), and Recall (R). \textbf{Bold} values indicate the best result for each metric.}
    \label{tab:classification_results}
    \begin{tabular}{lrrr}
\toprule
Model        & \multicolumn{1}{c}{F} & \multicolumn{1}{c}{P} & \multicolumn{1}{c}{R}  \\ 
\midrule
Dummy & 12.87 & 10.52 & 16.67 \\\midrule
CLIP & 25.12 & 21.99 & 52.02 \\
RemoteCLIP & 25.61 & 21.15 & 48.11 \\
SkyCLIP & 29.10 & 22.58 & 58.96 \\
SenCLIP & 17.73 & 20.33 & 49.61 \\
Llama3-CLIP & 22.30 & 30.28 & 49.82 \\\midrule
RemoteCLIP-T & 15.81 & 11.32 & 50.00 \\
SkyCLIP-T & 32.81 & 26.30 & 68.18 \\
SenCLIP-T & 5.41 & 4.48 & 33.34 \\
Llama3-CLIP-T & 36.45 & 30.89 & 75.29 \\\midrule
BiCLIP & 34.98 & 30.89 & 69.83 \\
CLOSP-RGB & 23.97 & 31.47 & 61.58 \\\midrule
CLOSP-RN & 41.56 & 35.59 & \textbf{82.25} \\
CLOSP-VS & \textbf{41.82} & 45.22 & 55.68 \\
CLOSP-VL & 37.31 & \textbf{62.48} & 37.40 \\
GeoCLOSP & 41.24 & 40.40 & 68.14 \\
\bottomrule
\end{tabular}

\end{table}

\subsection{Spatial Distance Correlation Analysis}
In this section, we analyze the correlation between the embedding distances and the corresponding image geographical distances to understand the effects of the location encoder.

\subsubsection{Settings}
To analyze the relationship, we first drew two disjoint sets of 10,000 images each via uniform random sampling from the entire corpus. We then created 10,000 pairs by matching the first image from set A with the first from set B, the second with the second, and so on. For each of these pairs, we computed two distance values: the geographic distance using the Haversine formula and the embedding distance using the cosine distance (computed as 1 $-$ cosine similarity) between the embeddings. We analyze Pearson and Spearman correlations to understand both linear and monotonic relationships.

\subsubsection{Results}
\begin{figure}[pos=htb]
    \centering
    \includegraphics[width=0.7\linewidth]{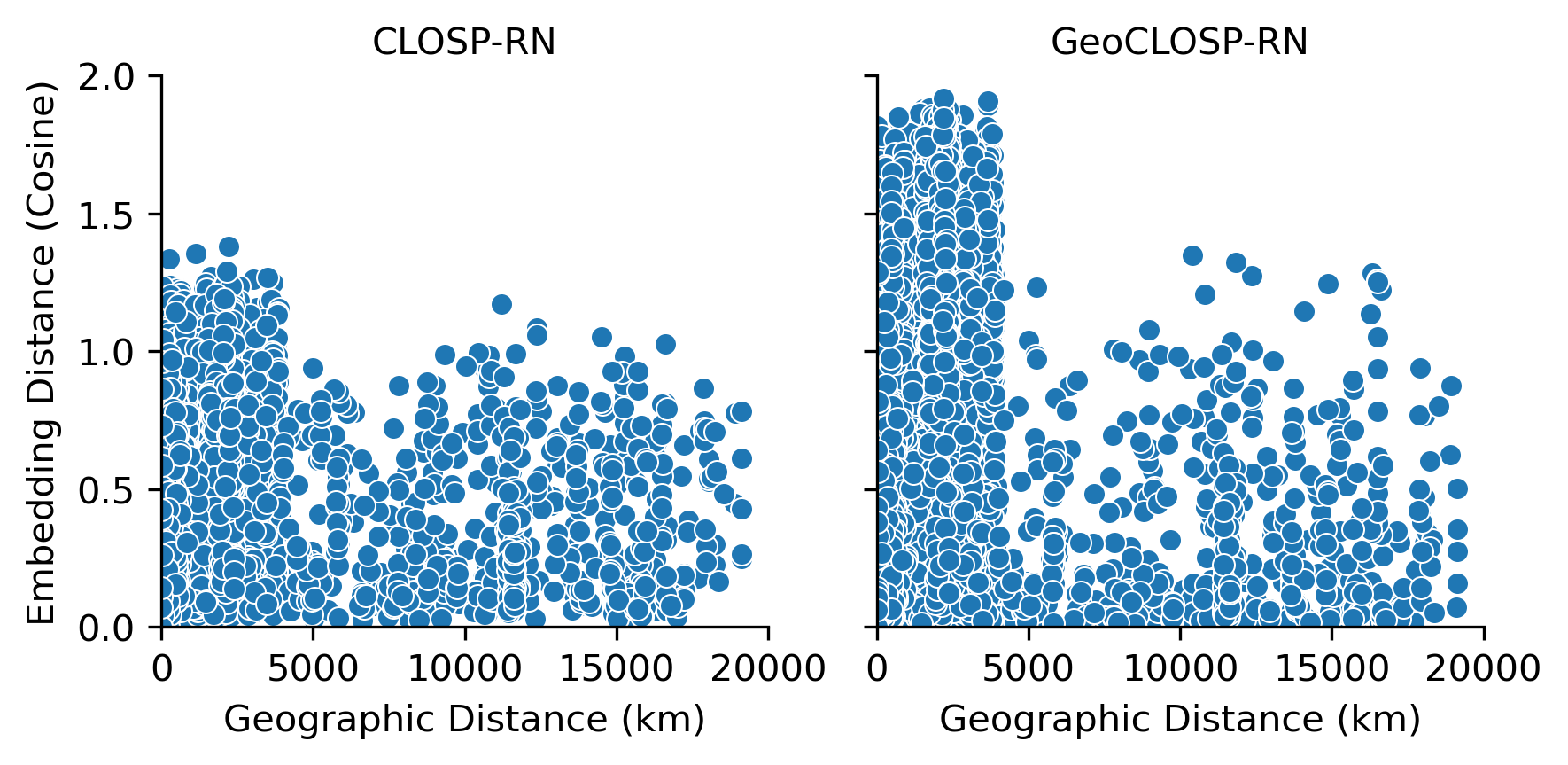}
    \caption{Relationship between geographic (Haversine) distance and embedding (cosine) distance for 10,000 image pairs.}
    \label{fig:geo_vs_embs}
\end{figure}

To directly assess the impact of the location encoder, we compare the results from GeoCLOSP against its location-unaware counterpart, CLOSP-RN. For the CLOSP-RN model, we observe a negligible linear correlation (Pearson's $r \approx -0.06$, $p < 0.01$) and a weak monotonic correlation (Spearman's $r_s \approx 0.13$, $p < 0.01$). In contrast, while the GeoCLOSP model also shows no linear relationship (Pearson's $r \approx -0.04$, $p < 0.01$), it achieves a stronger monotonic correlation (Spearman's $r_s \approx 0.34$, $p < 0.01$). The relationship between geographical and embedding distances of the sampled sets of points is illustrated in \Cref{fig:geo_vs_embs}.

This divergence demonstrates that our location encoder successfully introduces a moderate, monotonic structure into the latent space. The improvement in the Spearman coefficient confirms the effectiveness of our location-aware training approach. However, the modest absolute correlation value indicates that the embeddings remain primarily organized by semantic content rather than geographic proximity. This highlights a key challenge and a promising direction for future work: developing novel training strategies to achieve a more balanced representation of both semantic and geographic features.

\subsection{Single classes performance}
In this section, we report the performance by class in retrieval (nDCG) and classification (F1-score) for SkyCLIP-T, CLOSP-RN, and GeoCLOSP, to better understand their fine-grained differences. 
In this case, since we do not report mean results among classes, to determine statistical significance, we performed a bootstrap analysis \citep{Kuhn2013fr,efron1992bootstrap} with 1000 iterations for each class, applying a Bonferroni correction \citep{bonferroni1936} for multiple comparisons.

The per-class performance is detailed in \Cref{tab:per_class_comparison}, revealing a clear trade-off between general semantic performance and specialization on geographic and crisis-related events. While SkyCLIP-T serves as a strong baseline, our statistical analysis confirms that both CLOSP-RN and GeoCLOSP significantly outperform it across the majority of classes for both metrics. The statistical analysis confirms that \textbf{most observed performance differences are significant} ($p < 0.01$). The primary exceptions where no significant difference was found are:
\begin{itemize}
    \item For the F1-score, between CLOSP-RN and GeoCLOSP on the ``water” and ``burned area” classes.
    \item For the nDCG metric, between CLOSP-RN and GeoCLOSP on ``burned area”; between GeoCLOSP and SkyCLIP-T on ``trees”; and among all three models on the ``water'' class.
\end{itemize}

Our primary finding is that, while absolute F1-scores for rare classes are expectedly modest, GeoCLOSP improves performance on classes where location is a key factor. For crisis events such as earthquake damage and rare geographic features like snow and ice, GeoCLOSP shows a remarkable improvement in retrieval. Its nDCG score for ``earthquake damage” is 65.5\%, whereas the baseline CLOSP-RN completely fails with a score of 0. For ``snow and ice”, the nDCG score similarly improves from 18.47\% to 60.9\%. These results show that the location encoder introduces crucial contextual information that visual features alone cannot capture.

Conversely, for common, geographically widespread semantic classes like crops and trees, the baseline CLOSP-RN consistently achieves the highest scores in both classification and retrieval. For instance, CLOSP-RN scores an F1 of 84.84\% on ``trees”, while GeoCLOSP scores 78.95\%. This suggests that enforcing a geographic structure slightly dilutes the model's performance on these very general semantic concepts.

Interestingly, for some classes, the impact is more nuanced. For burned area and water, the performance between CLOSP-RN and GeoCLOSP is not statistically significantly different, suggesting that the strong visual and semantic signals of these classes dominate over the geographic information. The results clearly indicate that the choice between CLOSP-RN and GeoCLOSP presents a trade-off depending on the user's application: CLOSP-RN excels at general-purpose semantic tasks, while GeoCLOSP is a more powerful, specialized tool for location-specific and crisis-related analysis.

\begin{table}[pos=htb]
    \centering
    \caption{Performance by class in terms of F1-score (F) for zero-shot classification and nDCG@1000 (nDCG) for retrieval.}
    \label{tab:per_class_comparison}
    \sisetup{
    table-format=2.2,
    detect-weight
}
\begin{tabular}{ l S S S S S S }
\toprule
& \multicolumn{2}{c}{\textbf{CLOSP-RN}} & \multicolumn{2}{c}{\textbf{GeoCLOSP}} & \multicolumn{2}{c}{\textbf{SkyCLIP-T}} \\
\cmidrule(lr){2-3} \cmidrule(lr){4-5} \cmidrule(lr){6-7}
\textbf{Class} & {F} & {nDCG} & {F} & {nDCG} & {F} & {nDCG} \\
\midrule
bare                & 16.11 & \textbf{25.47} & \textbf{18.08} & 22.81 & 11.90 &  1.99 \\
built               & \textbf{46.15} & \textbf{72.08} & 42.92 & 58.39 & 30.93 & 27.59 \\
burned area         &  2.34 & 60.07 & \textbf{2.84}  & \textbf{61.76} &  1.39 &  0.00 \\
crops               & \textbf{76.76} & \textbf{90.53} & 68.82 & 67.84 & 67.00 & 79.21 \\
earthquake damage   & 16.33 &  0.00 & \textbf{21.45} & \textbf{65.50} & 14.65 & 23.29 \\
flooded area        & 38.87 & 32.09 & \textbf{45.94} & 15.91 & 20.39 & \textbf{32.40} \\
flooded vegetation  & 22.06 & \textbf{43.54} & \textbf{25.60} & 25.69 & 14.93 &  2.17 \\
grass               & \textbf{52.61} & \textbf{74.40} & 52.51 & 42.13 & 40.75 & 63.25 \\
shrub and scrub     & 58.58 & \textbf{50.87} & \textbf{60.60} & 30.81 & 53.65 & 40.05 \\
snow and ice        &  9.46 & 18.47 & \textbf{11.15} & \textbf{60.90} &  6.40 &  0.15 \\
trees               & \textbf{84.84} & \textbf{80.58} & 78.95 & 59.23 & 80.45 & 58.59 \\
water               & \textbf{74.61} & \textbf{99.77} & 65.96 & 99.30 & 51.30 & 99.67 \\
\bottomrule
\end{tabular}

\end{table}

\section{Discussion}
\label{sec:discussion}

Our study introduced CrisisLandMark, a novel large-scale corpus for Text-to-Remote-Sensing-Image Retrieval (T2RSIR), and CLOSP, a multimodal architecture designed to operate on satellite data beyond the visible spectrum. The experimental results demonstrate a significant improvement in performance over existing methods, and this section discusses the interpretation and implications of these findings.

\subsection{Principal Findings and Contributions}
The primary finding of this work is that by training a contrastive model on a diverse corpus of multispectral (Sentinel-2) and SAR (Sentinel-1) imagery, it is possible to substantially outperform state-of-the-art T2RSIR systems that are primarily designed for RGB data. Our proposed model, CLOSP, achieved a better retrieval and classification performance than the baselines, confirming that the rich information contained in non-RGB channels is not only useful but critical for accurately retrieving images based on land cover, land use, and crisis-event descriptions.

A key methodological insight from our experiments is that continual pre-training of an existing vision-language model like CLIP \citep{Radford2021} is not a prerequisite for high performance in this domain. Our results show that good performance can be achieved by aligning powerful, independently pre-trained unimodal encoders. By initializing CLOSP with a vision encoder from SSL4EO and a text encoder from Sentence Transformers, and then training them jointly with a contrastive objective, we demonstrate a flexible and effective alternative to extending pre-aligned models.

Furthermore, GeoCLOSP highlights the nuanced but critical role of location. We demonstrated that explicitly encoding geographic coordinates provides a contextual signal for location-dependent phenomena, such as earthquakes or specific climatic zones (e.g., snow and ice), while also showing that a trade-off exists for more semantically-defined classes.

\subsection{The Synergy of a Unified Optical-SAR Latent Space}
One of the most significant results is the demonstrated benefit of jointly training on optical and SAR data within a unified embedding space. As shown in our modality-specific evaluation (\Cref{tab:modality_results}), the proposed approach improved retrieval performance for Sentinel-1 (SAR) imagery, a data source notoriously difficult for semantic interpretation due to its different imaging physics compared to optical sensors. This suggests a powerful knowledge transfer mechanism: the model leverages the clear semantic signatures from Sentinel-2's multispectral bands (e.g., identifying vegetation or water) to learn corresponding structural and textural patterns in the Sentinel-1 SAR data, without any direct alignment strategy (there is no loss between the two vision encoders' embeddings).

This synergy is crucial for practical applications. It allows for the development of a single, robust retrieval system that can easily query across different sensor types. An end-user can submit a natural language query, such as ``flooded vegetation”, and retrieve relevant satellite images regardless of whether they are from an optical or SAR satellite. Our model effectively bridges the gap between the two modalities with an indirect alignment.

\subsection{The Trade-off Between Semantic and Geographic Representations}
The comparison between CLOSP and GeoCLOSP uncovers a fundamental trade-off in designing foundation models for Earth observation: the balance between semantic content and geographic context. Our results (Table \ref{tab:per_class_comparison}) clearly show that GeoCLOSP excels in retrieving images for classes where location is a key defining characteristic. For instance, its ability to retrieve images of ``earthquake damage” (a geographically localized event) and ``snow and ice” (a climatically constrained feature) was superior to the baseline CLOSP. In these cases, the location encoder provides an essential prior that visual features alone cannot capture.

Conversely, for geographically widespread classes such as ``crops” and ``trees”, the purely semantic CLOSP-RN model performed better. This suggests that enforcing a geographic structure on the latent space can slightly ``dilute” the model's focus on fine-grained visual semantics when the location is not a primary discriminative factor. This finding implies that the optimal retrieval strategy may depend on the nature of the query. Future systems could potentially employ an adaptive approach, dynamically weighting the influence of the location encoder based on the query's geographic specificity.

\subsection{Implications for Real-World Applications}
The advancements presented in this paper have direct implications for several domains. 
In \textbf{crisis management}, the ability to retrieve relevant imagery of events like floods or wildfires is a significant operational advantage. First responders and analysts can obtain a more complete and timely picture of the situation on the ground simply by knowing the possibly large area that is affected, leaving the search engine to find the exact points affected by the crisis (e.g., you may know a wildfire is near San Jose, but leave the engine search for all areas near the city that are affected). Importantly, our integration of SAR data can significantly boost performance in such scenarios, where optical data may be limited (e.g., due to cloud cover or night-time conditions). For \textbf{environmental monitoring and urban planning}, CLOSP makes large-scale land cover analysis more accessible. Researchers and policymakers can query vast archives using simple textual descriptions to track deforestation, monitor urban expansion, or assess agricultural health without needing to write complex code or manually filter thousands of images. For example, they could retrieve Sentinel-2 images to assess crop health via spectral indices and Sentinel-1 images to monitor changes in field structure or planting/harvesting activities, all through a single, unified system. This democratizes access to valuable Earth observation data.

\subsection{Limitations}
Despite the promising results, this study has several limitations that open avenues for future research. First, while CrisisLandMark is large, its source datasets may introduce geographic biases (e.g., a focus on Europe from the re-BEN dataset). 
Furthermore, our work focused on single-image retrieval. However, many remote sensing applications, such as change detection and trend analysis, rely on time-series data. A valuable and logical next step is to extend the CLOSP architecture to a temporal dimension. This would enable novel queries based on dynamic processes, for instance, allowing users to directly ask for ``deforestation'' between 2020 and 2024 or ``urban growth'' over the last decade. %
Additional new classes could benefit from a temporal perspective (e.g., deforestation, desertification). Additionally, our GeoCLOSP model uses a fixed weighting for its semantic and geographic loss components; future work could explore dynamic or adaptive weighting schemes to better balance these competing objectives based on the query type.

\section{Conclusions and future work}
In this paper, we addressed the limitations of existing text-to-image retrieval systems, which are often constrained to RGB data and cannot leverage the full richness of modern satellite sensors. We introduced CrisisLandMark, a new, large-scale corpus of over 647,000 Sentinel-1 (SAR) and Sentinel-2 (multispectral) images. The corpus's key innovation is its structured textual annotations, which are harmonized from authoritative land cover systems (CORINE and Dynamic World) and enriched with crisis-event tags. To leverage this resource, we developed CLOSP, a novel contrastive architecture that aligns text with both optical and SAR data. It uses text as a common ``bridge'' to create a unified semantic space from unpaired multisensor imagery, solving a fundamental challenge in satellite data fusion.

Our experiments led to three principal findings. First, by moving beyond the visible spectrum, the CLOSP framework significantly outperforms state-of-the-art baselines, proving that the rich information in SAR and multispectral data is crucial for retrieval. Second, we provide strong evidence for cross-modal knowledge transfer: the unified training strategy allows semantic concepts learned from optical data to improve the interpretation and retrieval of challenging SAR imagery. Finally, our work uncovers a fundamental and practical trade-off between semantic and geographic representation. While the baseline CLOSP model is a superior general-purpose semantic retriever, GeoCLOSP becomes a high-performing specialist for location-dependent queries, such as crisis events and rare geographic features.

This research opens several promising avenues for future work. The most critical next step is the integration of the temporal dimension, extending our framework to support time-aware queries and the analysis of dynamic environmental processes. Other promising directions include: expanding the CrisisLandMark corpus with more globally distributed data to mitigate geographic bias; exploring adaptive methods to dynamically balance the semantic-geographic trade-off within GeoCLOSP; and leveraging multimodal large language models to enable more descriptive annotations and queries.

Ultimately, this work provides both a powerful framework and a valuable resource for building the next generation of sensor-agnostic retrieval systems, making the wealth of information in global Earth observation archives more accessible and actionable than ever before.

\newpage

\section*{Code availability}
The source code is available for downloading under the Apache-2.0 license at the link: \url{https://github.com/DarthReca/closp}.

\section*{Data availability}
The data are available at \url{https://huggingface.co/datasets/DarthReca/crisislandmark}.

\section*{Declaration of competing interest}
The authors declare that they have no known competing financial interests or personal relationships that could have appeared to influence the work reported in this paper.

\appendix

\section{Analysis of Loss Weighting for GeoCLOSP}
\label{sec:loss_weighting}
To analyze the balance between the semantic ($L_{txt-img}$) and geographic ($L_{img-loc}$) loss components, for simplicity, we can rewrite the loss shown in \Cref{eq:total_geo_loss} as $L_g = \alpha L_{txt-img} + (1 - \alpha) L_{img-loc}$, where $\alpha = 0.5$ in this specific case. We tested $\alpha$ values of 0.25 and 0.75, comparing them to the baseline CLOSP-RN, which corresponds to $\alpha = 1$, and GeoCLOSP ($\alpha = 0.5$). The case where $\alpha=0$ (equivalent to SatCLIP) is omitted from the performance comparison as it discards the text query entirely, making it unsuitable for text-to-image retrieval. The results, reported in \Cref{tab:loss_weighting_analysis}, show that an equal weighting of $\alpha=0.5$ achieves the highest nDCG@1000 score of 57.76. This result outperforms the semantics-only baseline ($\alpha=1$) and suggests that incorporating geographic context is beneficial. Performance degrades as $\alpha$ shifts heavily towards either extreme, indicating that an even balance provides the optimal trade-off between semantic relevance and location-based alignment for this task.

\begin{table}[pos=h]
    \centering
    \caption{Impact of the loss weighting parameter $\alpha$ on text-to-image retrieval performance, measured by nDCG@1000. The parameter $\alpha$ balances the semantic loss ($L_{txt-img}$) and the geographic loss ($L_{img-loc}$).}
    \label{tab:loss_weighting_analysis}
    \begin{threeparttable}
    \begin{tabular}{l c c}
    \toprule
    \textbf{Configuration} & \textbf{$\alpha$ value} & \textbf{nDCG@1000} \\
    \midrule
    CLOSP-RN (Semantics-only) & 1.00 & 56.23 \\
    GeoCLOSP (High Semantics) & 0.75 & 55.18 \\
    \textbf{GeoCLOSP (Balanced)} & \textbf{0.50} & \textbf{57.76} \\
    GeoCLOSP (High Geography) & 0.25 & 52.82 \\
    SatCLIP (Geography-only) & 0.00 & N/A\tnote{*} \\
    \bottomrule
    \end{tabular}
    \begin{tablenotes}
        \item[*] It discards the text alignment, making the model unable to perform the text-to-image retrieval task.
    \end{tablenotes}
    \end{threeparttable}
\end{table}

\section{Visual Examples of Retrieval}
\label{sec:visual_examples}
\begin{figure}
    \centering
    \begin{tabular}{l|ccc|ccc}
     CLOSP-RN & 10 & 5 & 10 & 3 & 10 & 5\\
     &
     \includegraphics[width=0.1\linewidth]{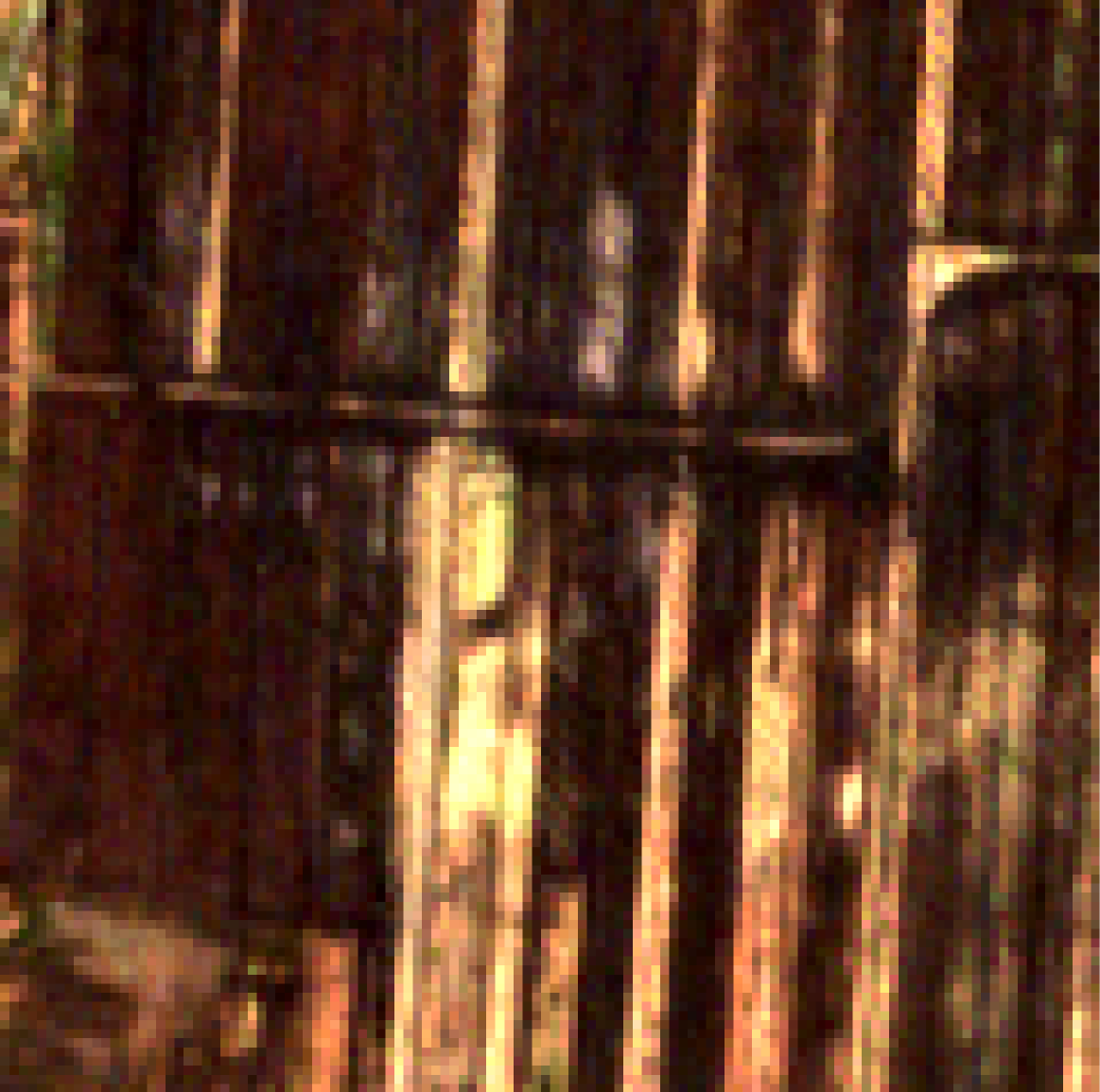}
     &
     \includegraphics[width=0.1\linewidth]{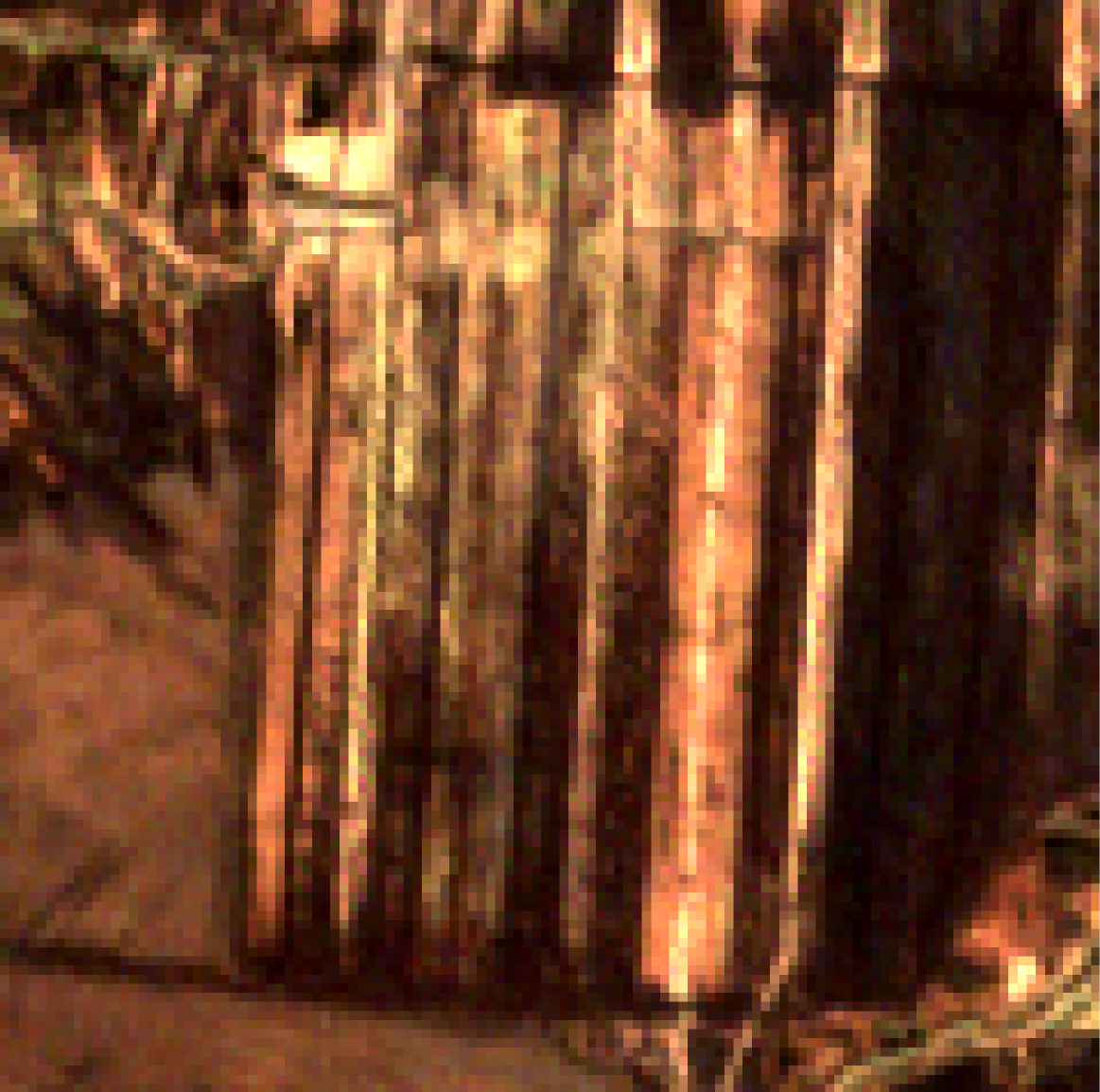}
     &
     \includegraphics[width=0.1\linewidth]{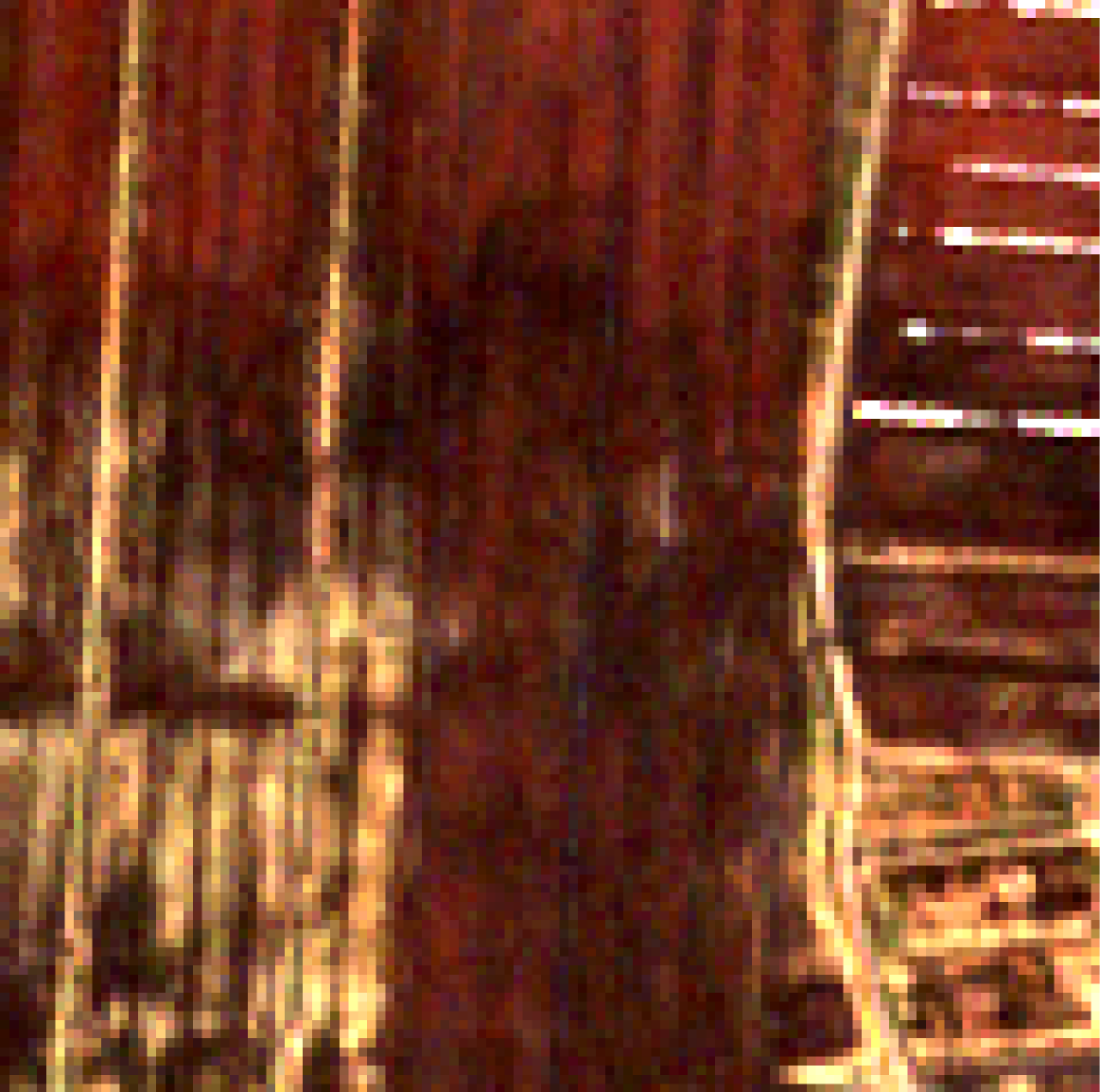}
     &
     \includegraphics[width=0.1\linewidth]{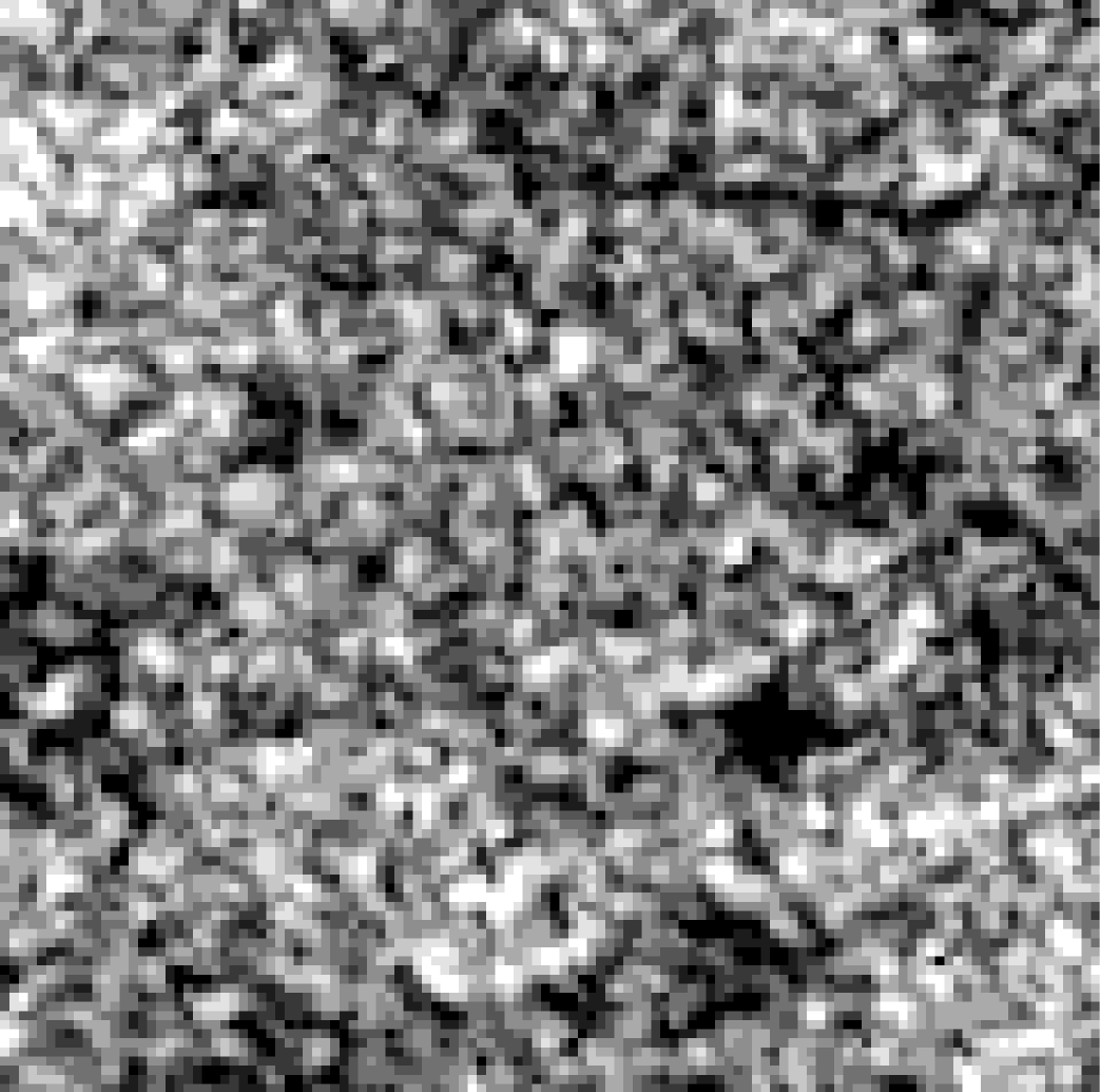}
     &
     \includegraphics[width=0.1\linewidth]{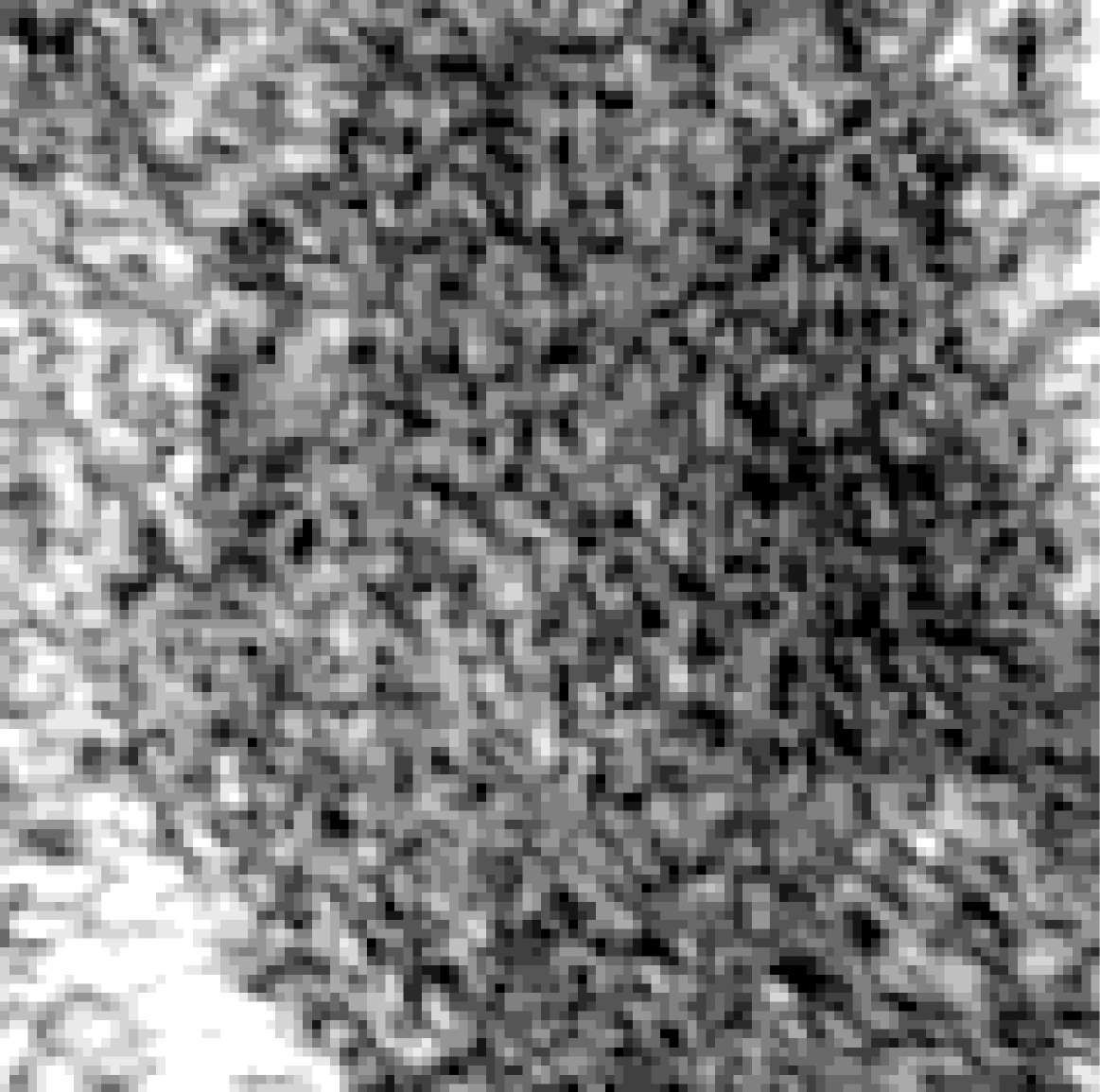}
     &
     \includegraphics[width=0.1\linewidth]{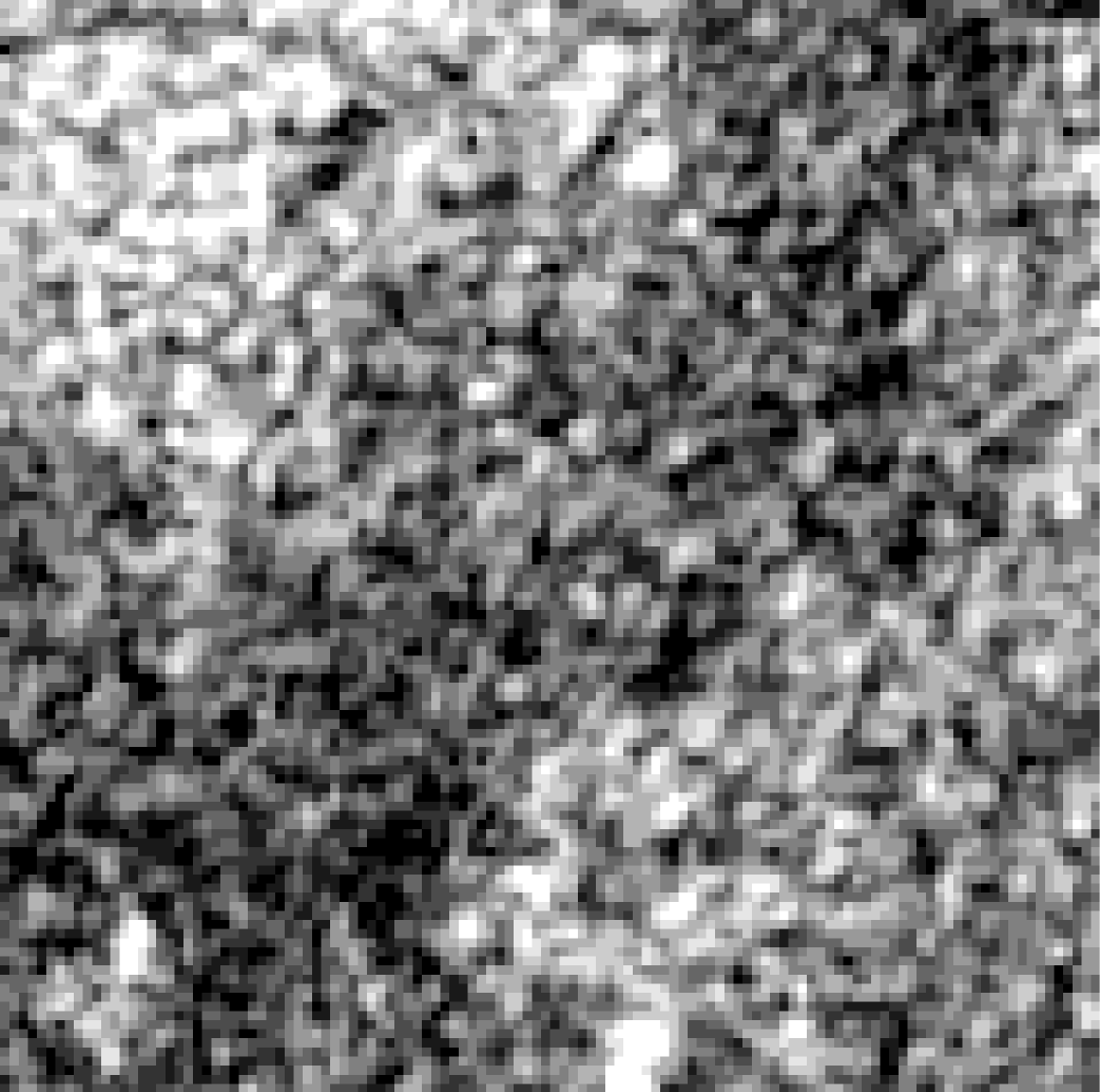}
     \\\midrule
     BiCLIP & 5 & 5 & 5 & 2 & 3 & 7\\
     &
     \includegraphics[width=0.1\linewidth]{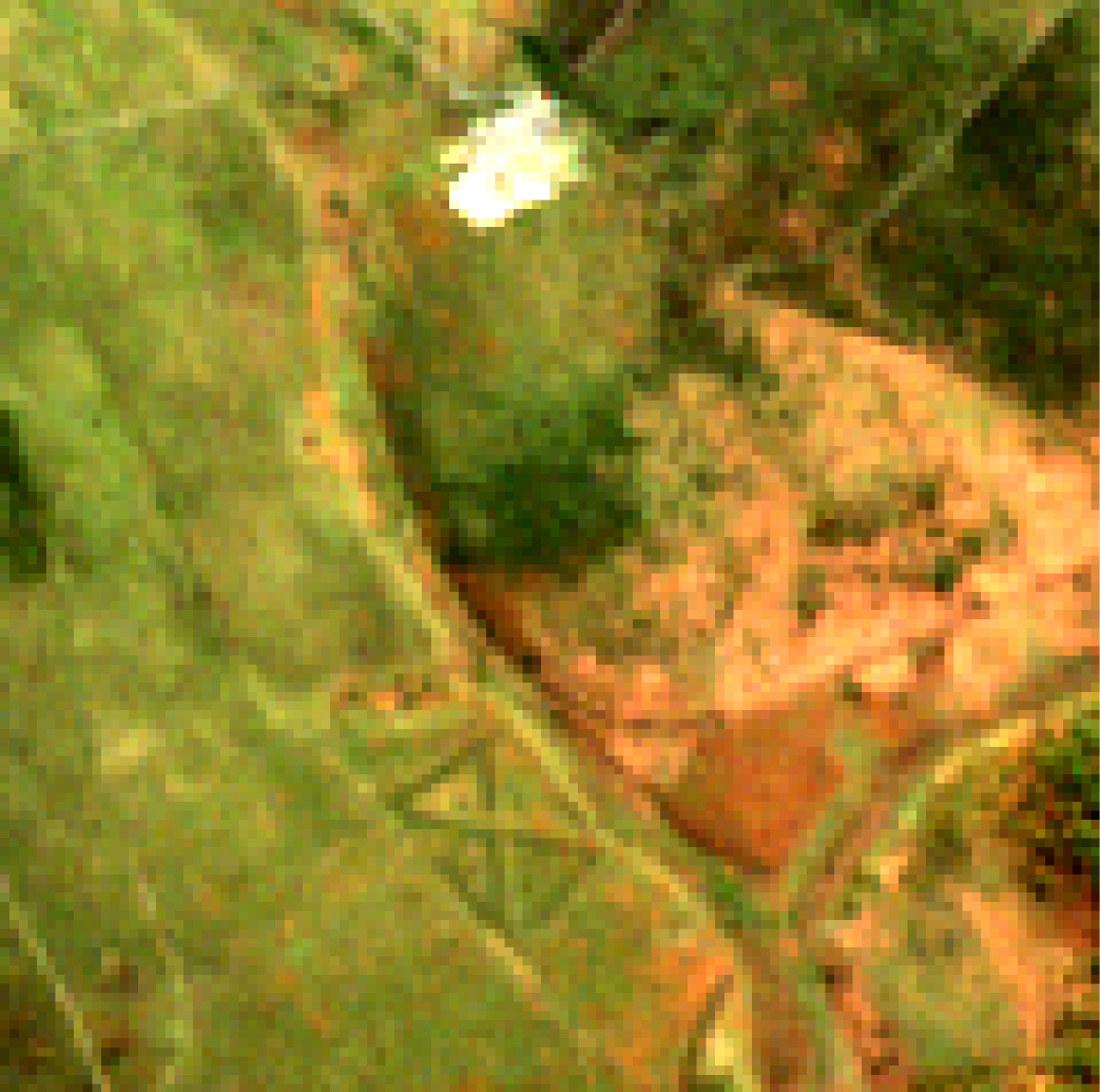}
     &
     \includegraphics[width=0.1\linewidth]{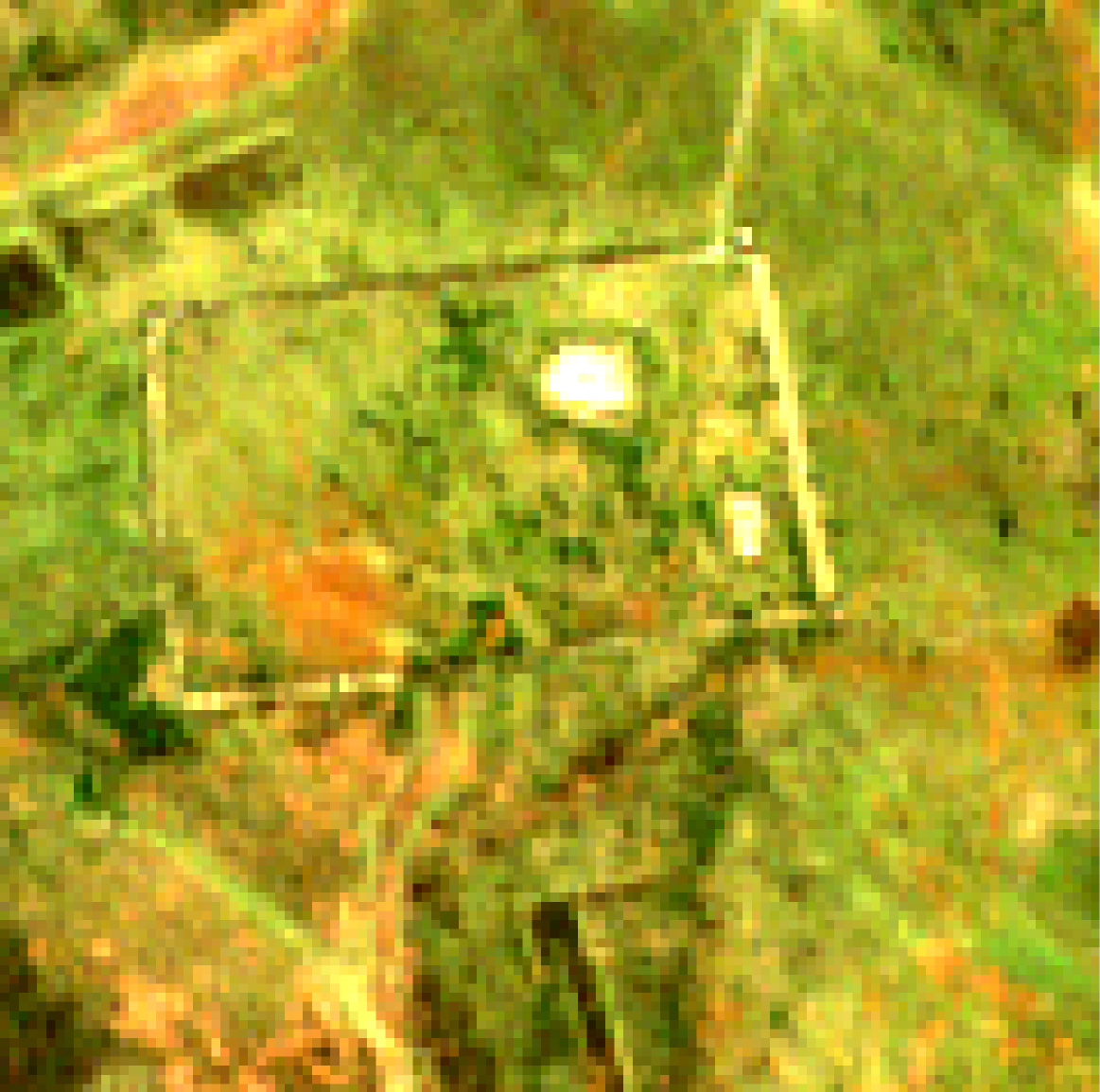}
     &
     \includegraphics[width=0.1\linewidth]{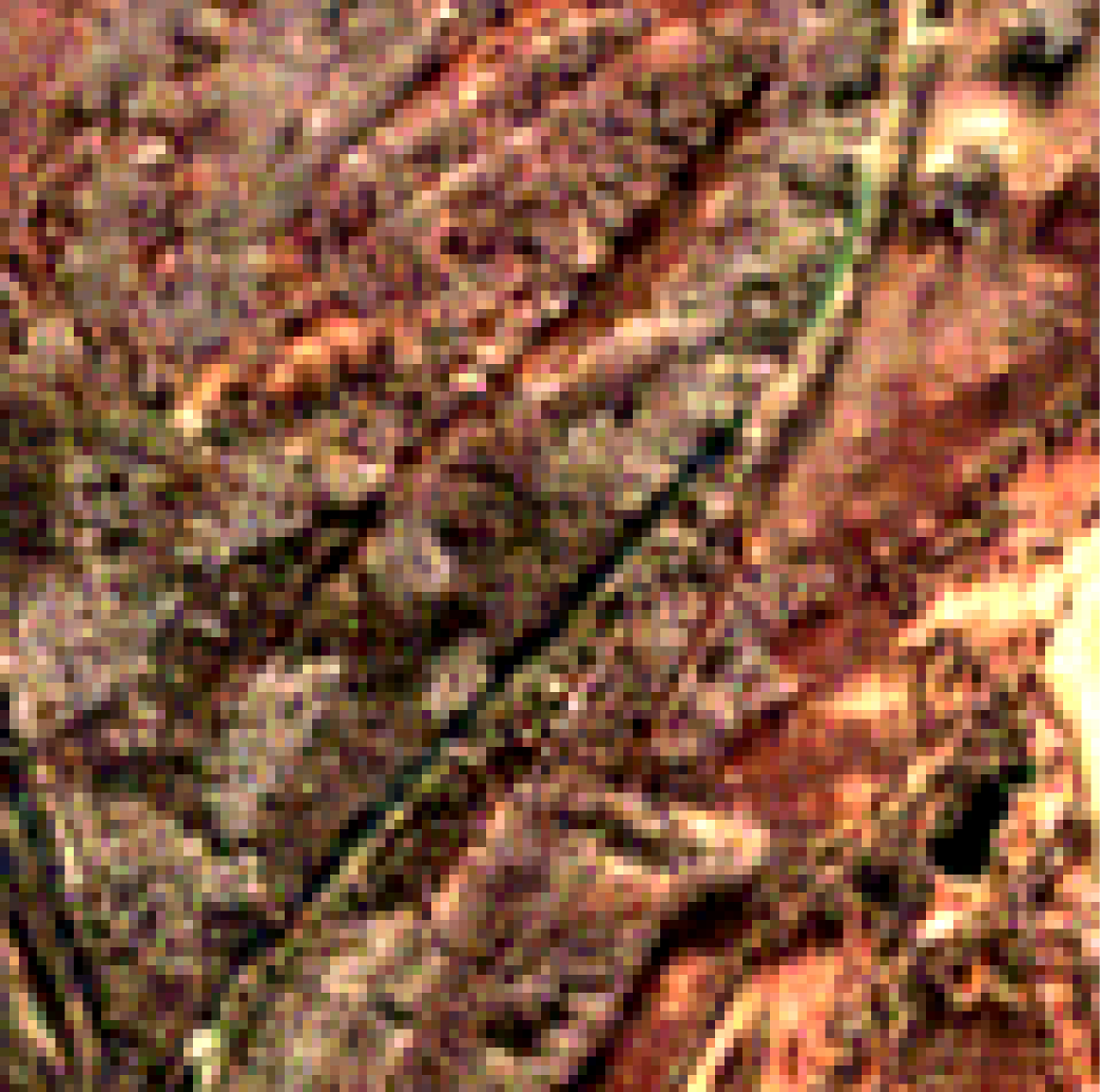}
     &
     \includegraphics[width=0.1\linewidth]{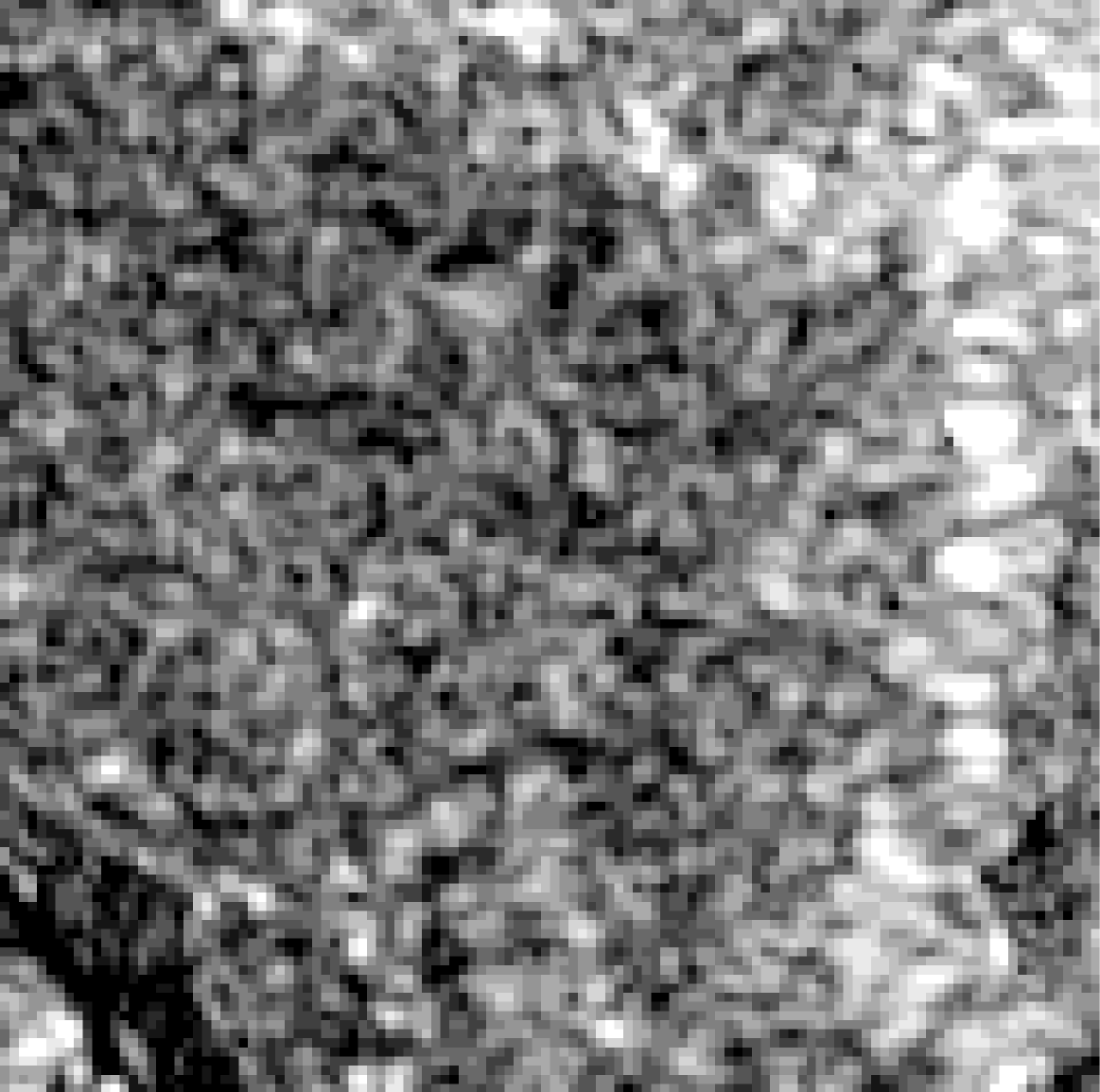}
     &
     \includegraphics[width=0.1\linewidth]{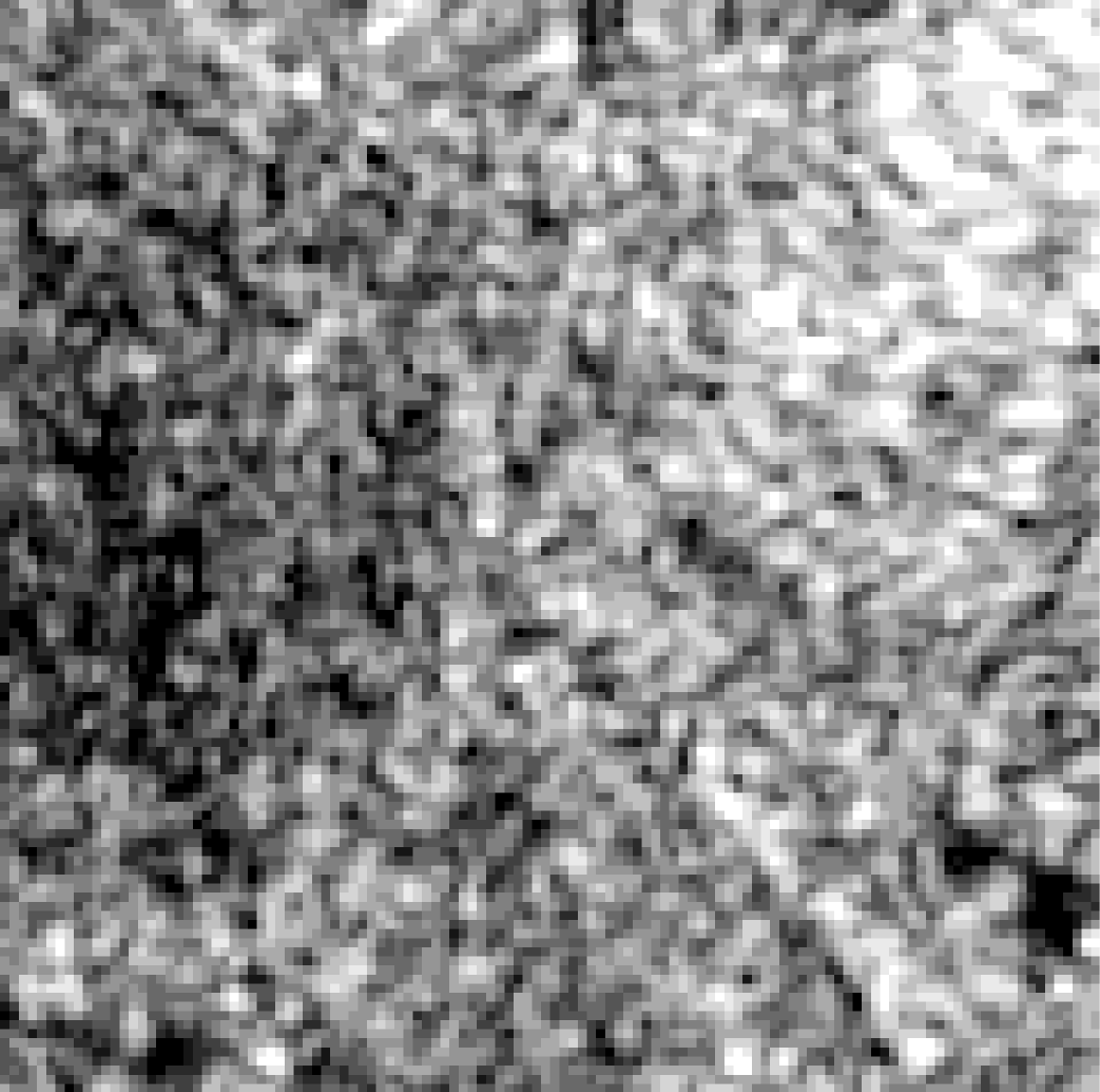}
     &
     \includegraphics[width=0.1\linewidth]{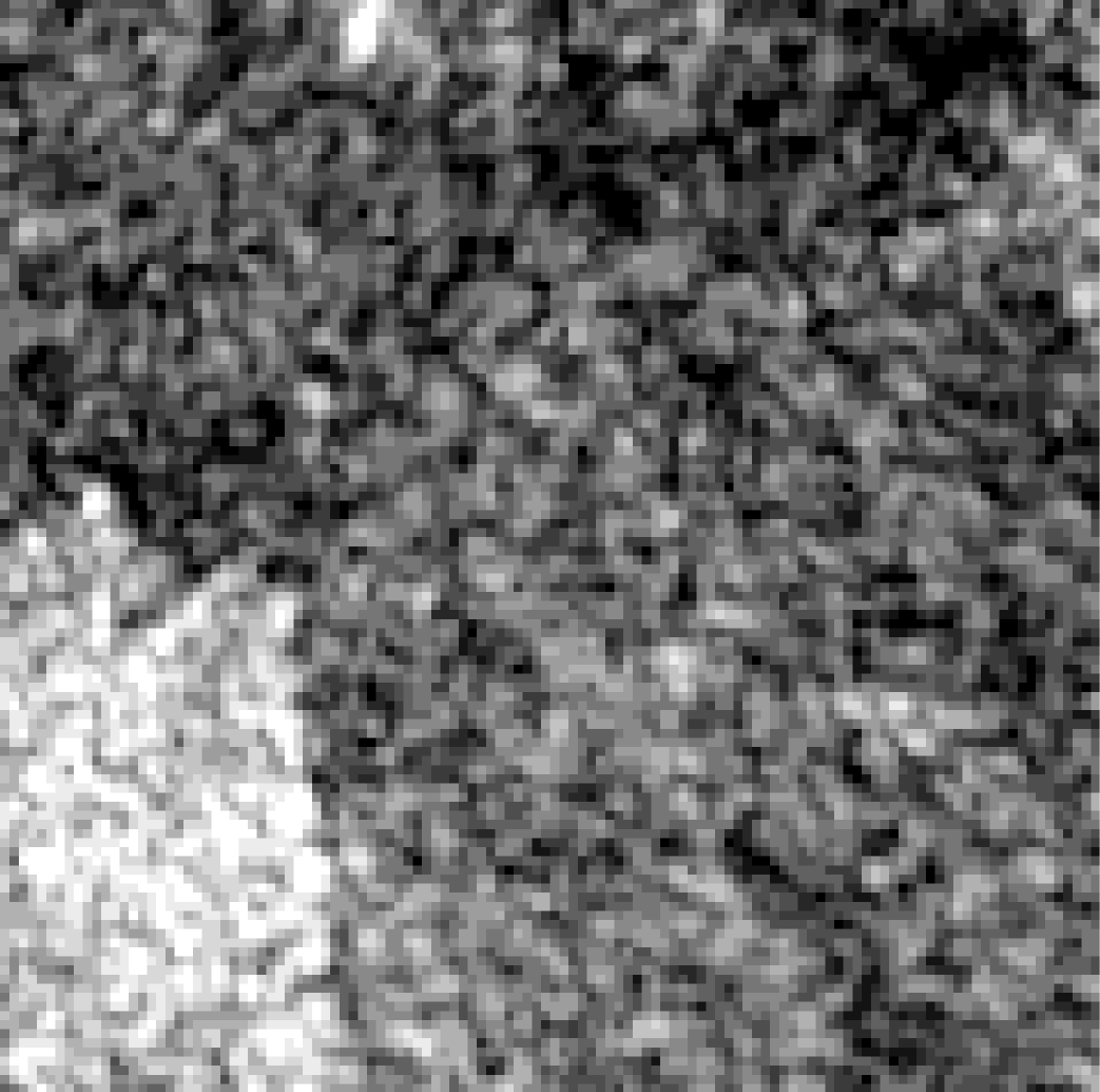}
     \\\midrule
     SkyCLIP-T & 3 & 3 & 3 & - & - & - \\
     &
     \includegraphics[width=0.1\linewidth]{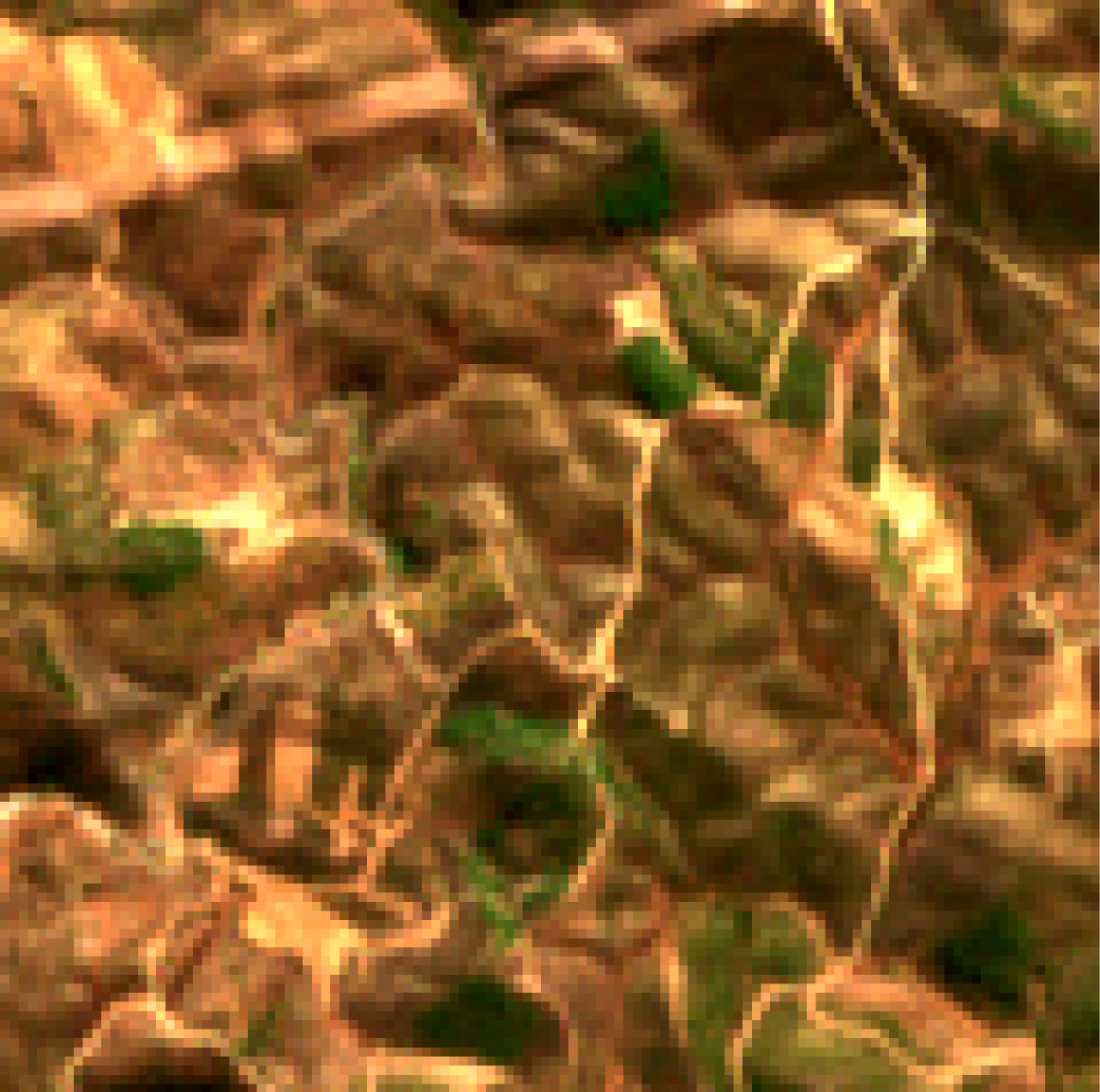}
     &
     \includegraphics[width=0.1\linewidth]{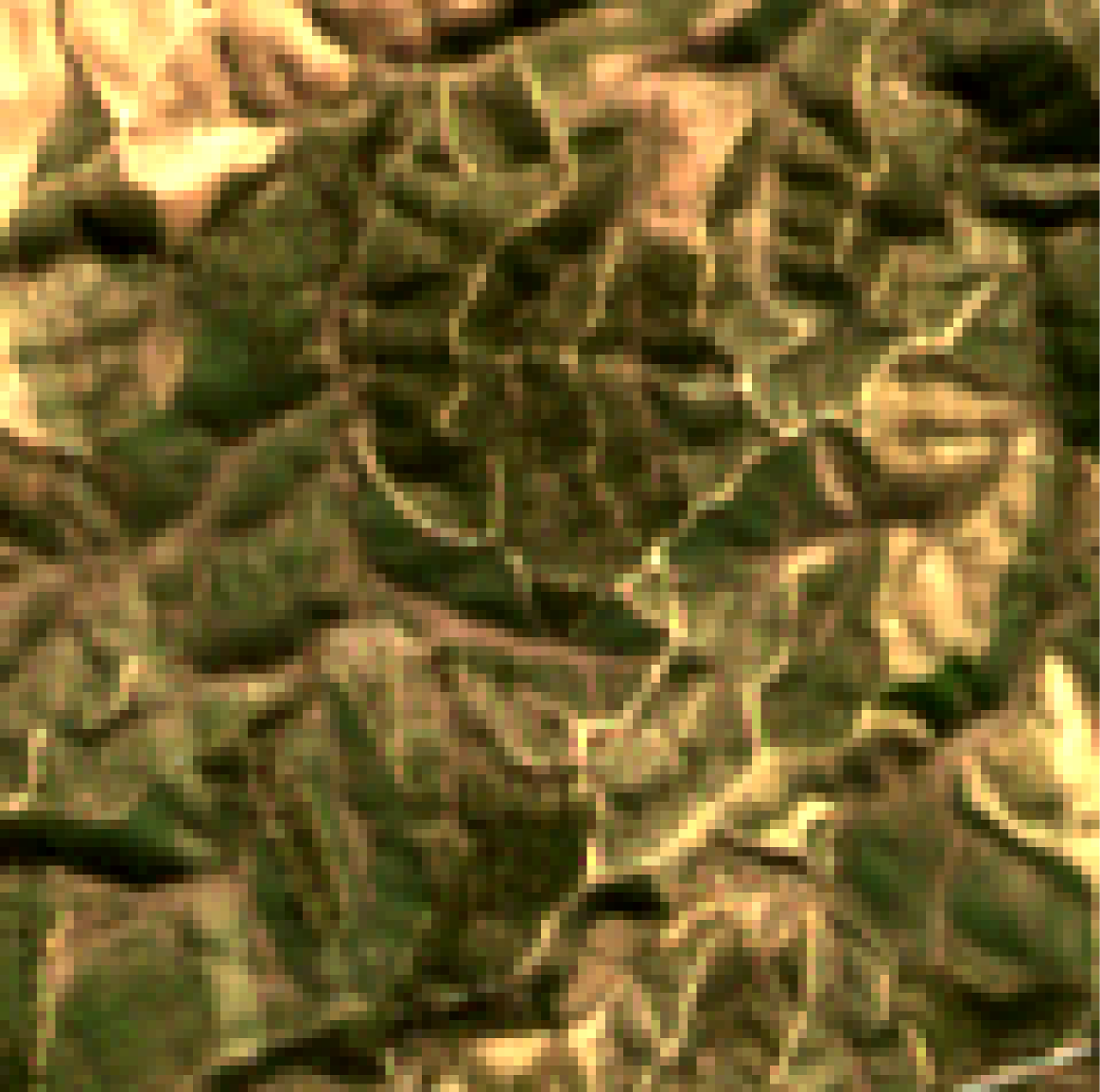}
     &
     \includegraphics[width=0.1\linewidth]{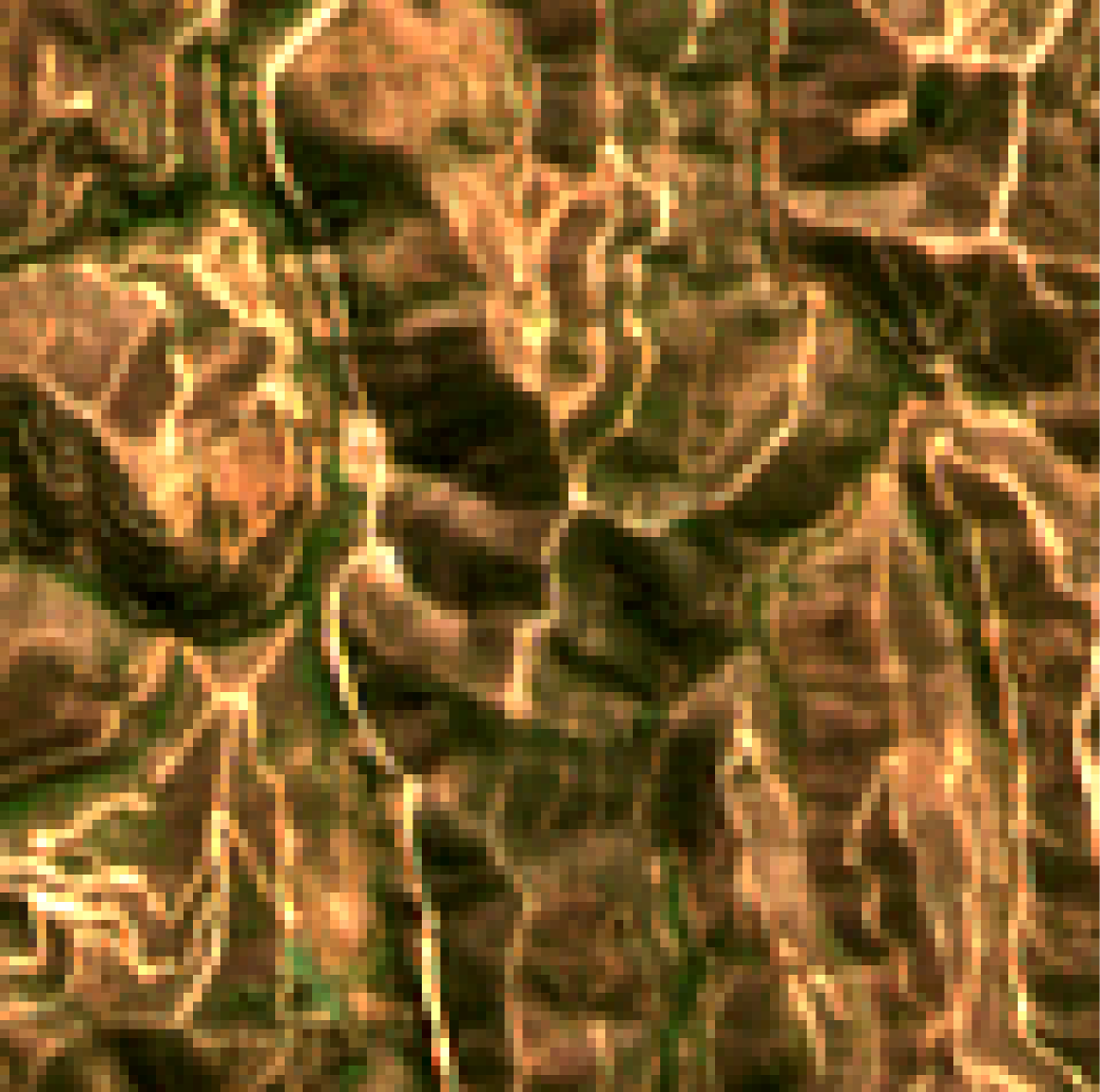}
     &
     &
     &
     \\\bottomrule
\end{tabular}

    \caption{Top-3 images by modality retrieved by the models for the provided query. We reported the RGB for Sentinel-2 and the VV polarization for Sentinel-1. Each image is coupled with the corresponding real relevance. Images are missing due to the use of approximated KNN in vector databases. \textbf{Query:} Flooded vegetation. Shrub and scrub}
    \label{fig:query_35}
\end{figure}

\begin{figure}
    \centering
    \begin{tabular}{l|ccc|ccc}
     CLOSP-RN & 8 & 8 & 3 & 8 & 5 & 6\\
     &
     \includegraphics[width=0.1\linewidth]{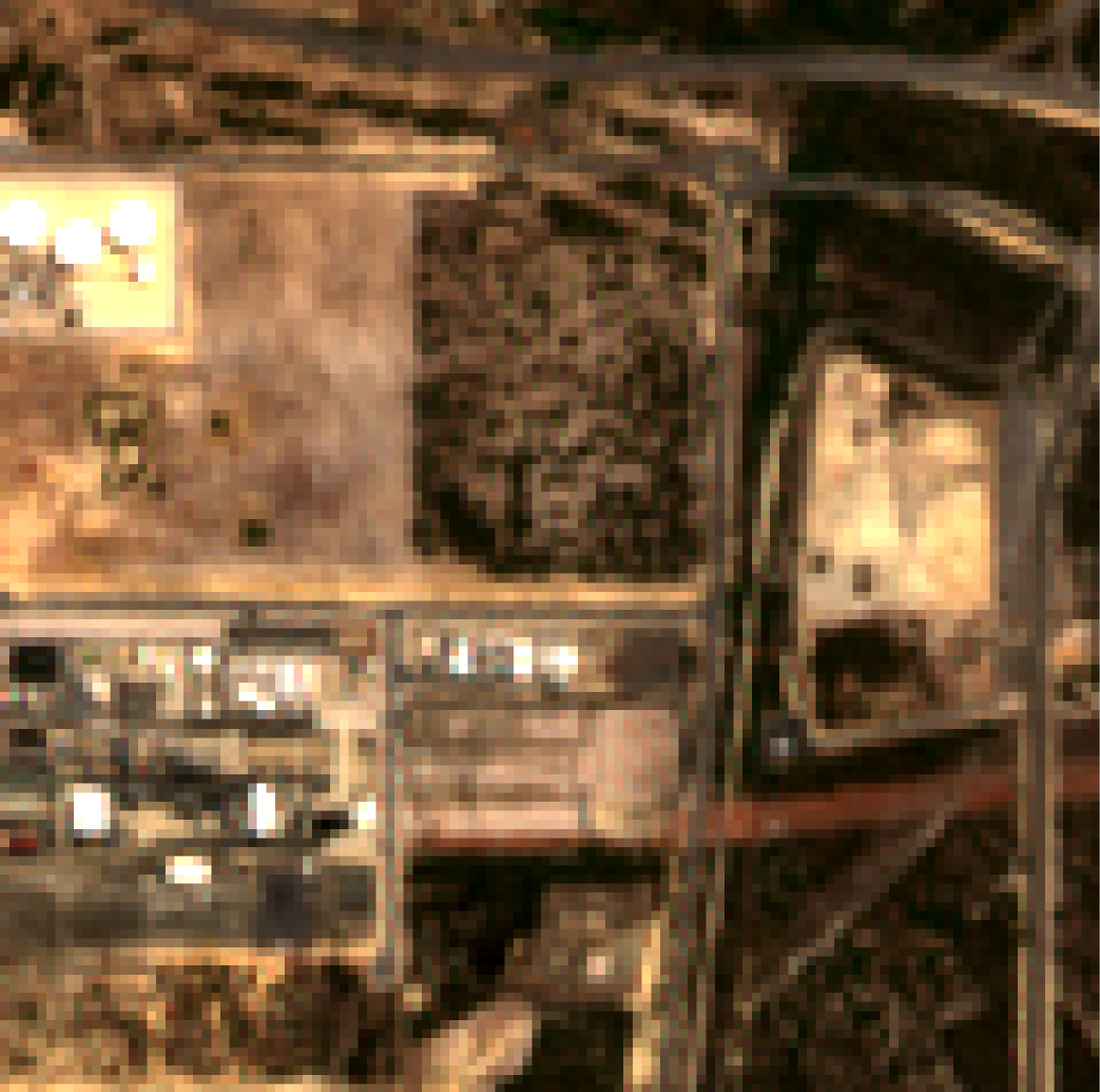}
     &
     \includegraphics[width=0.1\linewidth]{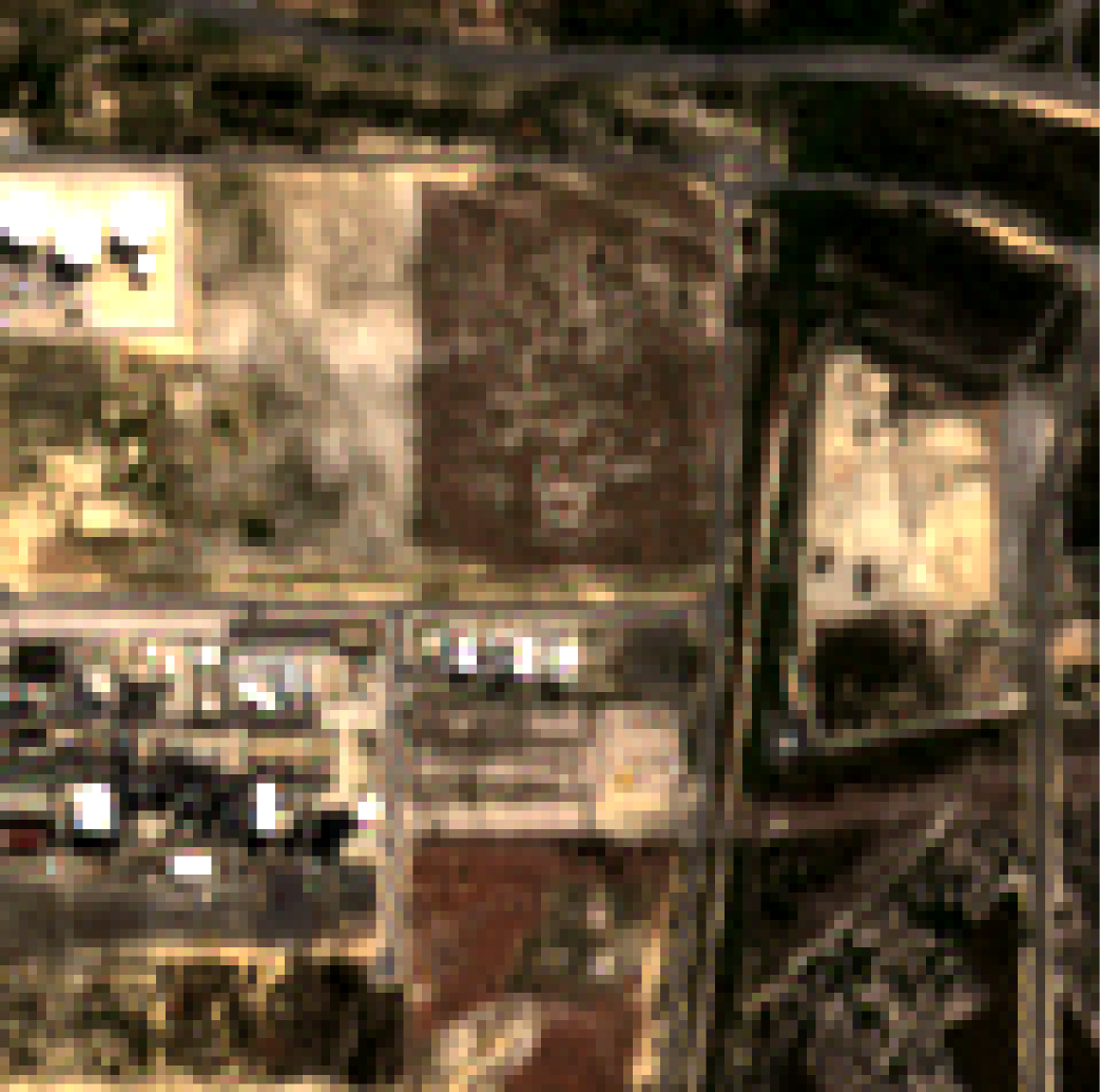}
     &
     \includegraphics[width=0.1\linewidth]{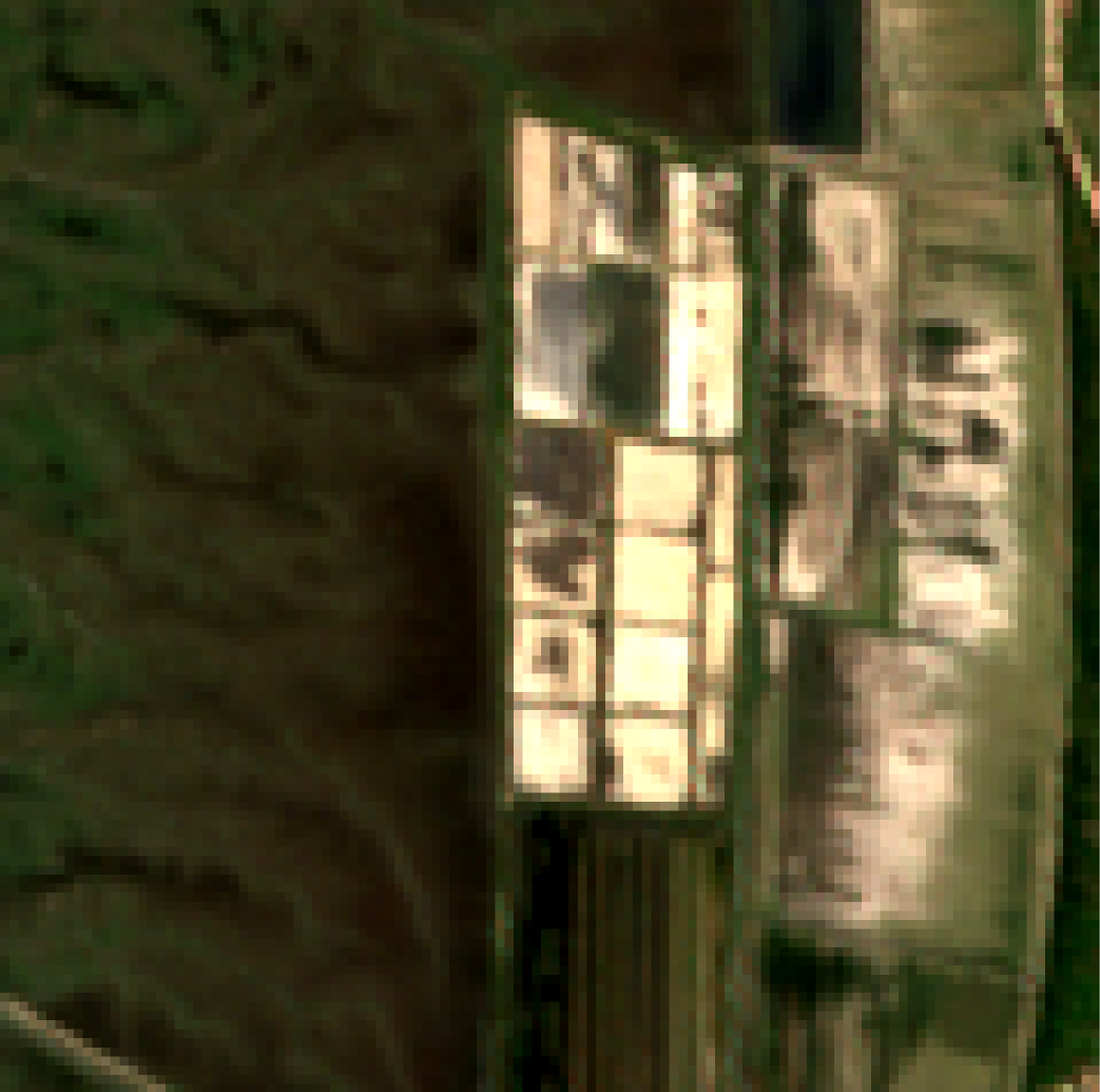}
     &
     \includegraphics[width=0.1\linewidth]{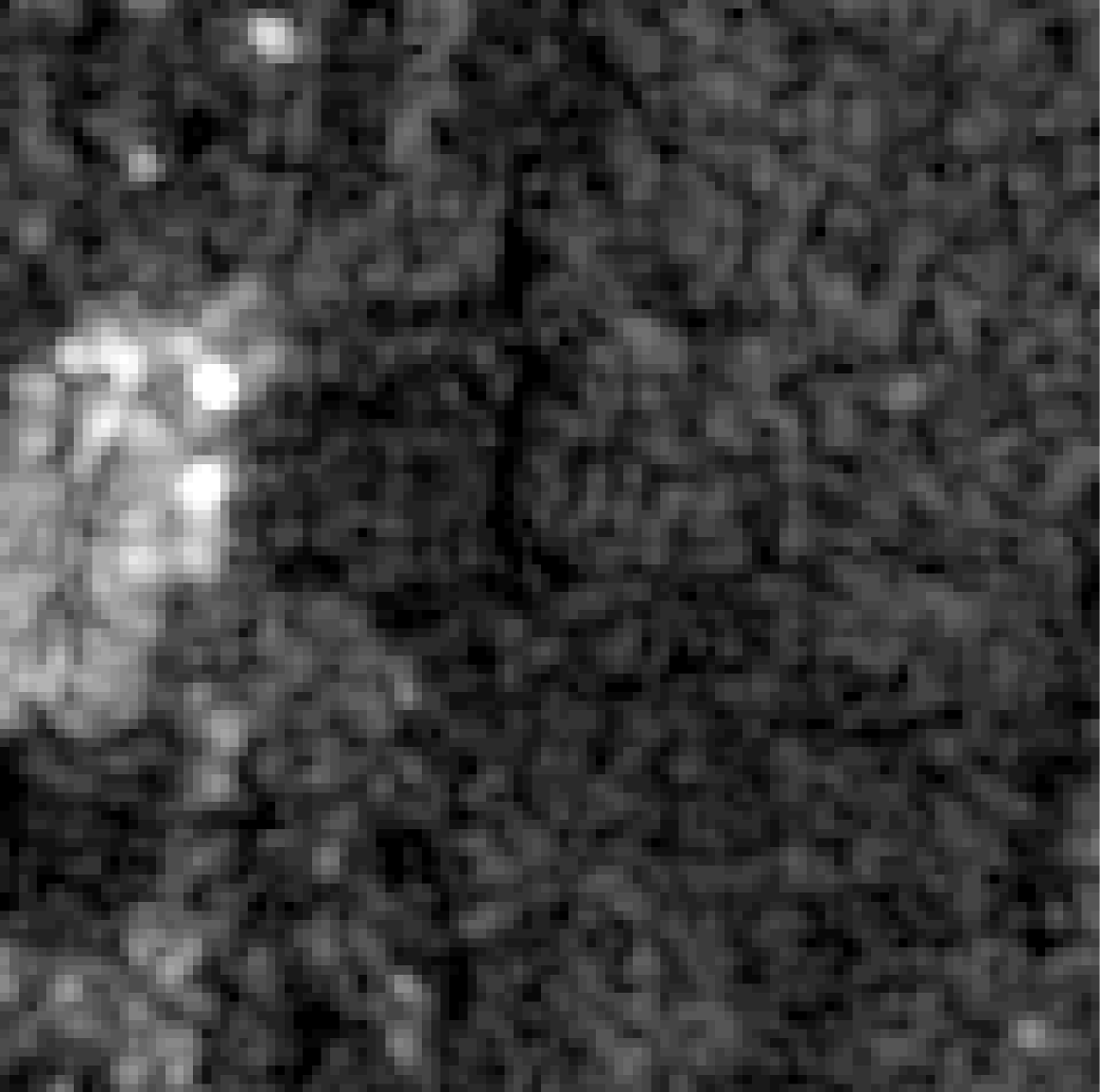}
     &
     \includegraphics[width=0.1\linewidth]{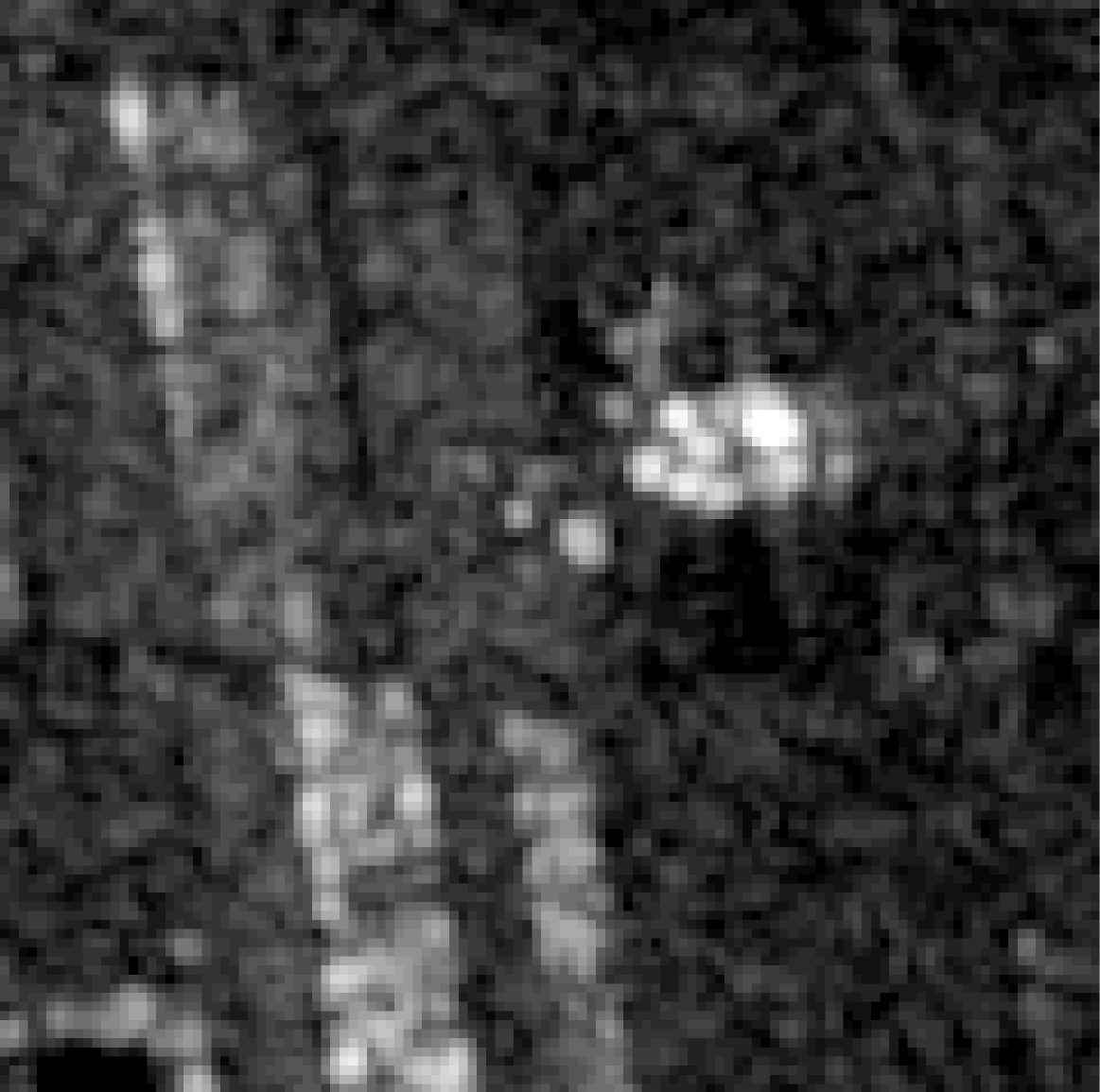}
     &
     \includegraphics[width=0.1\linewidth]{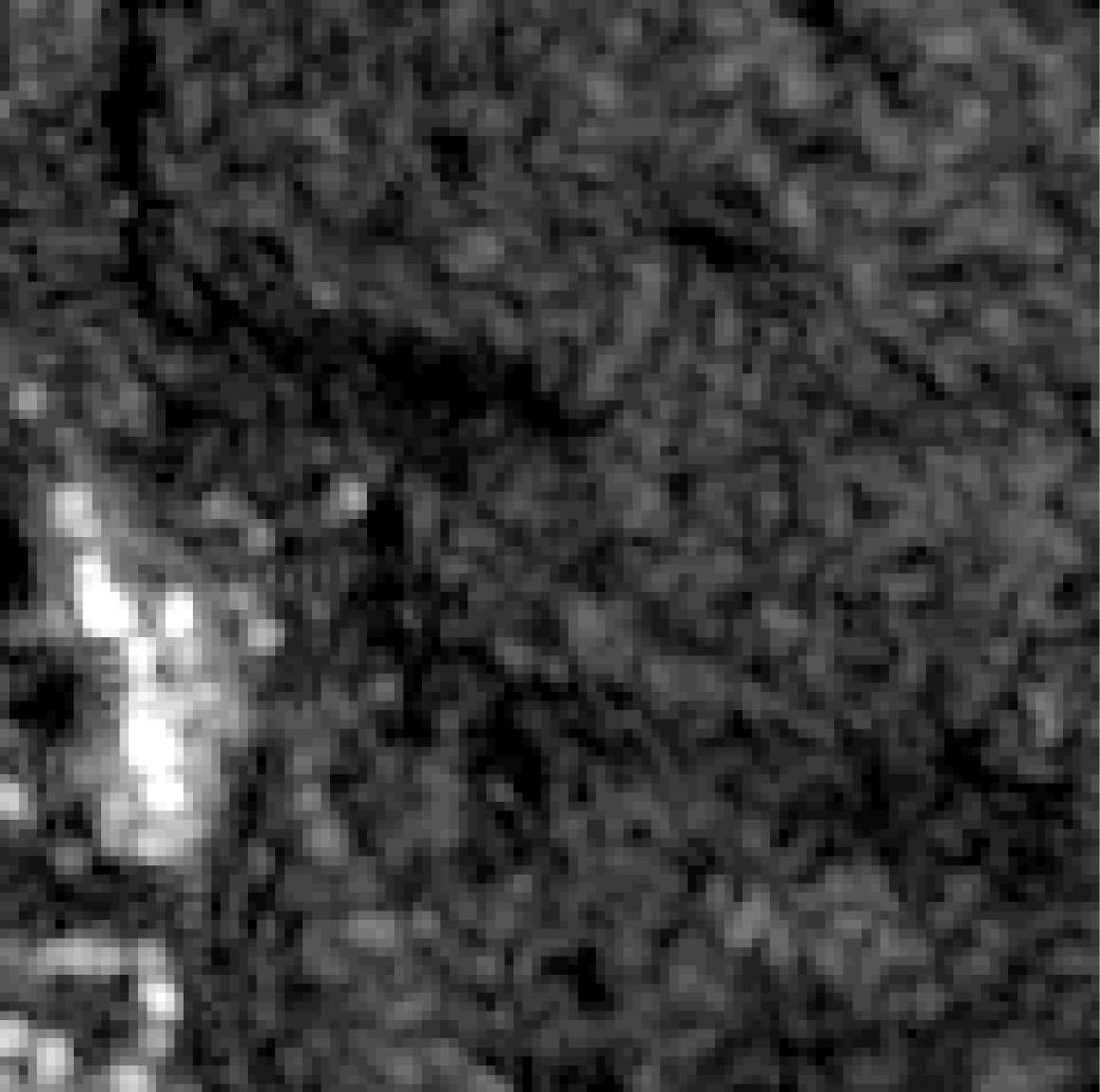}
     \\\midrule
     BiCLIP & 4 & 8 & 5 & 4 & 6 & 8\\
     &
     \includegraphics[width=0.1\linewidth]{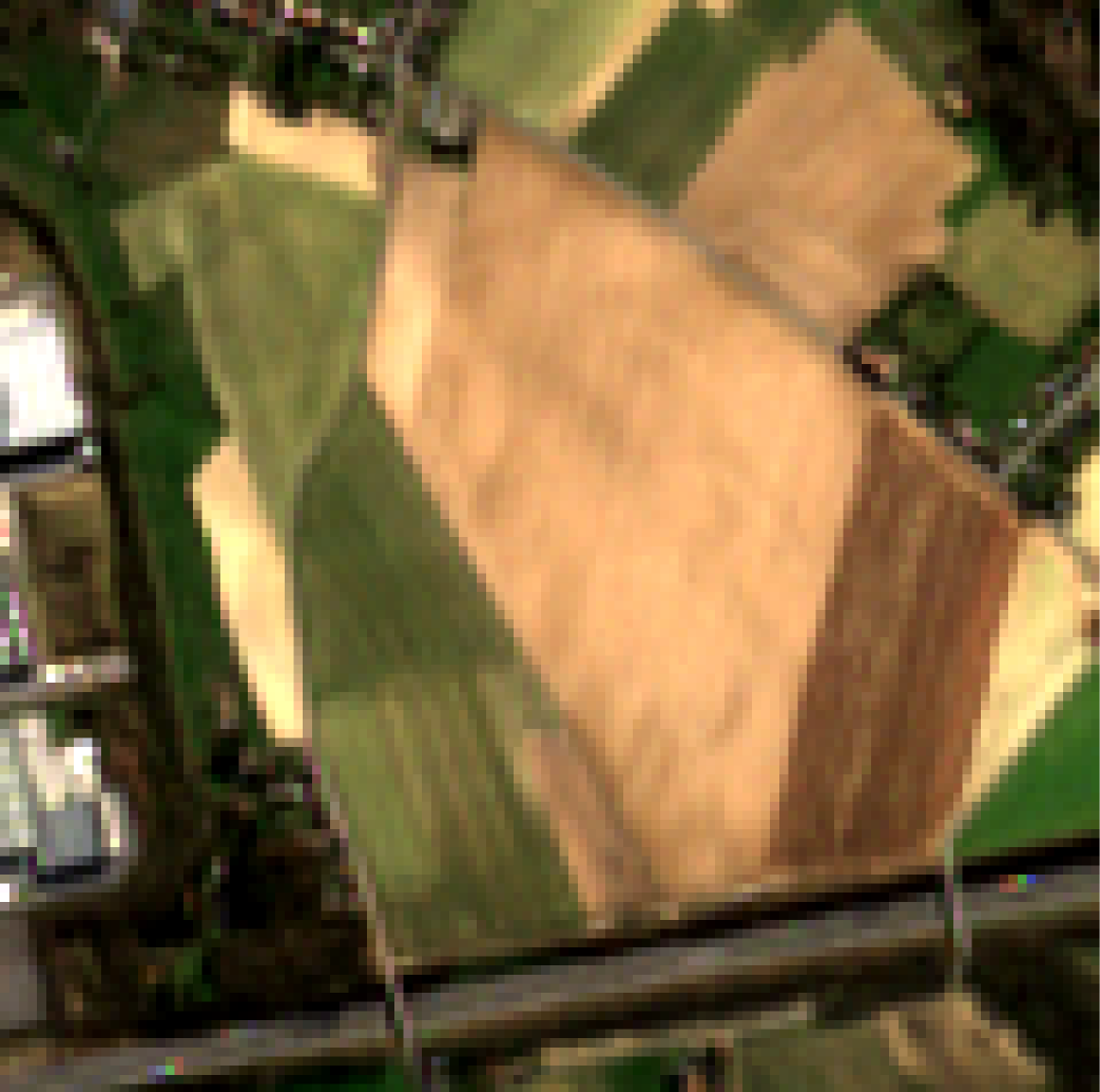}
     &
     \includegraphics[width=0.1\linewidth]{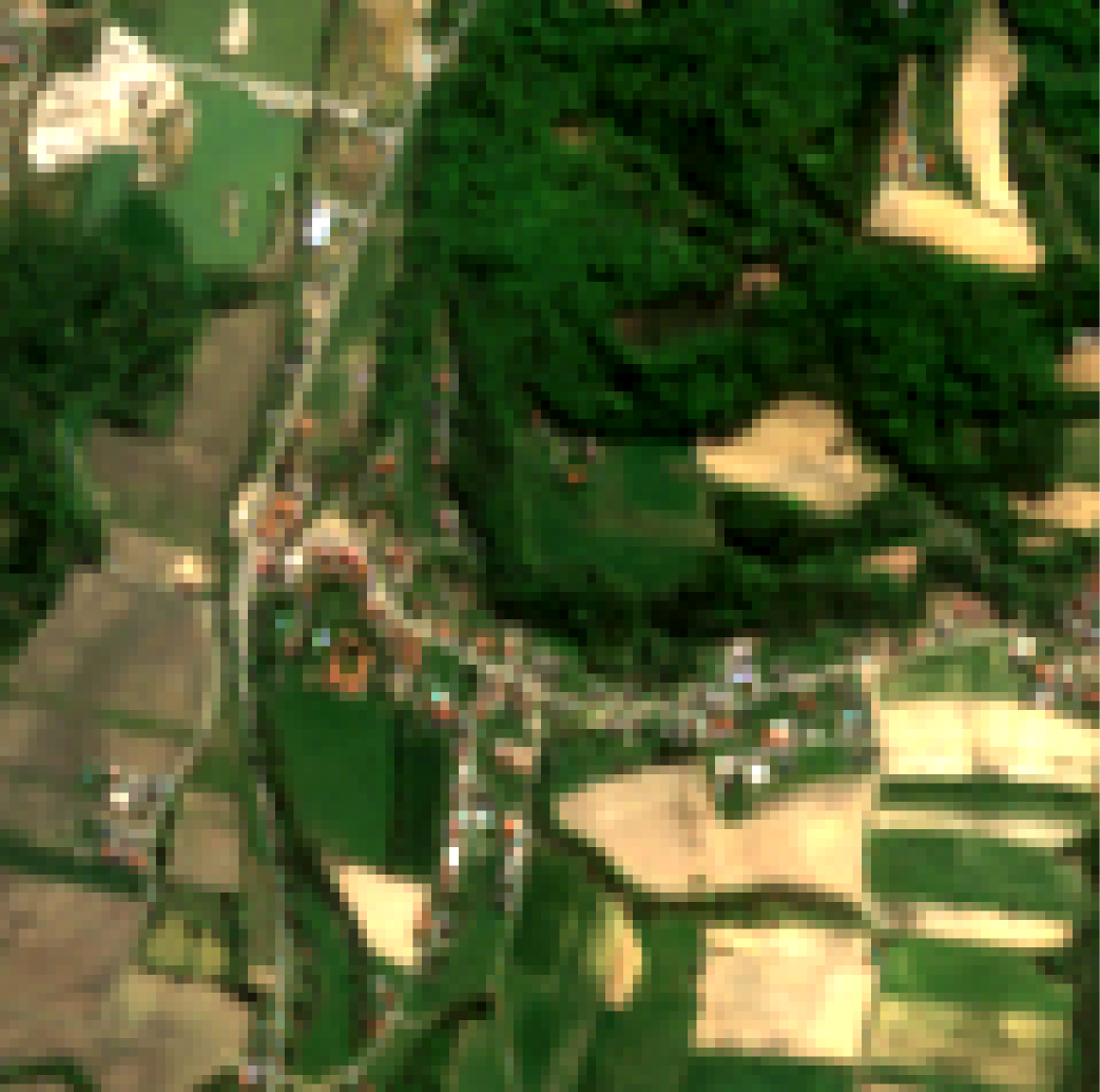}
     &
     \includegraphics[width=0.1\linewidth]{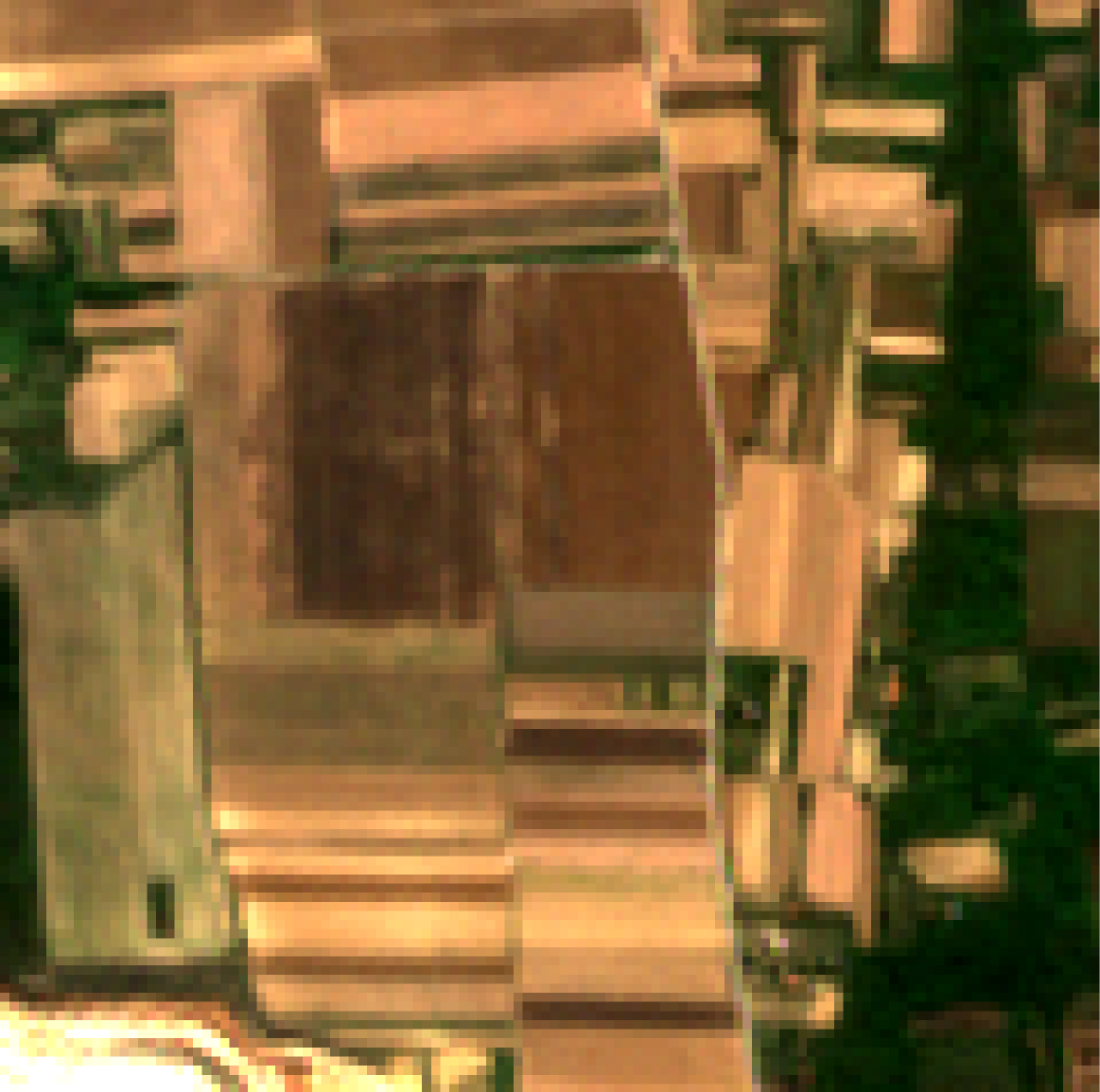}
     &
     \includegraphics[width=0.1\linewidth]{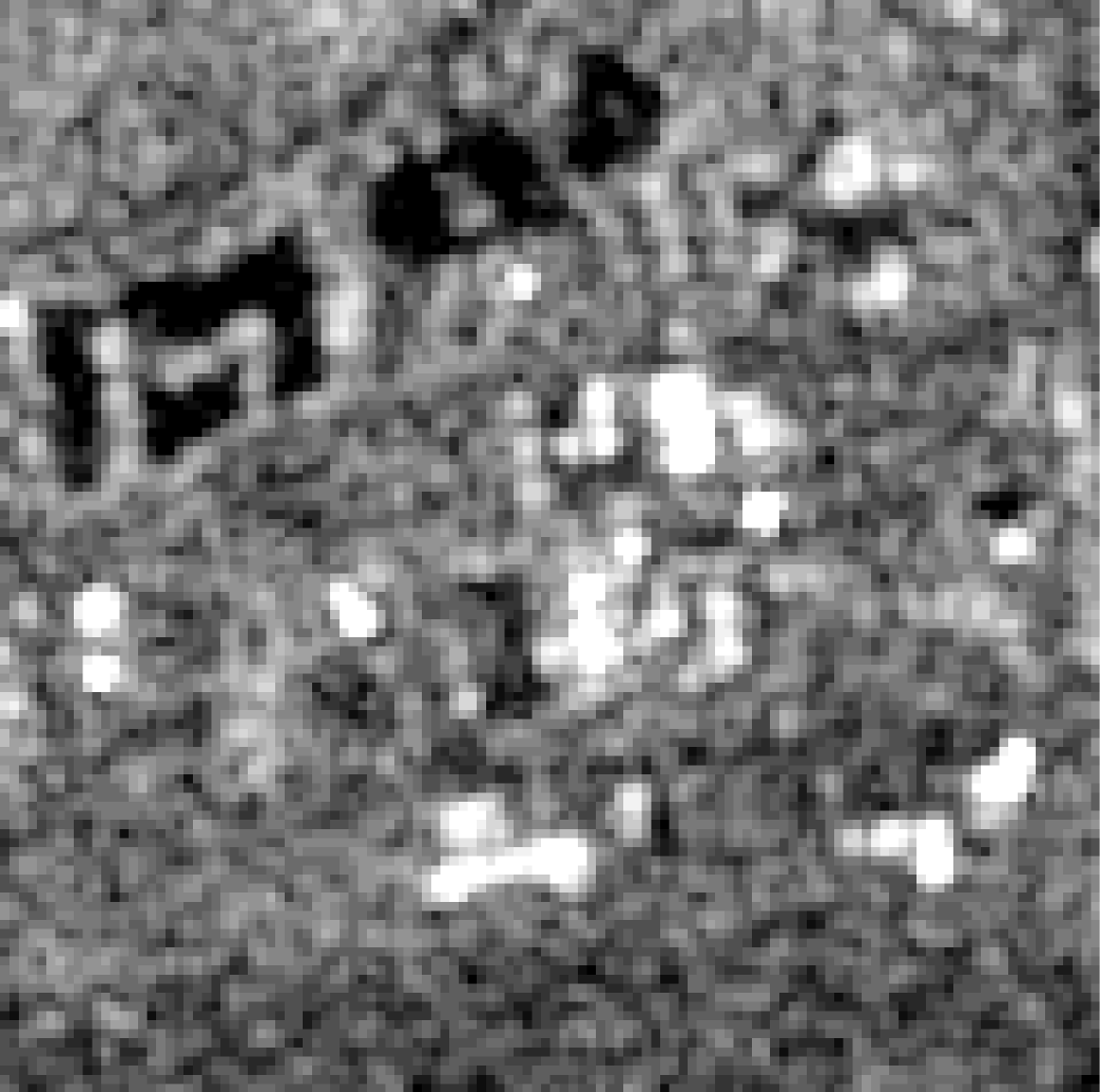}
     &
     \includegraphics[width=0.1\linewidth]{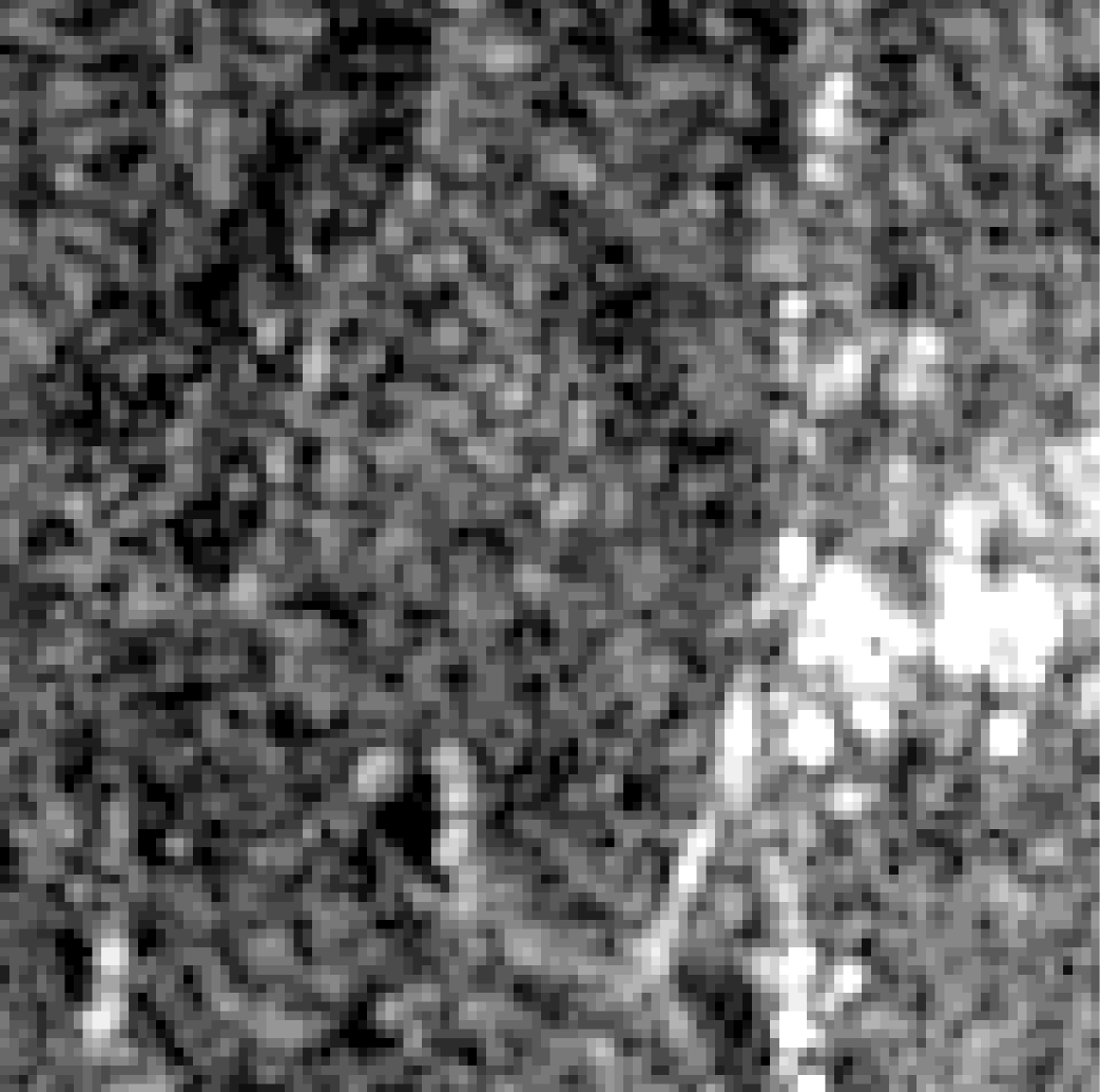}
     &
     \includegraphics[width=0.1\linewidth]{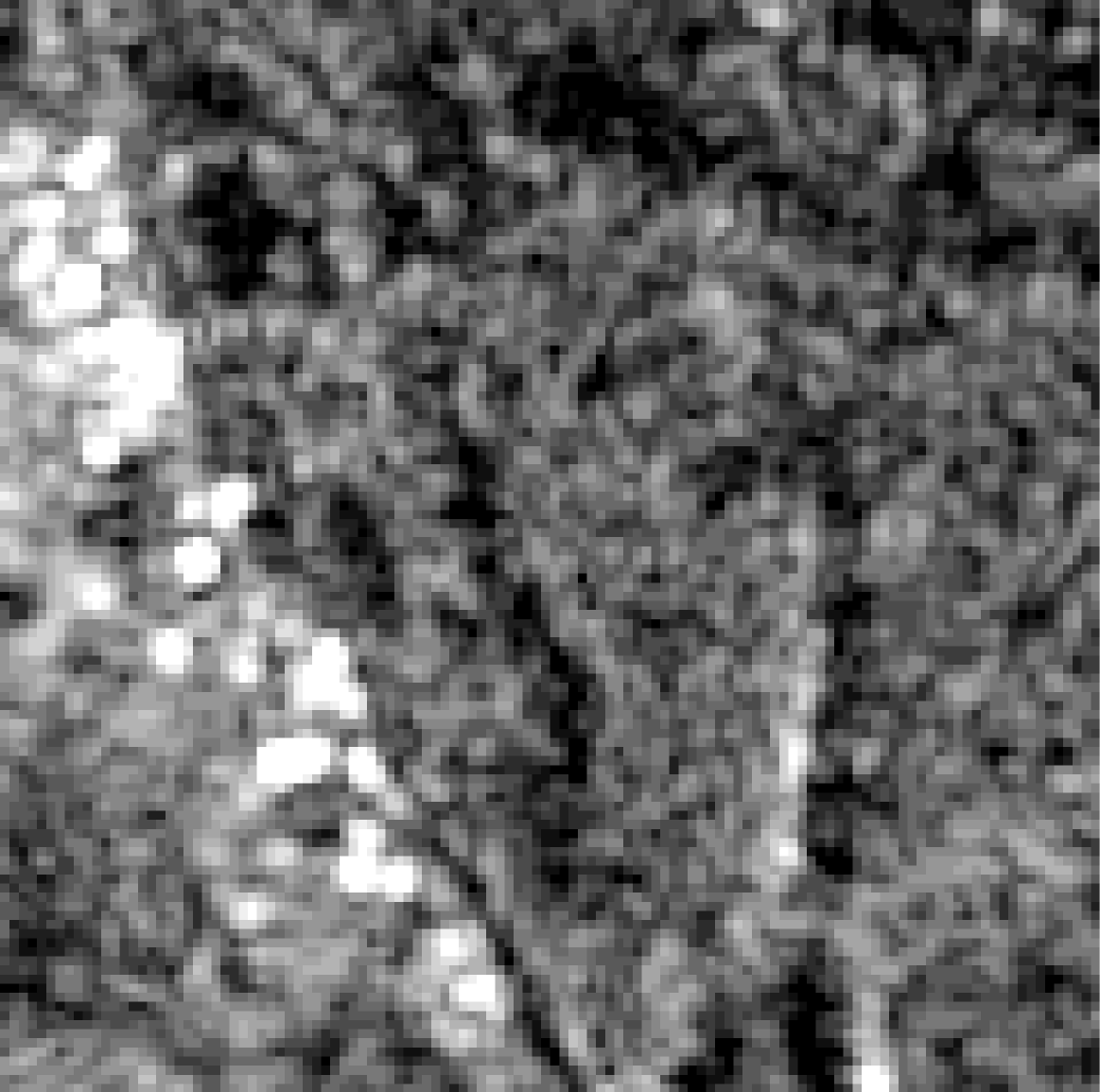}
     \\\midrule
     SkyCLIP-T & 6 & 2 & 4 & - & - & - \\
     &
     \includegraphics[width=0.1\linewidth]{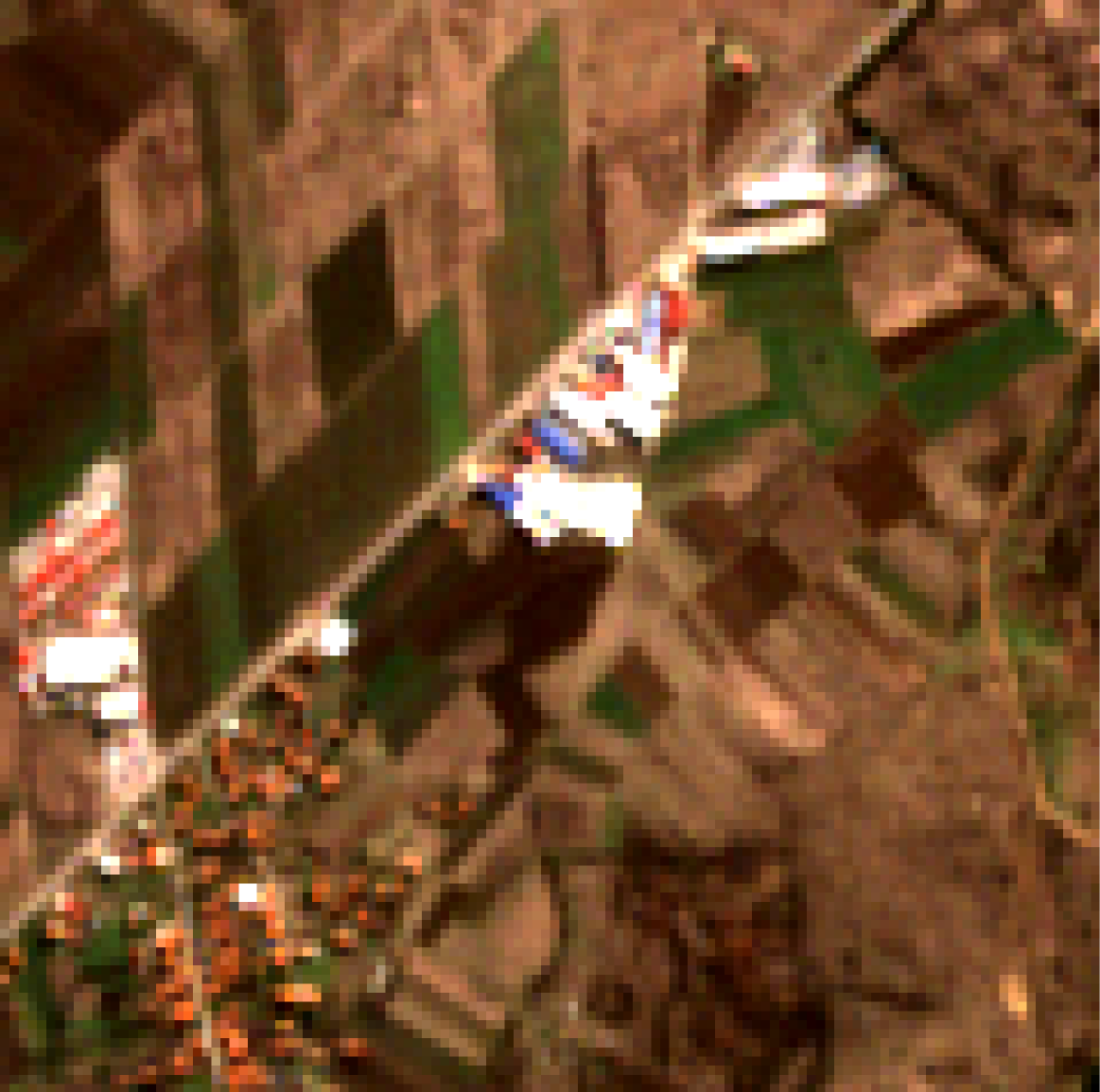}
     &
     \includegraphics[width=0.1\linewidth]{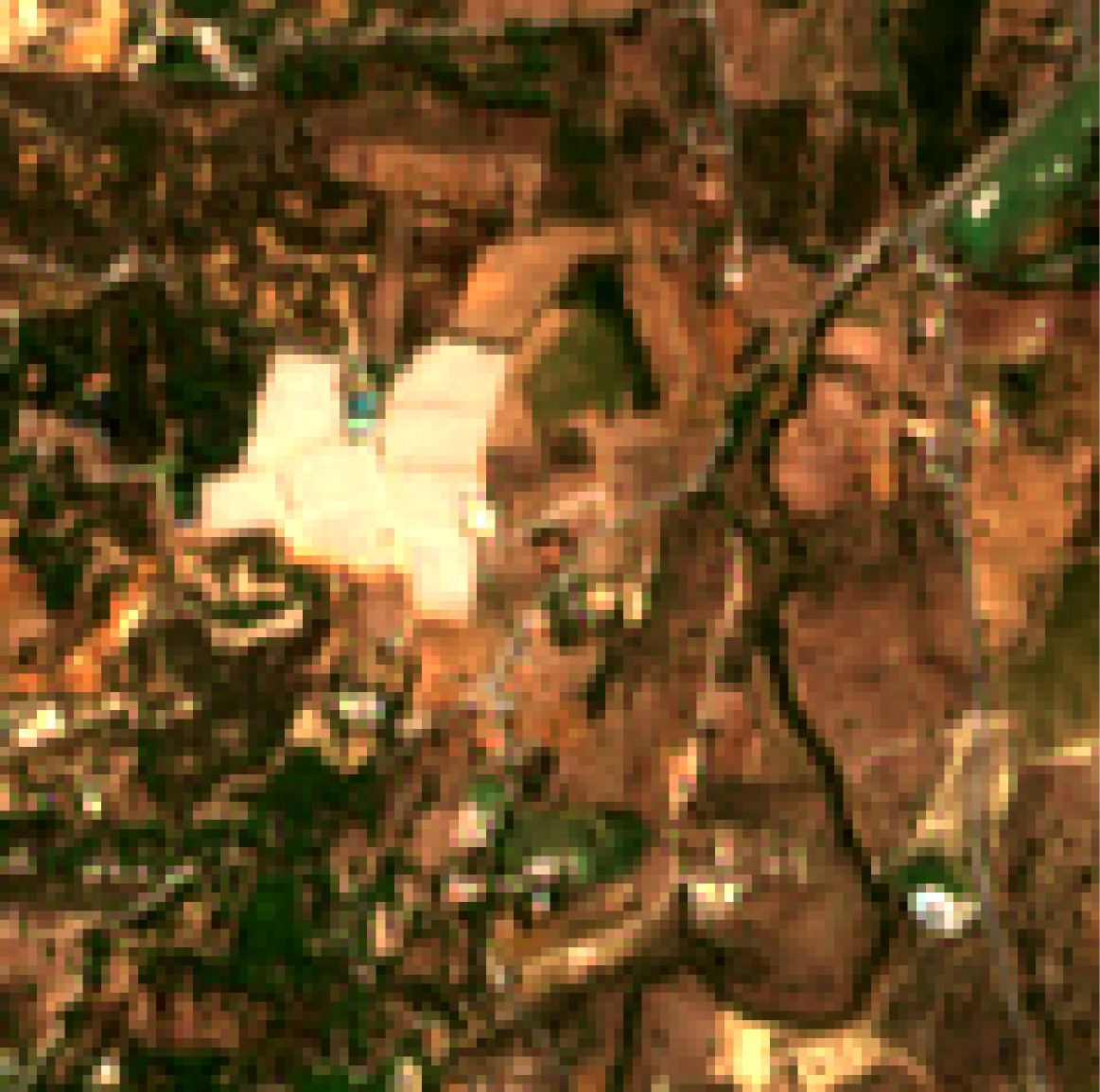}
     &
     \includegraphics[width=0.1\linewidth]{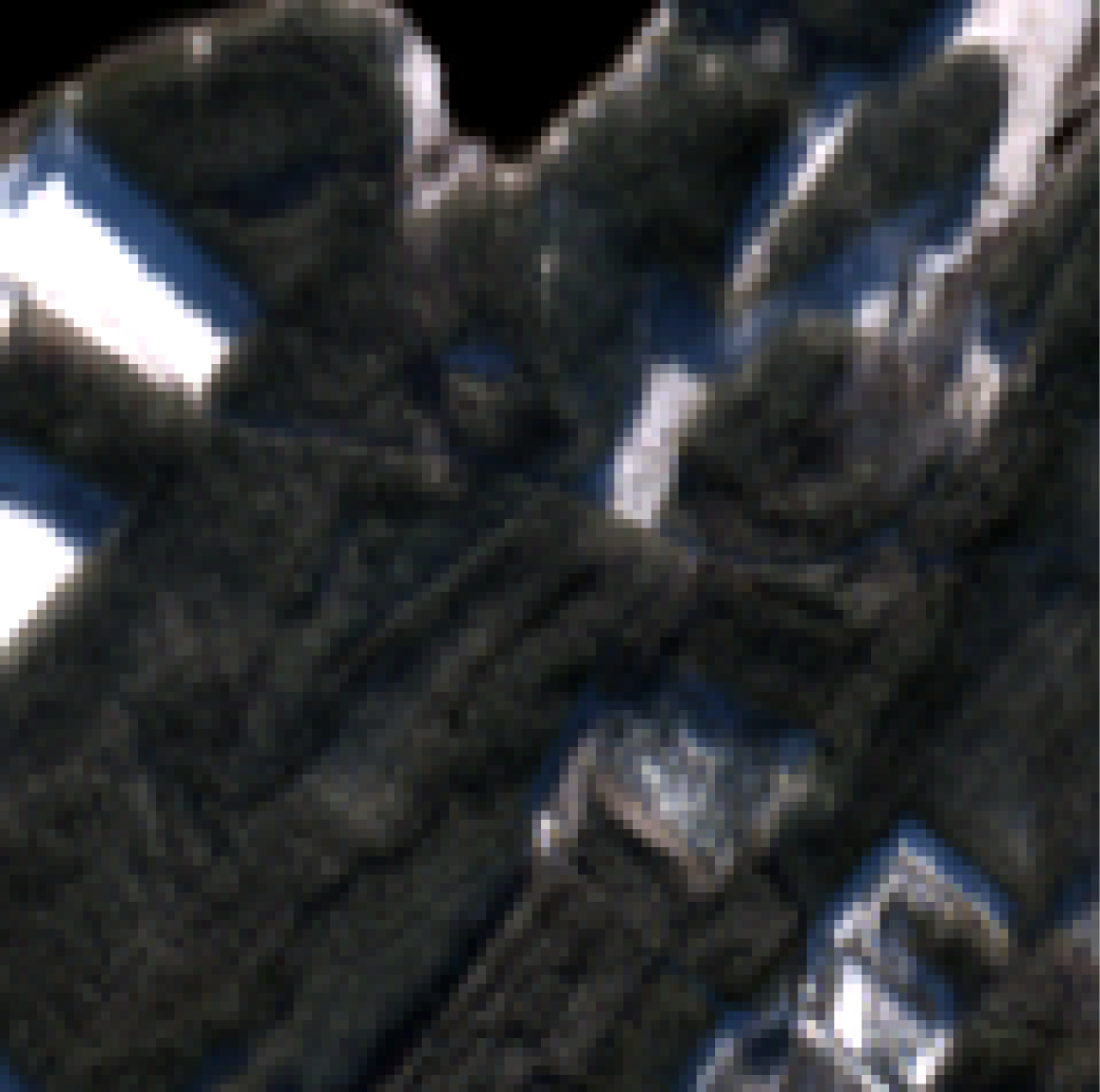}
     &
     &
     &
     \\\bottomrule
\end{tabular}

    \caption{Top-3 images by modality retrieved by the models for the provided query. We reported the RGB for Sentinel-2 and the VV polarization for Sentinel-1. Each image is coupled with the corresponding real relevance. Images are missing due to the use of approximated KNN in vector databases. \textbf{Query:} Bare. Built. Crops. Trees}
    \label{fig:query_318}
\end{figure}

\begin{figure}
    \centering
    \begin{tabular}{l|ccc|ccc}
     CLOSP-RN & 7 & 5 & 7 & 4 & 4 & 3\\
     &
     \includegraphics[width=0.1\linewidth]{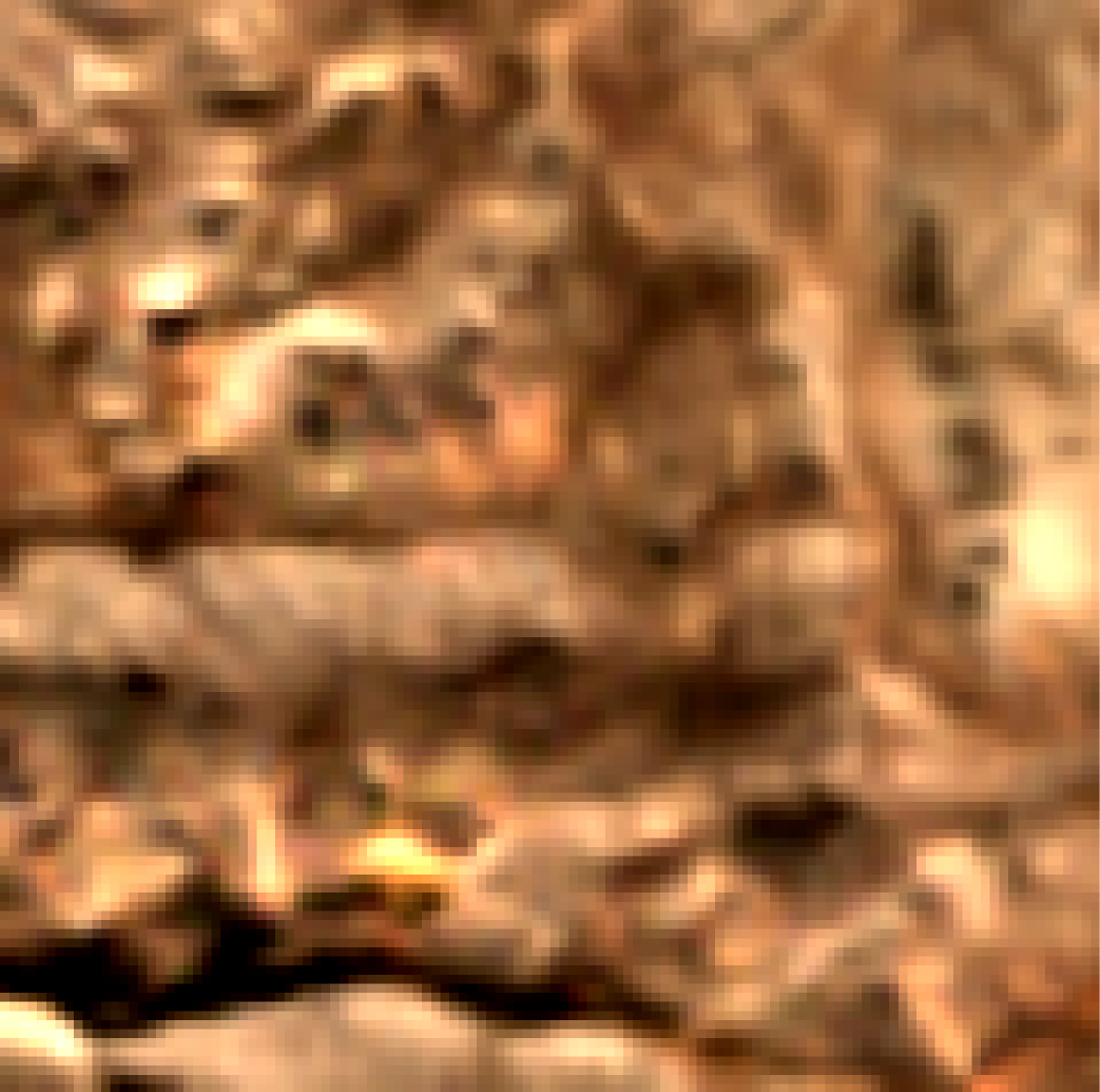}
     &
     \includegraphics[width=0.1\linewidth]{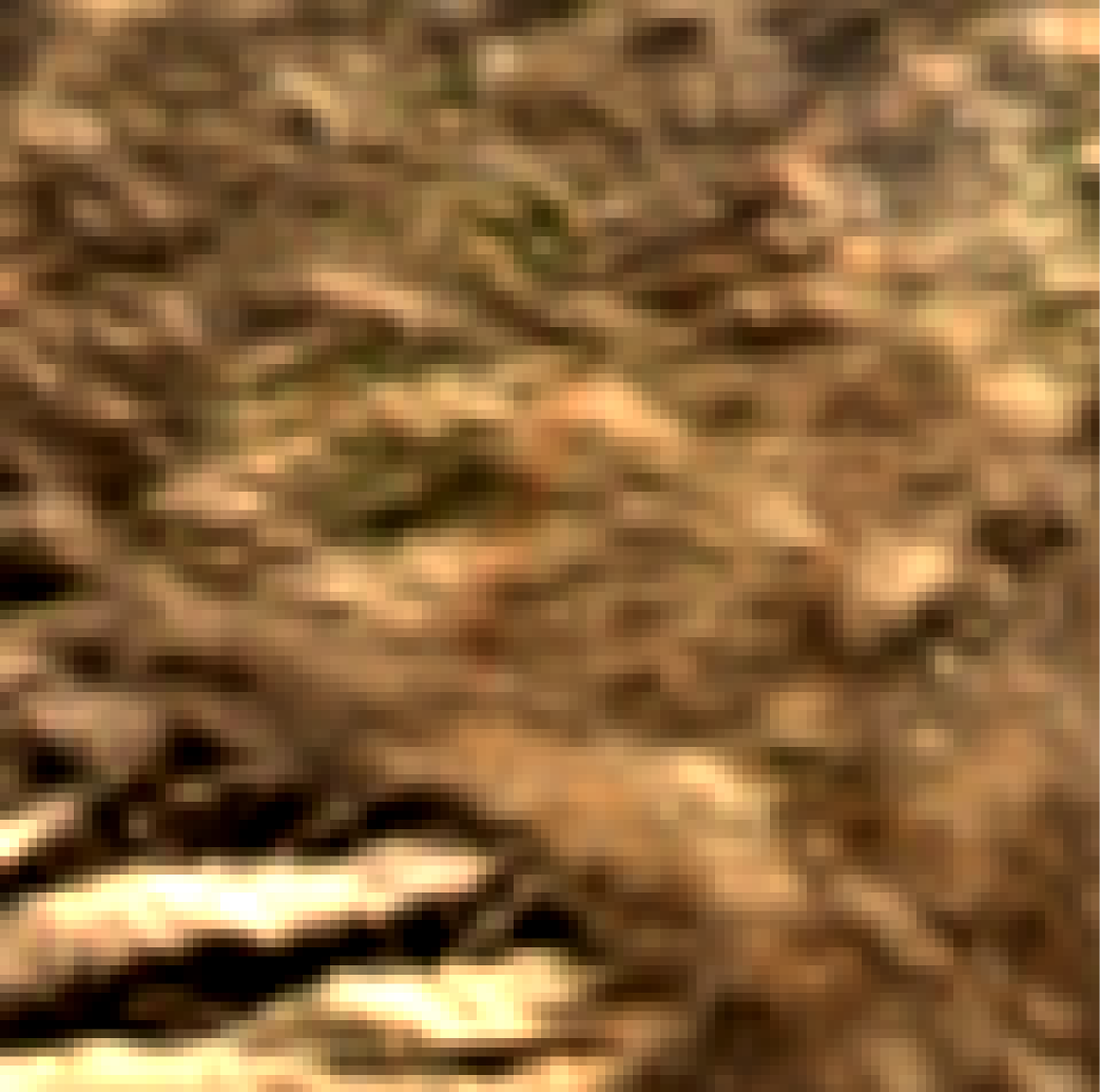}
     &
     \includegraphics[width=0.1\linewidth]{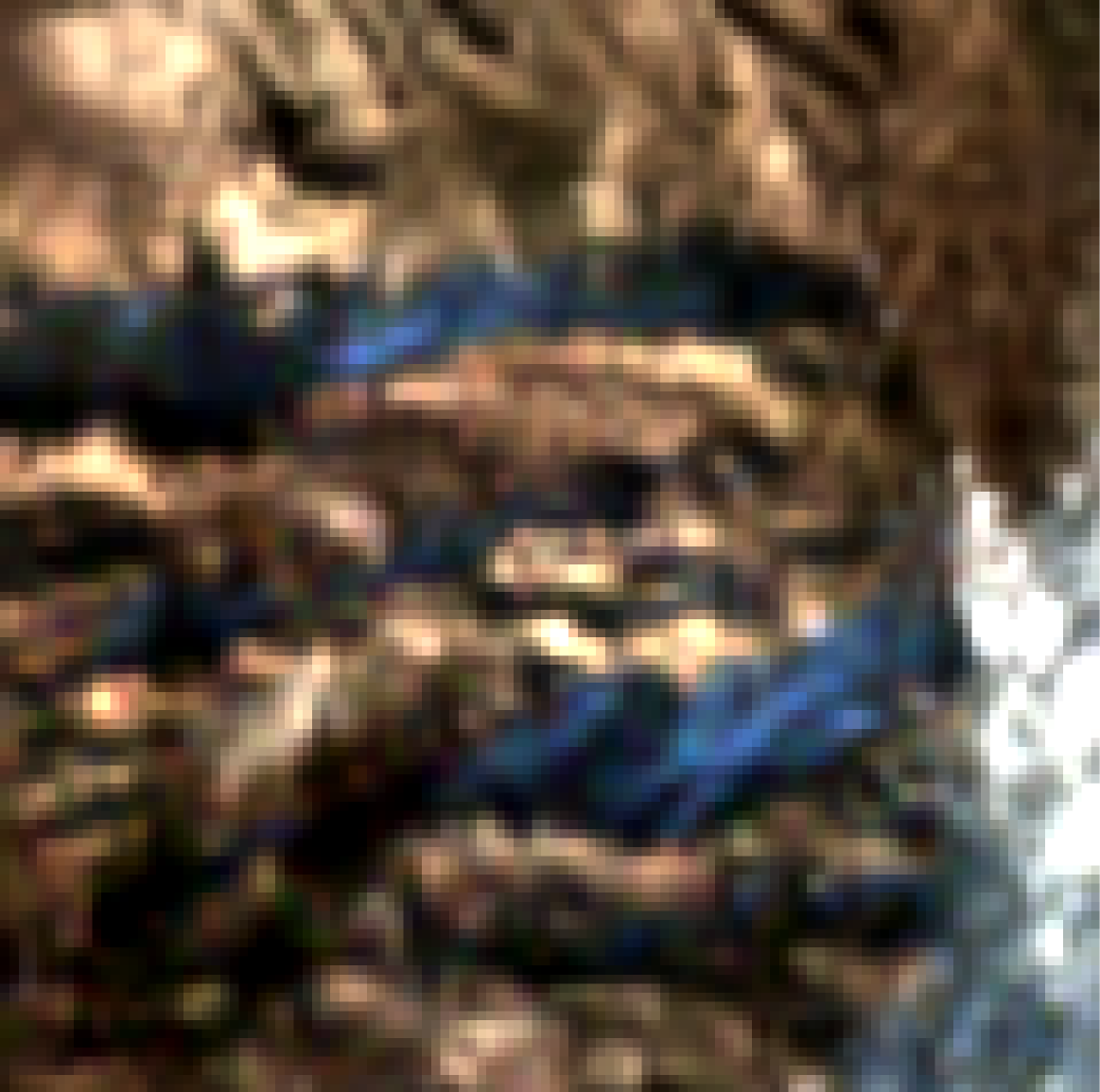}
     &
     \includegraphics[width=0.1\linewidth]{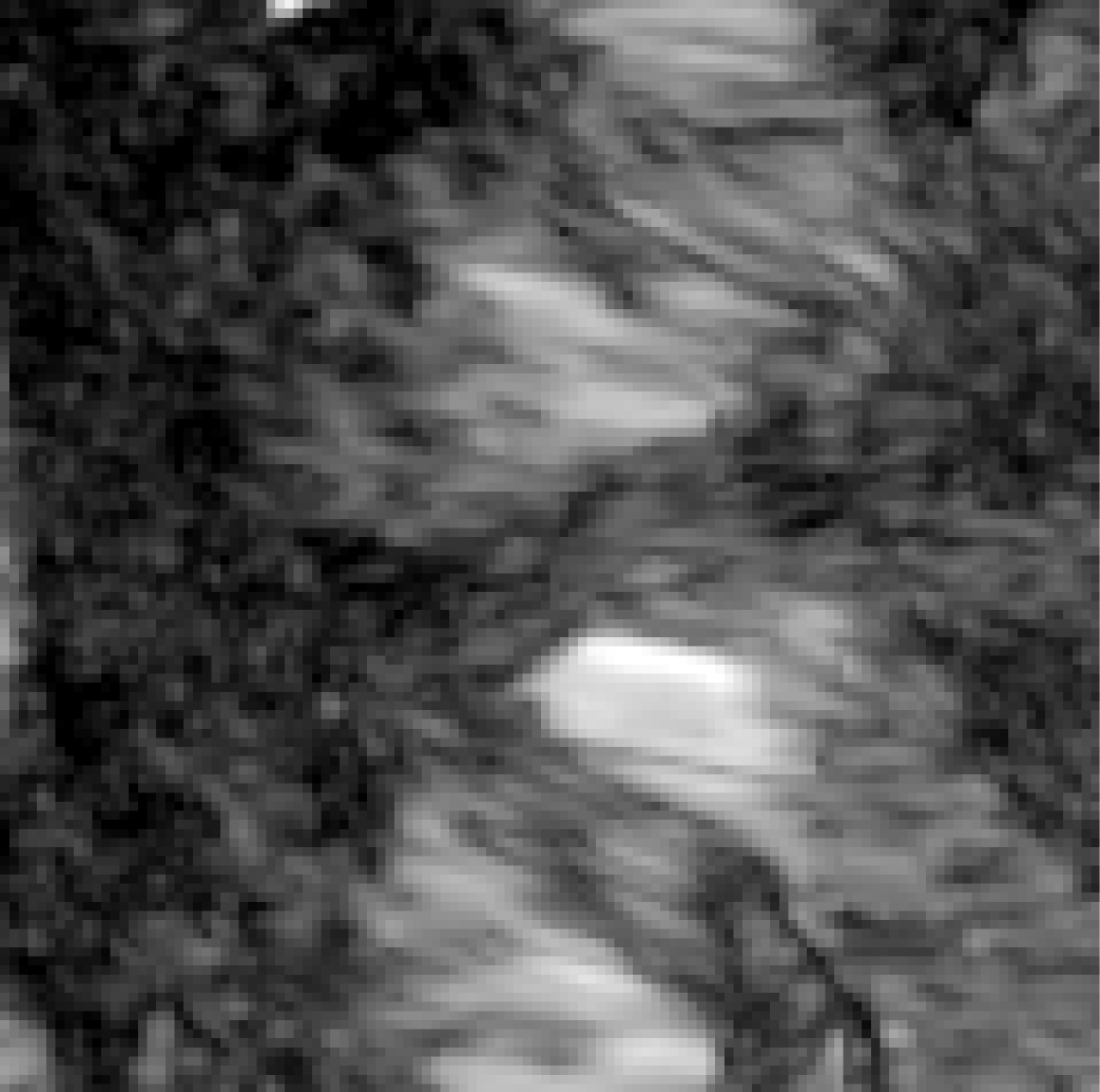}
     &
     \includegraphics[width=0.1\linewidth]{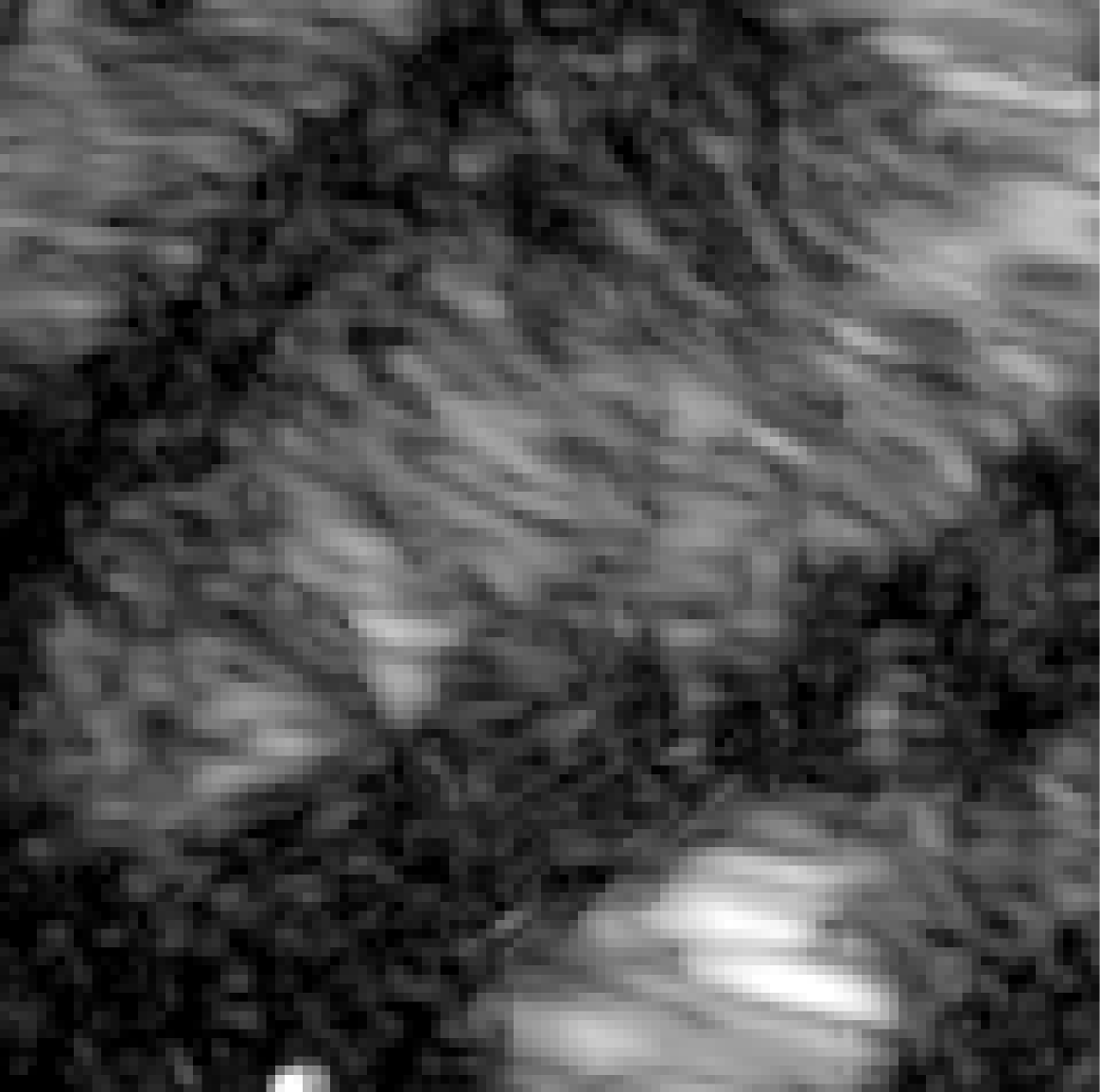}
     &
     \includegraphics[width=0.1\linewidth]{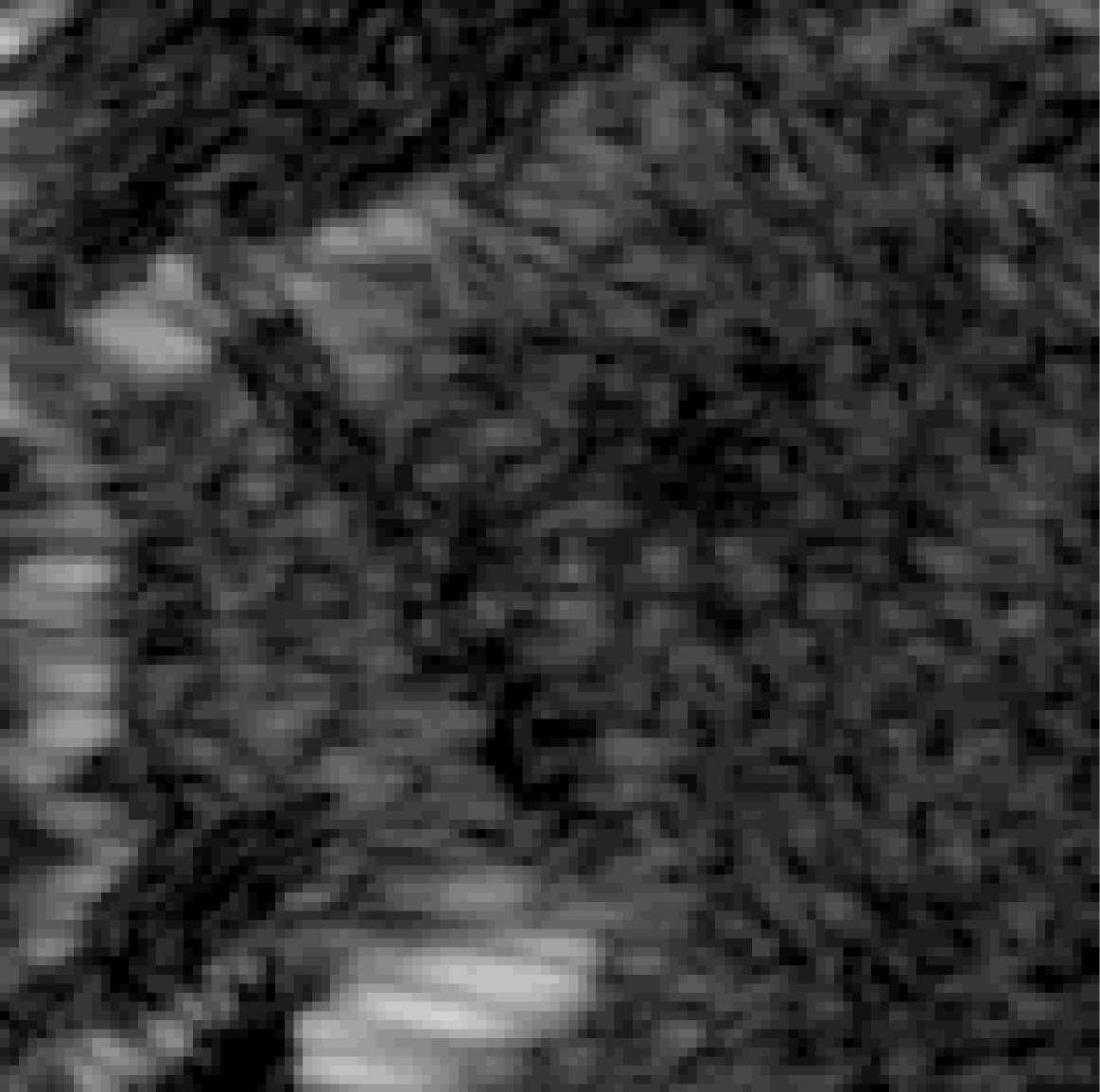}
     \\\midrule
     BiCLIP & 4 & 5 & 6 & 3 & 3 & 3\\
     &
     \includegraphics[width=0.1\linewidth]{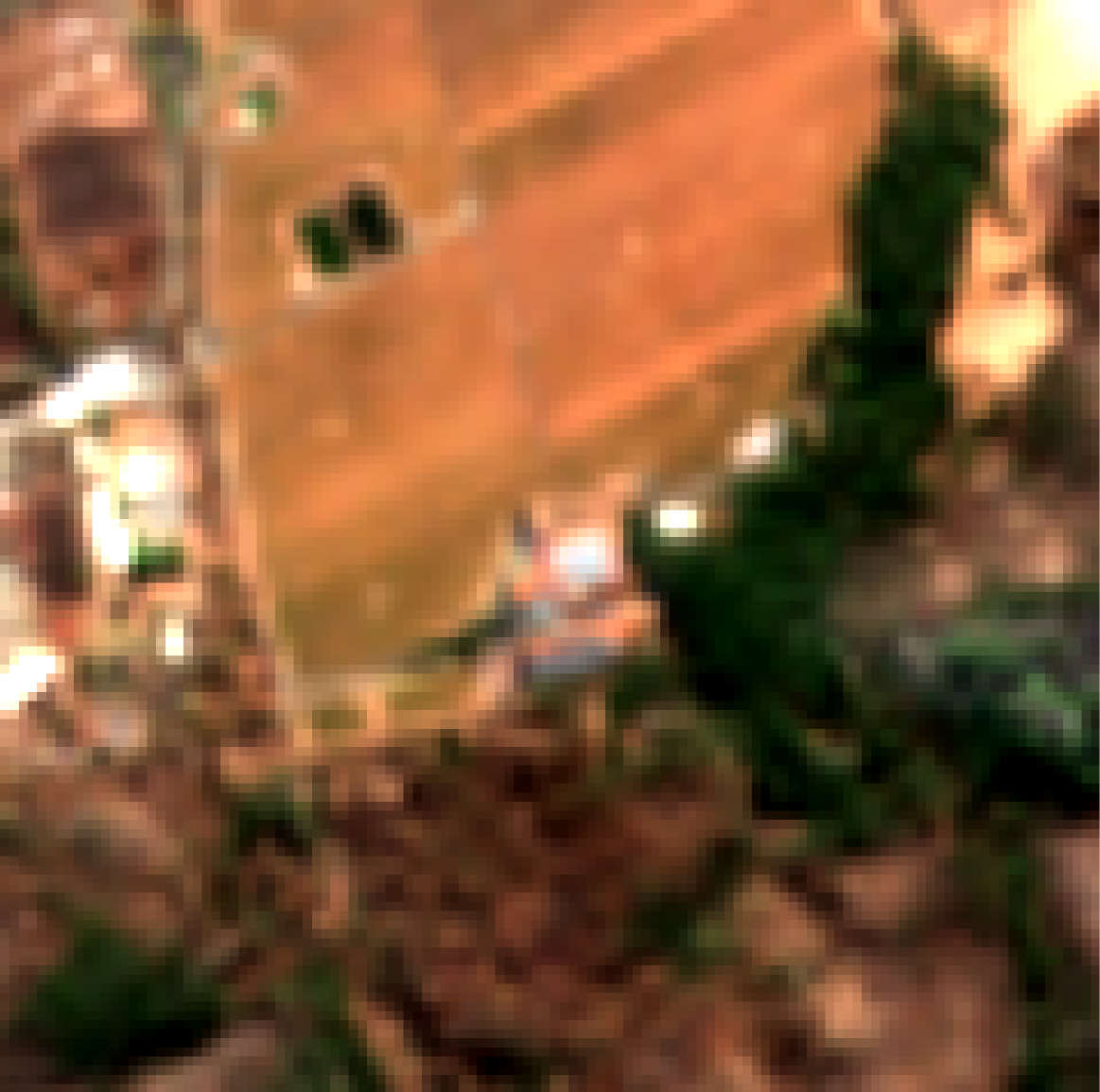}
     &
     \includegraphics[width=0.1\linewidth]{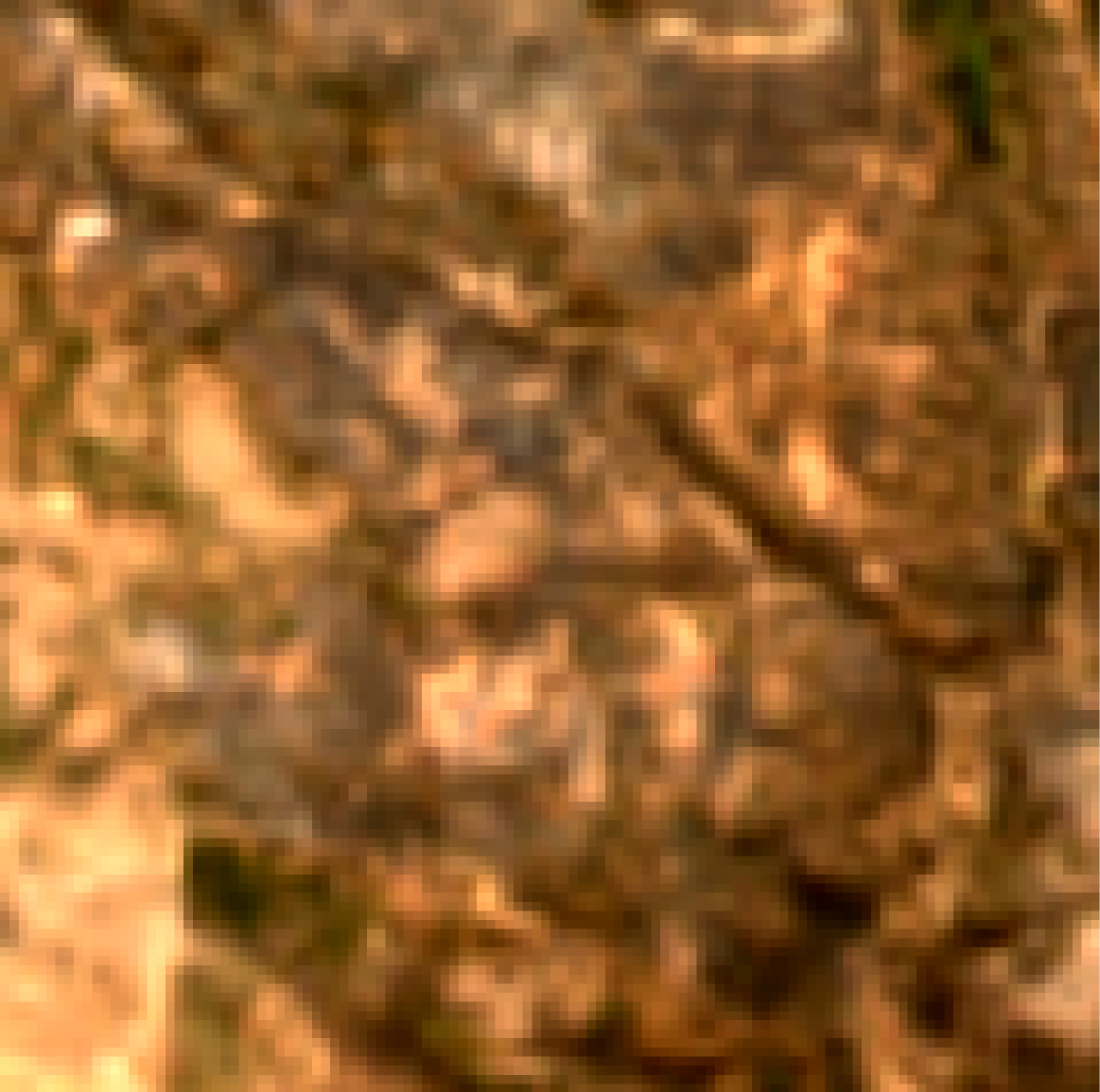}
     &
     \includegraphics[width=0.1\linewidth]{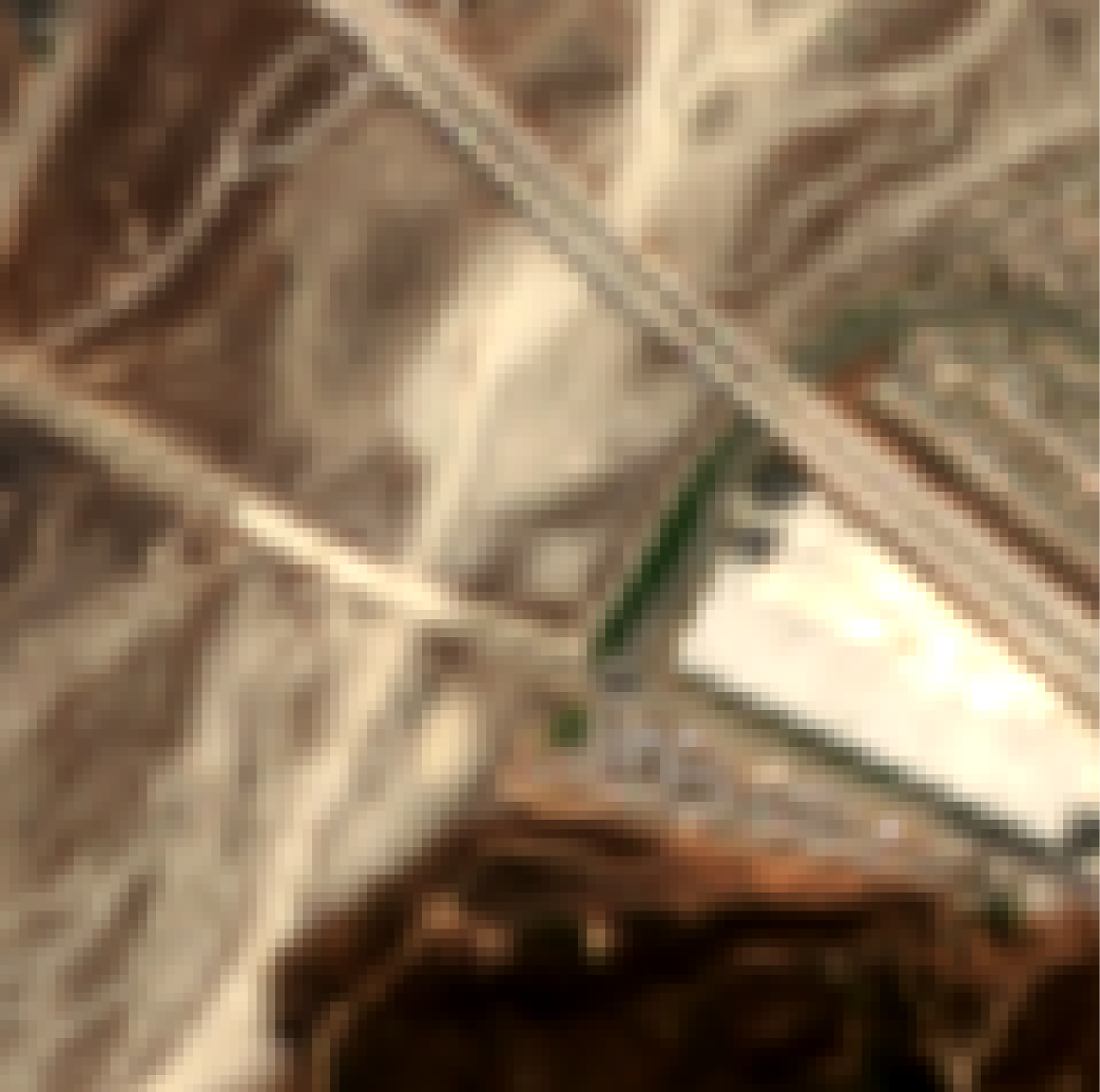}
     &
     \includegraphics[width=0.1\linewidth]{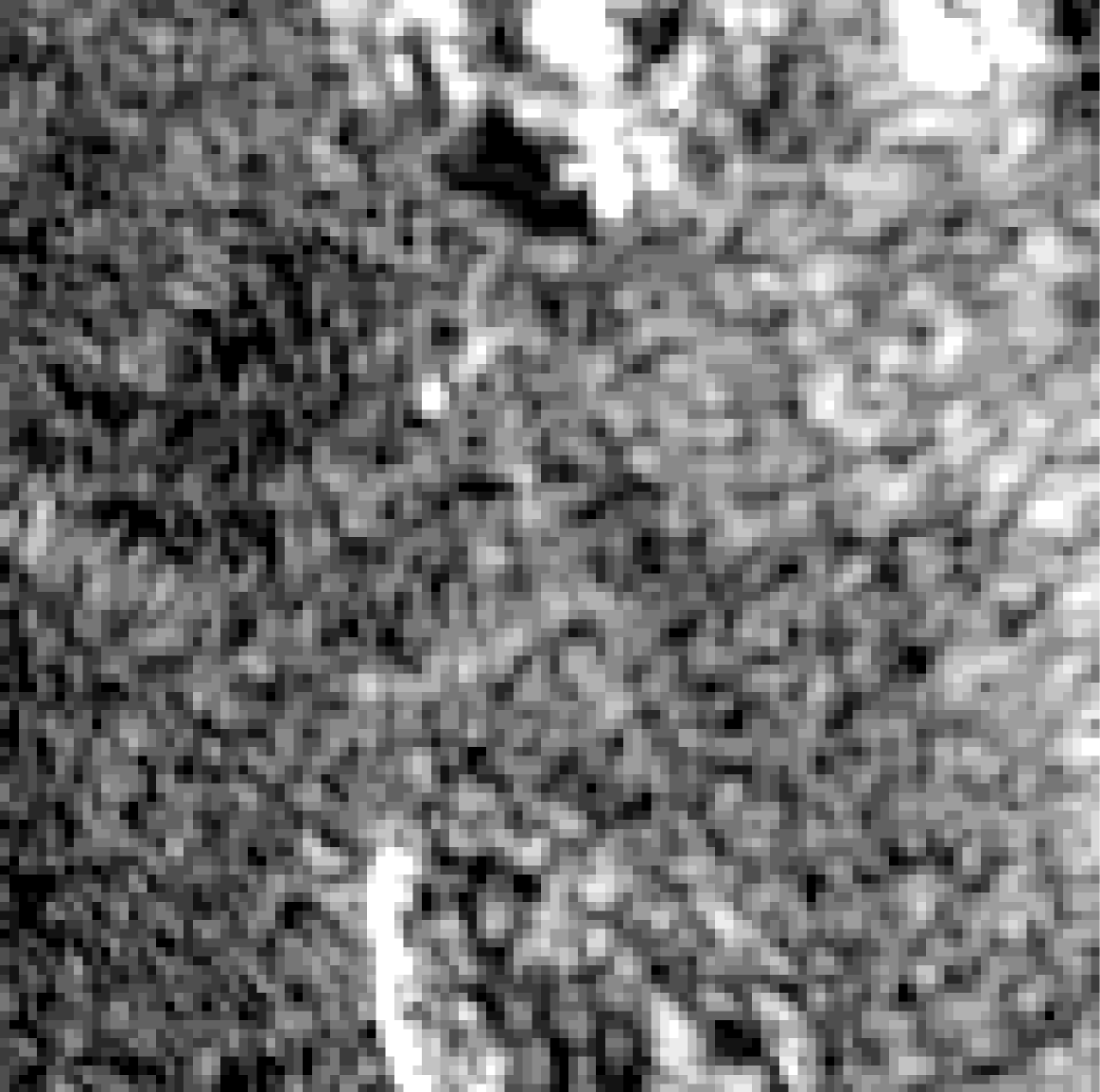}
     &
     \includegraphics[width=0.1\linewidth]{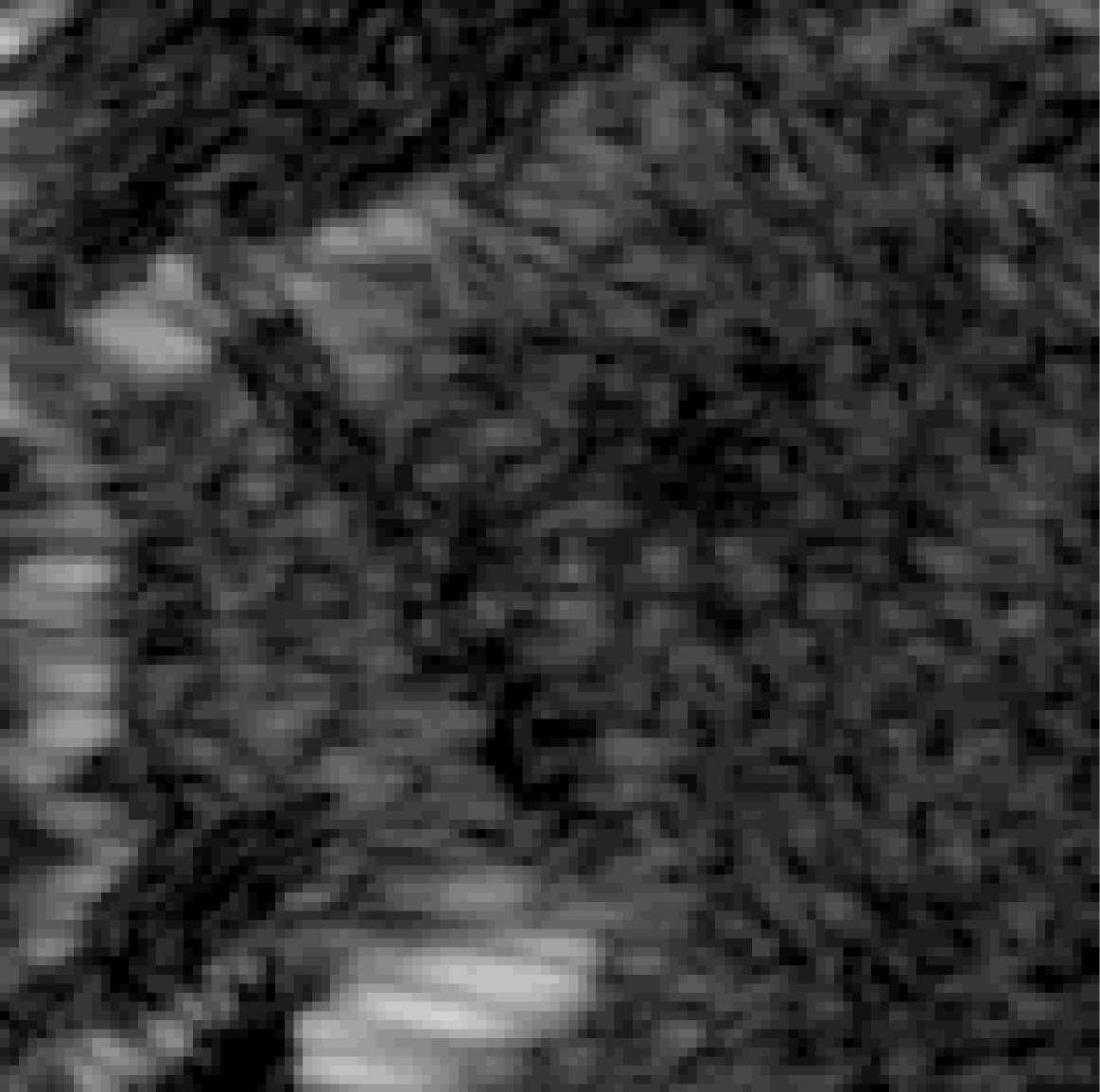}
     &
     \includegraphics[width=0.1\linewidth]{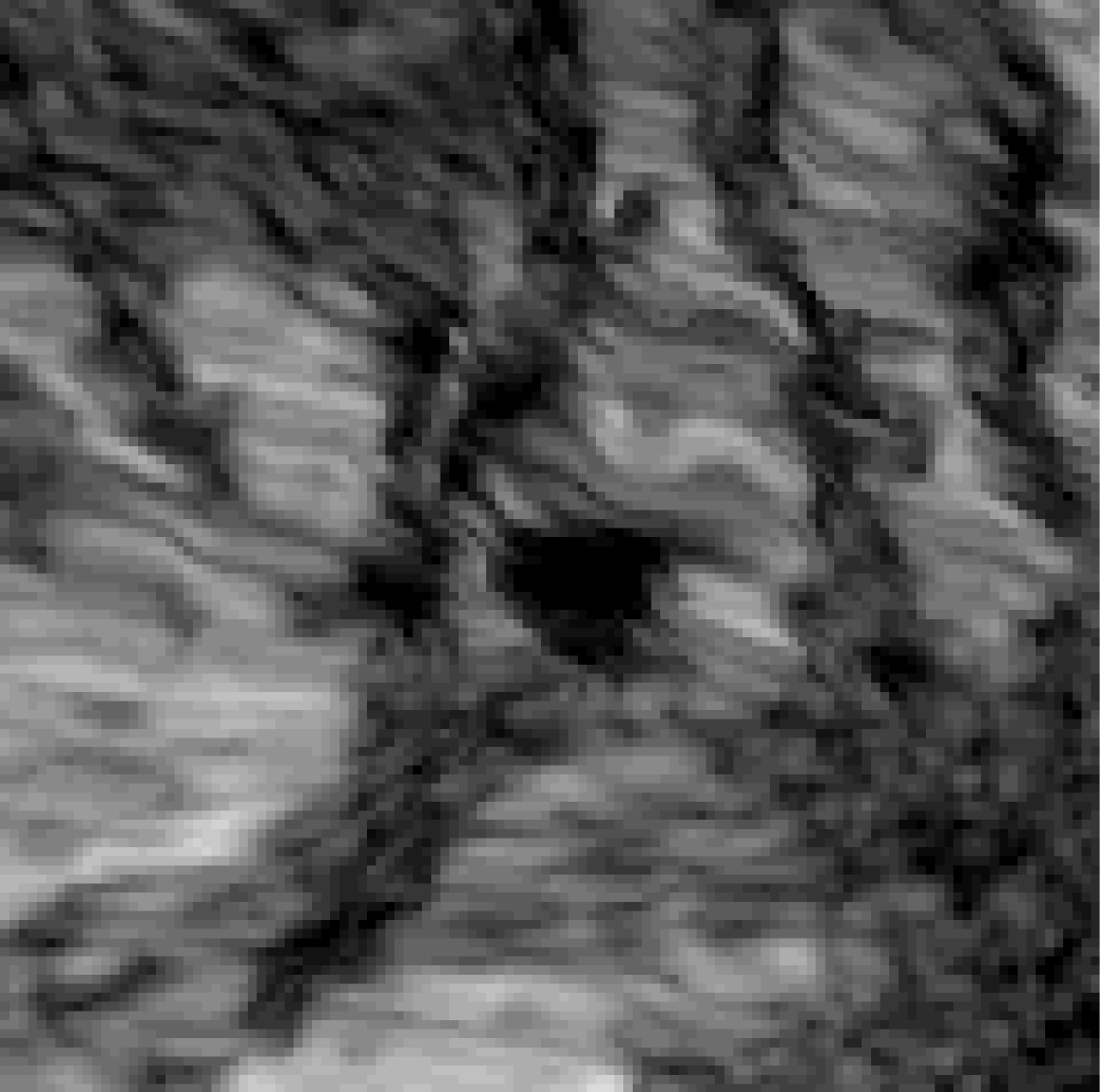}
     \\\midrule
     SkyCLIP-T & - & - & - & 1 & 2 & 1 \\
     &
     &
     &
     &
     \includegraphics[width=0.1\linewidth]{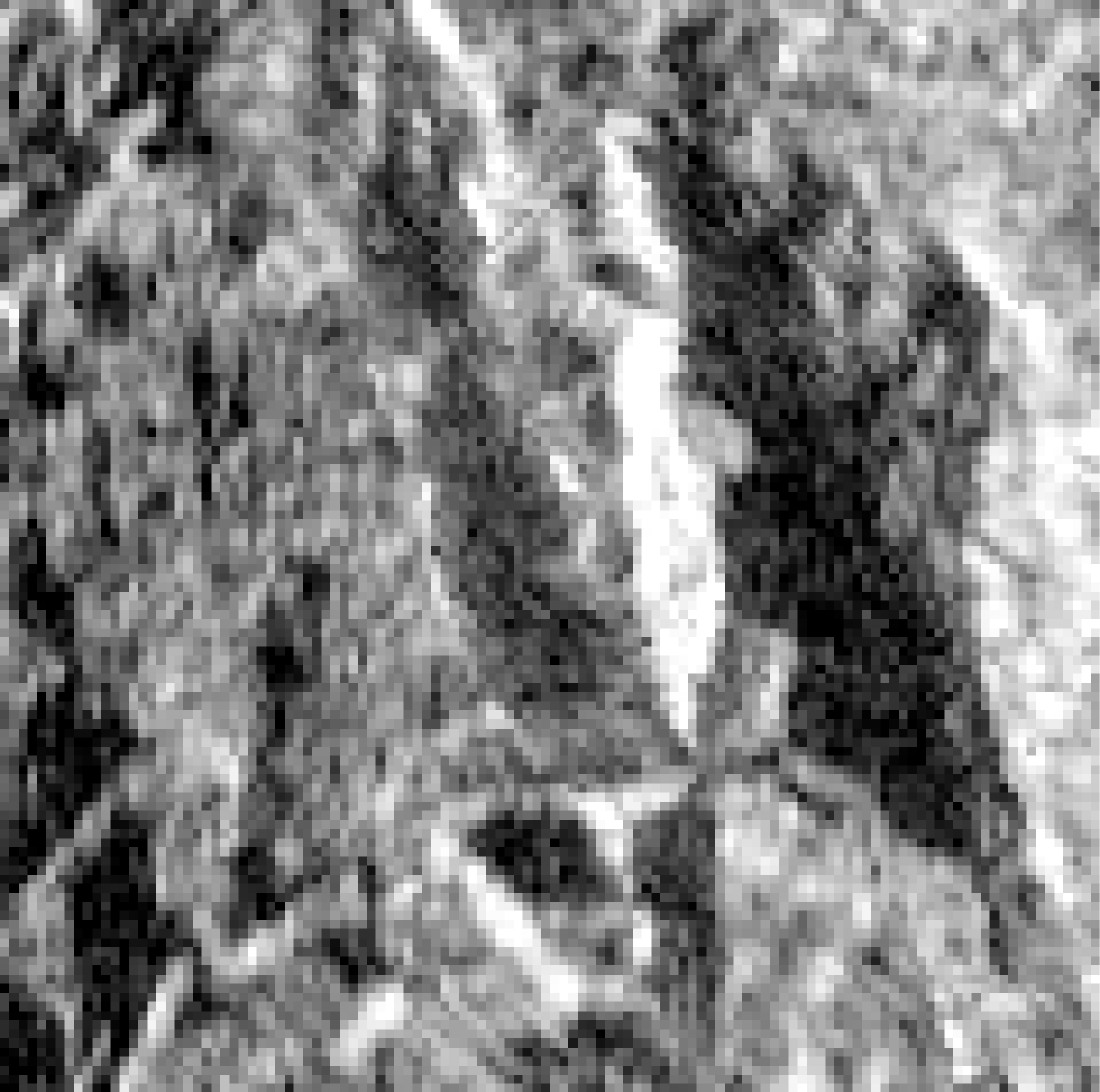}
     &
     \includegraphics[width=0.1\linewidth]{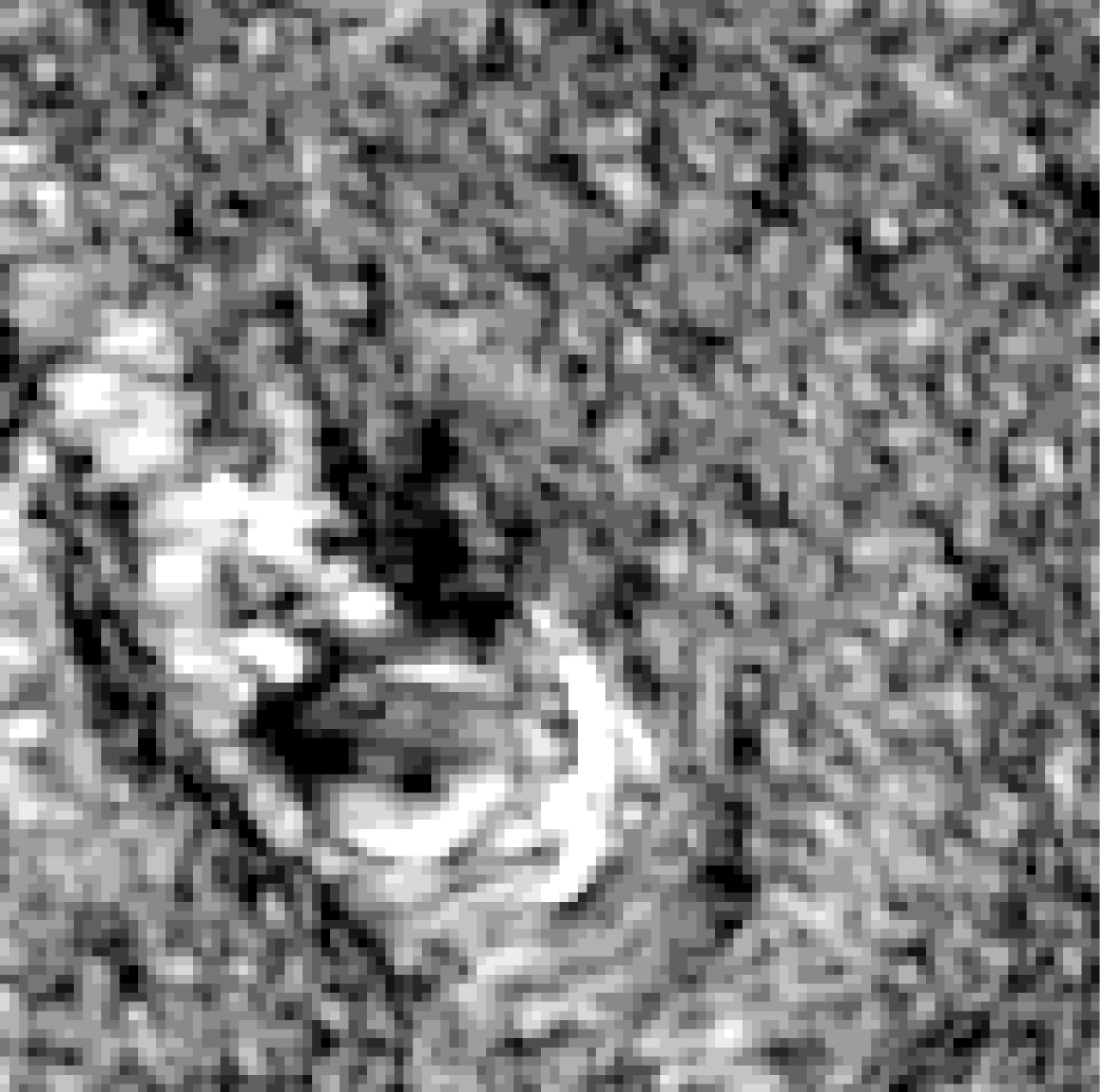}
     &
     \includegraphics[width=0.1\linewidth]{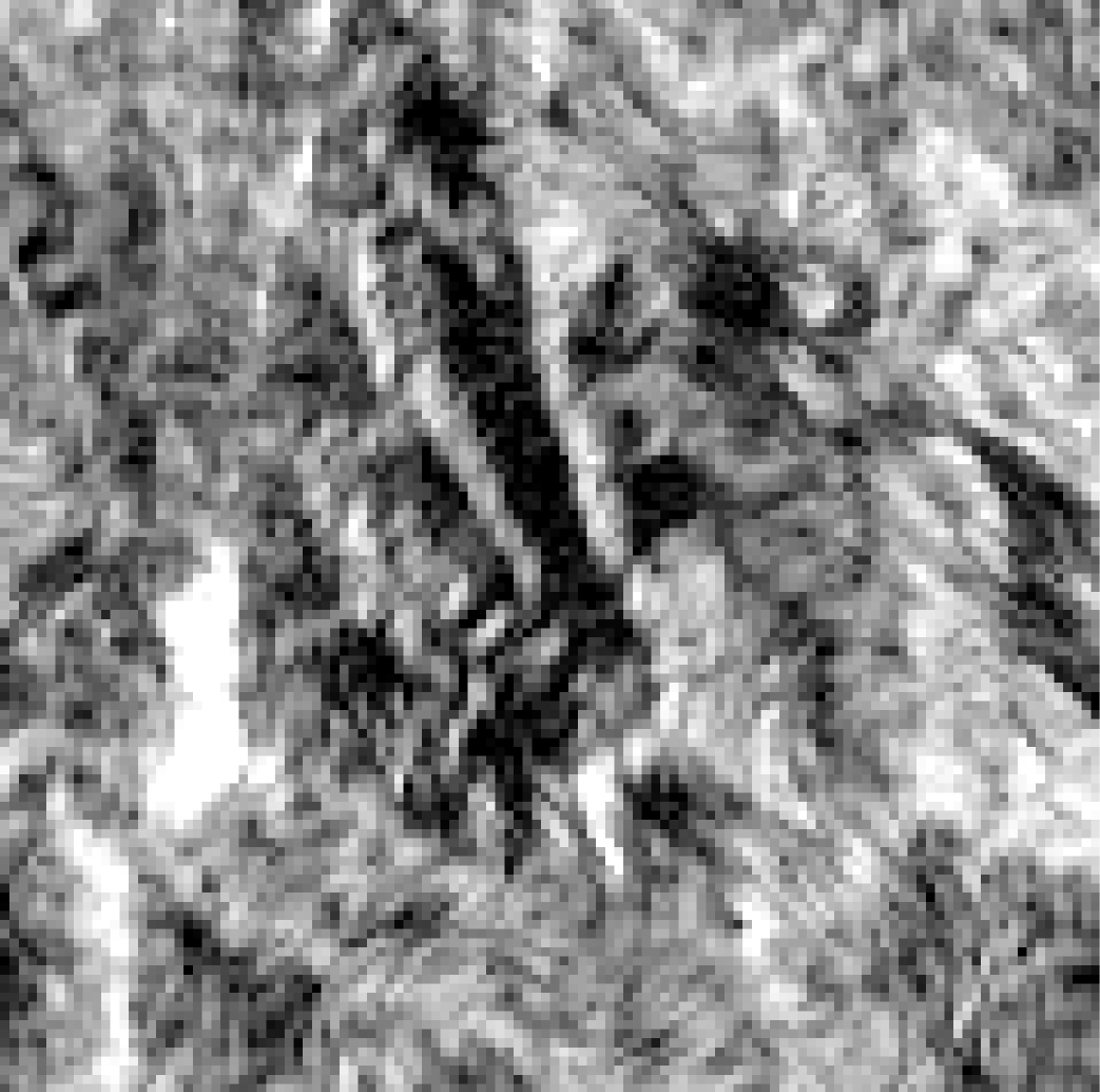}
     \\\bottomrule
\end{tabular}

    \caption{Top-3 images by modality retrieved by the models for the provided query. We reported the RGB for Sentinel-2 and the VV polarization for Sentinel-1. Each image is coupled with the corresponding real relevance. Images are missing due to the use of approximated KNN in vector databases. \textbf{Query:} Bare. Burned area. Crops. Shrub and scrub. Snow and ice. Trees}
    \label{fig:query_1503}
\end{figure}

We reported some visual examples of retrieved images for increasingly complex queries in \Cref{fig:query_35,fig:query_318,fig:query_1503}. It is possible that nothing will be retrieved since vector databases (i.e., ChromaDB) use approximated KNN to navigate a large dataset. Similar images correspond to the same area collected at different timestamps.

\section{Corine LC mapping}
\label{sec:clc_mapping}
In \Cref{tab:clc_mapping}, we present the mapping used between CORINE Land Cover classes and the Dynamic World classes.

\begin{longtable}{ll}
\caption{Mapping of CORINE Land Cover (CLC) classes to Dynamic World (DW) classes.}
\label{tab:clc_mapping} \\
\toprule
\textbf{CORINE Land Cover Class} & \textbf{Dynamic World Class} \\
\midrule
\endfirsthead

\multicolumn{2}{c}%
{{\tablename\ \thetable{} -- continued from previous page}} \\
\toprule
\textbf{CORINE Land Cover Class} & \textbf{Dynamic World Class} \\
\midrule
\endhead

\bottomrule
\multicolumn{2}{r}{{Continued on next page}} \\
\endfoot

\endlastfoot

Continuous urban fabric & Built \\
Discontinuous urban fabric & Built \\
Industrial or commercial units & Built \\
Road and rail networks and associated land & Built \\
Port areas & Built \\
Airports & Built \\
Mineral extraction sites & Bare \\
Dump sites & Bare \\
Construction sites & Bare \\
Green urban areas & Grass \\
Sport and leisure facilities & Grass \\
Non-irrigated arable land & Crops \\
Permanently irrigated land & Crops \\
Rice fields & Flooded vegetation \\
Vineyards & Crops \\
Fruit trees and berry plantations & Crops \\
Olive groves & Crops \\
Pastures & Grass \\
Annual crops associated with permanent crops & Crops \\
Complex cultivation patterns & Crops \\
Land principally occupied by agriculture, with & Crops \\
significant areas of natural vegetation & \\
Agro-forestry areas & Trees \\
Broad-leaved forest & Trees \\
Coniferous forest & Trees \\
Mixed forest & Trees \\
Natural grassland & Grass \\
Moors and heathland & Shrub and Scrub \\
Sclerophyllous vegetation & Shrub and Scrub \\
Transitional woodland/shrub & Shrub and Scrub \\
Beaches, dunes, sands & Bare \\
Bare rock & Bare \\
Sparsely vegetated areas & Bare \\
Burnt areas & Burned Area \\
Glaciers and perpetual snow & Snow and Ice \\
Inland marshes & Flooded vegetation \\
Peatbogs & Flooded vegetation \\
Salt marshes & Flooded vegetation \\
Salines & Bare \\
Intertidal flats & Bare \\
Water courses & Water \\
Water bodies & Water \\
Coastal lagoons & Water \\
Estuaries & Water \\
Sea and ocean & Water \\
\end{longtable}

\bibliographystyle{cas-model2-names}
\bibliography{bibliography} 

\end{document}